\documentclass[pdflatex,sn-mathphys-num]{sn-jnl}%

\usepackage{graphicx}%
\usepackage{multirow}%
\usepackage{amsmath,amssymb,amsfonts}%
\usepackage{amsthm}%
\usepackage[title]{appendix}%
\usepackage{xcolor}%
\usepackage{textcomp}%
\usepackage{booktabs}%
\usepackage{tabularx}
\usepackage{array}
\usepackage{adjustbox}
\usepackage{float}
\usepackage{subcaption}%

\usepackage{hyperref}     %

\newcommand{\an}{\textcolor{black}}
\newcommand{\ans}{\textcolor{black}}

\begin{document}

\title{From Deterministic to Generative Deep Learning for Urban Air Quality Reconstruction from Sparse Observations}

\author[1,2]{\fnm{Abhishek Ajit} \sur{Sabnis}}
\email{abhishek-ajit.sabnis@enpc.fr}

\author[2]{\fnm{Mihai} \sur{Mitrea}}
\email{mihai.mitrea24@imperial.ac.uk}

\author[1]{\fnm{Lya} \sur{Lugon}}
\email{lya.lugon@enpc.fr}

\author[1]{\fnm{Karine} \sur{Sartelet}}
\email{karine.sartelet@enpc.fr}

\author[1]{\fnm{Marc} \sur{Bocquet}}
\email{marc.bocquet@enpc.fr}

\author[3]{\fnm{Xiaoyuan} \sur{Cheng}}
\email{xiaoyuan.cheng.21@ucl.ac.uk}

\author[4]{\fnm{Shupeng} \sur{Zhu}}
\email{shupengz@zju.edu.cn}

\author*[1]{\fnm{Sibo} \sur{Cheng}}
\email{sibo.cheng@enpc.fr}

\affil*[1]{\orgdiv{CEREA}, \orgname{ENPC, EDF R\&D, Institut Polytechnique de Paris}, \orgaddress{\city{Île-de-France}, \country{France}}}

\affil[2]{\orgdiv{Department of Computing}, \orgname{Imperial College London}, \orgaddress{\city{London}, \country{UK}}}

\affil[3]{\orgdiv{Dynamic Systems Lab}, \orgname{University College London}, \orgaddress{\city{London}, \country{UK}}}

\affil[4]{\orgdiv{Department of Atmospheric Sciences, School of Earth Sciences}, \orgname{Zhejiang University}, \orgaddress{\city{Hangzhou}, \country{China}}}

\abstract{Full-field reconstruction of air pollution is essential for evaluating pollution exposure and supporting public health decision-making. However, the complex interactions among pollutants, hard-to-predict weather patterns, and limited monitoring-station coverage make this a complex task. We apply deep learning techniques to provide fast and accurate reconstructions from sparse observations of four key pollutants: NO$_2$, O$_3$, PM$_{2.5}$ and PM$_{10}$. Models are trained on full-field simulation data and evaluated on real-world observations collected from $9$ to $28$ monitoring stations in the city of Paris. We introduce a diffusion-based generative framework for multi-pollutant reconstruction and benchmark its performance against deterministic deep learning models. Despite noisy observations and strong spatial variability, the models achieve high structural similarity on simulated \an{test} data and \an{produce spatial patterns on real-world observations that are consistent with the simulated fields.} We introduce data augmentation methods that enable transfer to real-world observations without \an{requiring fine-tuning}, allowing the models to generalise beyond the training period. \an{These findings highlight the potential of ML models for air pollution reconstruction under extremely sparse and noisy observation conditions.}}


\maketitle

\section*{Introduction}\label{Intro}

Increased human activity makes it difficult to predict variations in air quality. Exposure to pollutants such as nitrogen dioxide (NO$_2$), particulate matter (PM$_{10}$, PM$_{2.5}$) and ozone (O$_3$) is strongly associated with respiratory illness~\cite{respiratory_illness}, cardiovascular disease~\cite{cardiovascular}, and premature mortality~\cite{mortality}. Although global PM$_{10}$ concentration tends to decrease, PM$_{2.5}$, NO$_2$ and O$_3$ have increased in 65\%, 71\% and 89\% of cities around the world~\cite{SICARD2023}. Accurate estimation of pollutant concentrations is essential for assessing pollution exposure, addressing associated health impacts~\cite{brauer2016ambient}, and helping policymakers design effective measures~\cite{barwick2024fog}.

Deterministic modelling with chemistry-transport (CT) models can be used to estimate pollution exposure by providing spatially and temporally resolved concentration fields \citep{PASCAL2016, sartelet2025}. However, these estimates remain subject to uncertainties arising from several sources, including emission inventories, meteorological fields, and the representation of atmospheric chemical and physical processes. To reduce these uncertainties, simulated concentrations can be corrected using ground-based observations through spatial interpolation-based methods, such as kriging~\citep{SHUKLA2020}, or data assimilation techniques~\citep{tilloy2013, sartelet2025}. These approaches are generally associated with higher computational costs and rely on precise prior knowledge of error distributions and their spatial correlations, which limits their applicability in real-time monitoring scenarios.

To address this, we investigate the reconstruction of spatial fields directly from sensor observations, without relying on CT models or data assimilation during inference. While data from monitoring stations provide us with real-time observation values, these stations are usually sparse and unevenly distributed. High-precision ground stations are expensive to deploy and maintain. By contrast, low-cost sensors that are more scalable often suffer from limited accuracy, higher sensitivity, and require regular calibration~\citep{hayward2024low, stowe2026assessing}.
Additionally, pollutant concentration fields can exhibit abrupt transitions, leading to nonlinear and nonstationary spatial patterns that vary across pollutants~\cite{he2022spatial}. Urban air-quality reconstruction remains an intrinsically ill-posed inverse problem
~\cite{hase2017atmospheric,rodgers2000inverse} without prior simulations. Sparse monitoring observations do not uniquely determine the underlying pollution field because multiple atmospheric states can be consistent with the same set of measurements. Because observations are often sparse and noisy, there is a need for reconstruction methods that remain robust under these conditions.

In recent years, the use of machine learning techniques has shown rapid progress in the reconstruction of air pollution fields~\cite{le2020spatiotemporal,sasaki2022airex,feng2024spatio}. Although these methods have higher reconstruction accuracy than traditional approaches, several important limitations remain unaddressed. First, most studies model pollutants individually or train separate models for each~\cite{le2020spatiotemporal}, ignoring strong inter-pollutant correlations~\cite{seinfeld2016atmospheric}. Second, studies using methods trained on synthetically generated full-field data rarely evaluate the reconstruction quality on real-world observations, as noisy measurements limit model transfer to new data. Moreover, real-world sensor networks switch on and off over time, leading to temporally incomplete observations. This practical challenge is often overlooked in existing modelling frameworks~\cite{santos2023development}.

In order to address these highlighted gaps, we adopt a strategy that jointly models multiple pollutants and performs an ML-based reconstruction. We work with two different datasets, the \textit{simulation} data to train the model and \textit{observation} data to evaluate on real-world air quality measurements from the Paris metropolitan area. \an{The simulations account for pollutant emissions, including road traffic, meteorological conditions, urban morphology, and initial and boundary conditions. Lugon et al.~\cite{lugon2022} evaluated Polyphemus (simulation model) concentrations against measurements at urban stations in Paris for 2014 and reported agreement with the urban model-performance criteria adopted in their study for PM${10}$, PM${2.5}$, NO$_2$. This evaluation supports the use of the simulations as a physically informed reference, while not implying that the simulated fields fully represent the real-world concentrations.} To provide a spatially consistent representation of the sensor network to the ML model, Voronoi tessellations are constructed. This also mitigates the issues arising from temporally missing observations.

The main contribution of this study is the application of a diffusion-based generative model~\cite{song2020score} to enhance the reconstruction of air pollution fields. The diffusion models generate multiple solutions guided by sparse observations. This stochastic formulation enables reconstruction of finer details, particularly in regions of high spatial variability of pollution fields. We compare this approach with various deterministic models. To our knowledge, this study is among the first to reconstruct high resolution spatial air pollution fields using a diffusion- or flow-based generative AI model~\cite{yang2023diffusion}. Previous studies have used diffusion models to perform reconstruction of pollution in the temporal domain~\cite{liu2023pristi} (interpreted as signal recovery). Others have performed reconstruction in the spatial domain~\cite{zheng2026noise} without training on full-space data. These methods perform probabilistic inference at unknown sensor locations rather than supervised learning to predict the reference field.

In this paper, the evaluation of the reconstruction is conducted on an independent dataset of observed real-world sensor readings. To provide a fair assessment and evaluate the model’s generalisability \ans{beyond the training period}, the models are trained on simulation data for 10 months, from January to October 2014, \an{validated on November 2014 for hyperparameter tuning and tested on out-of-distribution real-world observations in December 2014.} Since full-field spatial data is not available for the observation dataset, we introduce point-wise cross-validation and spectral plots to evaluate the model's performance. \ans{The results demonstrate that our models generalise well to real-world data. Combined with their ability to generate reconstructions in real-time after training, these findings highlight the potential of ML models for monitoring and mapping of operational urban air quality.}

\begin{figure}
    \centering
    \includegraphics[width=1\linewidth]{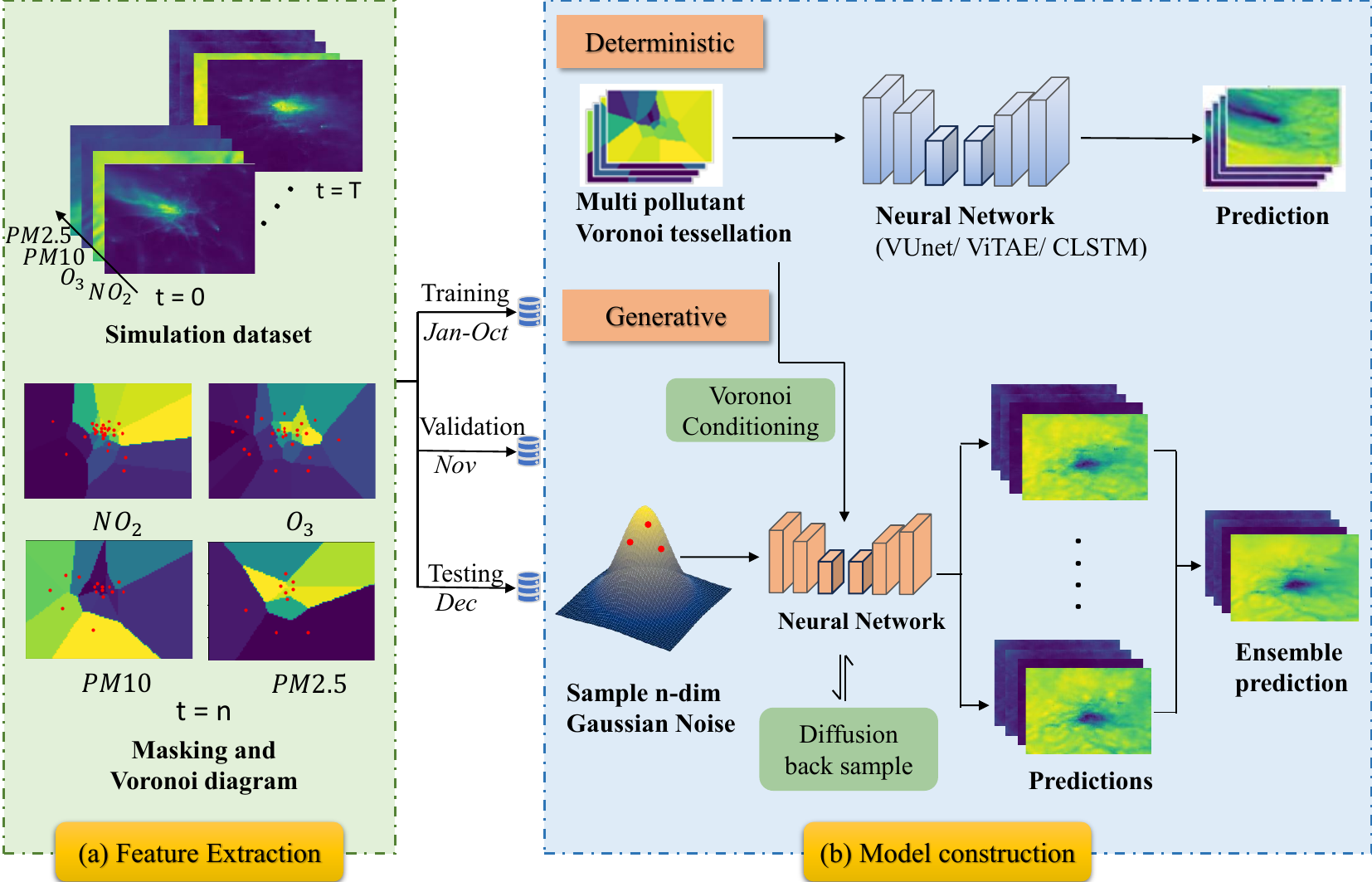}
    \caption{\textbf{Overview of model training.} Illustration of the training procedure of various models using simulation data. Panel (a) illustrates the feature extraction step which creates masks and generates Voronoi tessellations for simulation data at each time step. Panel (b) illustrates the differences in output generation by deterministic and generative models.}
    \label{fig:train_flow}
\end{figure}

\section*{Results}\label{result}

\subsection*{Performance of deterministic models on simulation data}\label{result_determi}

The ML models are trained and evaluated using full-field simulation data generated by Polyphemus~\cite{lugon2022}, a chemistry-transport model. The deterministic deep learning models include VUNet~\cite{RefWorks:RefID:4-fukami2021global,ronneberger2015u}, ViTAE~\cite{RefWorks:RefID:3-fan2025vitae-sl:}, and CLSTM~\cite{peehole_lstm,shi2015convolutional}, each selected for its different strengths in processing complex spatio-temporal datasets (more details in the "Model architectures" section in Supplementary materials). The results of the Kriging interpolation method~\cite{RefWorks:RefID:36-kleijnen2009kriging}, widely used in operational scenarios, are provided as a baseline. Details of model construction and data processing are provided in the "Methods" section. The previous $k$ hourly observations are used to reconstruct the field at the current time step. This experiment (Table \ref{tab:simulation_results}) investigates the performance of models when considering time windows of different lengths, $k = \{1,2,3,4,6,8,12 \}$.

\newcommand{\nmetrics}{4}

\begin{table}[htbp]
    \centering
    \setlength{\tabcolsep}{3pt}
    \renewcommand{\arraystretch}{1.2}
    \begin{adjustbox}{max width=\linewidth,center}
\begin{tabular}{ll*{7}{c}}
        \toprule
        \multirow{2}{*}{\textbf{Model}} &
        \multirow{2}{*}{\textbf{Metric}} &
        \multicolumn{7}{c}{\textbf{Number of input time steps}} \\
        \cmidrule(lr){3-9}
        & & \textbf{1} & \textbf{2} & \textbf{3} & \textbf{4} & \textbf{6} & \textbf{8} & \textbf{12}\\
        \midrule
        \multirow{\nmetrics}{*}{\textbf{CLSTM}}
            & \an{Relative L2} $\blacktriangledown$ & 0.130 & 0.120 & 0.118 & 0.117 & 0.115 & \textbf{0.112} & 0.114 \\
            & SSIM $\vartriangle$ & 0.563 & 0.598 & 0.606 & 0.617 & 0.626 & 0.636 & \textbf{0.637}\\
            & MFE $\blacktriangledown$ & 0.182 & 0.173 & 0.172 & 0.171 & 0.166 & 0.160 & \textbf{0.158} \\
            & MFB $\approxeq0$ & 0.012 & 0.030 & 0.029 & 0.029 & 0.011 & -0.009 & \textbf{-0.004} \\
        \midrule
        \multirow{\nmetrics}{*}{\textbf{VUNet}}
            & \an{Relative L2} $\blacktriangledown$ & 0.119 & 0.118 & 0.116 & \textbf{0.109} & 0.118 & 0.120 & 0.121\\
            & SSIM $\vartriangle$ & 0.669 & 0.674 & 0.679 & \textbf{0.685} & 0.667 & 0.659 & 0.663 \\
            & MFE $\blacktriangledown$ & 0.178 & 0.181 & \textbf{0.172} & 0.202 & 0.176 & 0.185 & 0.186 \\
            & MFB $\approxeq0$ & -0.009 & 0.005 & -0.004 & 0.044 & -0.009 & 0.035 & \textbf{0.003}\\
        \midrule
        \multirow{\nmetrics}{*}{\textbf{ViTAE}}
            & \an{Relative L2} $\blacktriangledown$ & 0.142 & 0.141 & 0.138 & 0.132 & \textbf{0.129} & 0.133 & 0.136 \\
            & SSIM $\vartriangle$ & 0.607 & \textbf{0.669} & 0.635 & 0.629 & 0.655 & 0.647 & 0.647 \\
            & MFE $\blacktriangledown$ & 0.195 & 0.195 & 0.198 & 0.184 & \textbf{0.182} & 0.185 & 0.182 \\
            & MFB $\approxeq0$ & 0.014 & -0.033 & -0.062 & 0.016 & -0.019 & \textbf{-0.007} & -0.013 \\

        \midrule
        \multirow{\nmetrics}{*}{\textbf{Kriging}}
            & \an{Relative L2} $\blacktriangledown$ & 0.422 \\
            & SSIM $\vartriangle$ & 0.506 \\
            & MFE $\blacktriangledown$ & 0.336 \\
            & MFB $\approxeq0$ & -0.113 \\

        \bottomrule
    \end{tabular}
\end{adjustbox}
    \caption{Performance of deterministic models when evaluated on the \an{test dataset} of the synthetic dataset with differing lengths of time windows. The best value for each metric within each model across the tested time-window lengths is bolded. The $\vartriangle$ symbol next to a metric indicates that a higher value is considered better, $\blacktriangledown$ indicates that a lower value is better, and $\approxeq0$ indicates that a value closer to $0$ is better.}
    \label{tab:simulation_results}
\end{table}

As the Kriging approach has been mainly designed for static data, it is evaluated on a single time step. As detailed in Table \ref{tab:simulation_results}, Kriging is significantly outperformed by ML-based models across all metrics. \an{Among all models, VUNet achieves the highest SSIM of 0.685, indicating the} model's capability to capture key spatial patterns and preserve the visual structures formed by pollution fields.

Each model has a different optimal window length $k$. From Table \ref{tab:simulation_results}, CLSTM achieves the best performance at $k=8$, whereas VUNet and ViTAE achieve at $k=4$ and $k=6$. The variation in performance can be explained by the differing architecture of each model. CLSTM is specifically designed for sequential data. This architecture enables the output quality to improve steadily with increasing $k$. In contrast, the VUNet model processes temporal data as channel-wise concatenation of input time steps and lacks an explicit mechanism to model sequential dependencies. The ViTAE model, based on a Vision Transformer architecture, contains a large number of neural network parameters (see Table~\ref{tab:model_parameter} in the Supplementary material). The increased model complexity may make optimisation more challenging under the limited training data available (10 months of hourly snapshots) and cause higher errors.

\subsection*{Generative model and the ensemble effect}\label{result_generative}

Details of the diffusion model are provided in the Section Generative model. The stochasticity inherent in diffusion models enables them to output diverse samples. By aggregating multiple predictions, the ensemble methods improve the quality of the output. During output generation, the ensemble size $E$ measures the number of samples to be averaged and the number of inference steps $R$ measures the number of integration steps used to generate each sample prediction. The numerical experiments evaluate the effect of $E$ and $R$ on the performance of the diffusion model.

\begin{figure}[htbp]
    \centering
    \includegraphics[width=\linewidth]{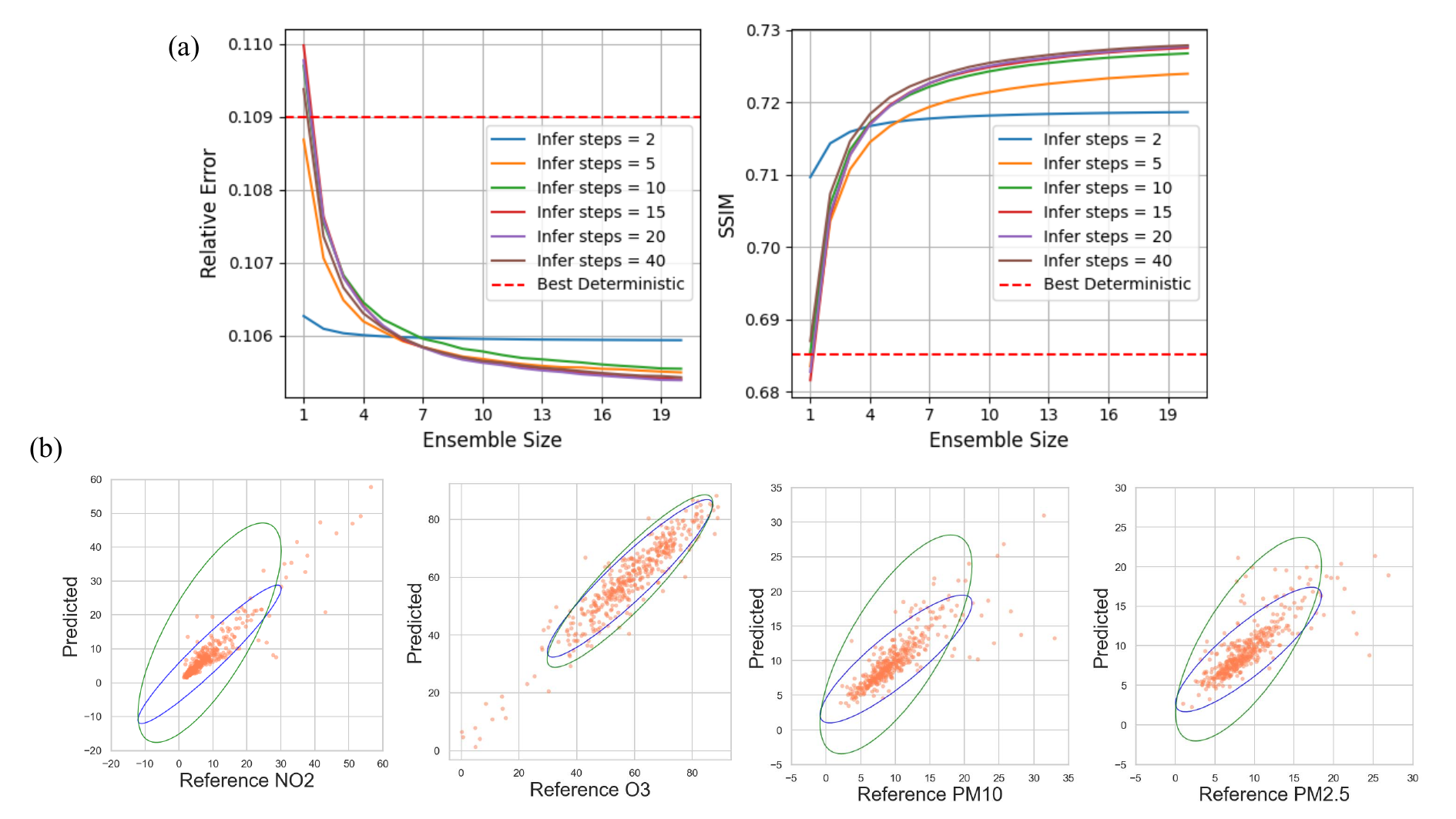}

    \caption{\textbf{The result of the generative model on simulation data.} Panel (a) shows the effect of the parameters Ensemble Size (\textit{E}) and Inference steps (\textit{R}) on the diffusion model on \an{relative L2 error} and SSIM metrics averaged over the four pollutants. The dashed red line in each plot represents the best-performing deterministic model from Table \ref{tab:simulation_results}. Panel (b) plots an \an{covariance} ellipse bounding the points for ground truth versus prediction. The blue ellipse is for the diffusion model ($E=20$) and the green ellipse represents the Kriging model.}
    \label{fig:ensemble_metric}
\end{figure}

Figure \ref{fig:ensemble_metric} plots the evaluation metrics for the average of all pollutants. It shows that the prediction quality improves with increasing ensemble size and inference steps. Even with fewer sampling steps, $R \approx 10$, the model demonstrates strong performance, highlighting the efficiency of DPM Solver++~\cite{lu2025dpm} for sampling. Compared to the best-performing deterministic model from Table \ref{tab:simulation_results}, ensembling allows the diffusion model \an{to achieve lower relative L2 error and higher SSIM of 0.727}. It helps reduce stochastic variation in outputs, but after the variance is sufficiently reduced and the model bias remains, there is no further measurable improvement. This is observed in Figure \ref{fig:ensemble_metric}, as the performance gain plateaus at $E\approx 20$. To reduce the computational inference time without compromising the quality of the result, we choose $E=20$ and $R=10$ for the diffusion model.

Additionally, Figure \ref{fig:ensemble_metric} plots the reference simulation against the predicted values for all grid points across samples and overlays a \an{covariance} ellipse for the diffusion model ($E=20$, $R=10$) and the Kriging baseline. For NO$_2$, PM$_{10}$ and PM$_{2.5}$, the diffusion model has a tighter and more compact ellipse, indicating a reduced dispersion of the prediction error \an{when compared to the Kriging model.}

\begin{figure}[htbp]
    \centering
    \includegraphics[width=0.8\linewidth]{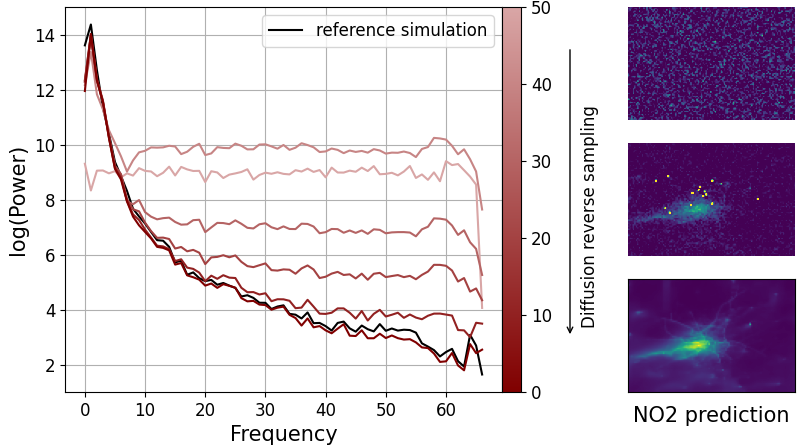}
    \caption{\textbf{Spectral plot of denoising steps of diffusion model.} From pure Gaussian noise to generated output, progressive denoising mechanism in the diffusion model of a single NO$_2$ sample from the simulation dataset. In a $50$-step denoising ($R=50$), the power spectrum plot shows the logarithm of the power values of intermediate steps until the final output and compare it to the reference simulation.}
    \label{fig:psd_diff_sampling}
\end{figure}

Diffusion models exhibit better performance as they progressively denoise the data, allowing a coarse-to-fine generation process~\cite{psd_diffusion_rissanen2022generative}. To quantify this behaviour, we conduct spectral analysis on intermediate outputs during the reverse sampling process (Figure \ref{fig:psd_diff_sampling}). Using a 50-step denoising process ($R=50$), we observe that the generated fields progressively match the reference simulation, with low-frequency components (corresponding to coarse structure) recovered first, followed by higher-frequency values (corresponding to fine details). This result is a strong indicator of the advantage that diffusion models provide. Rather than purely increasing the model size or capacity, diffusion models implicitly perform a more complex process of aligning the spectral structure.

Figure~\ref{fig:psd_pollutants} in Supplementary material compares the power spectral plots for deterministic and generative outputs for each pollutant in the \an{test} dataset of the simulation data. Since NO$_2$ and O$_3$ have shorter lifetimes and greater spatial variability, they exhibit higher power at high frequencies compared to PM$_{10}$ and PM$_{2.5}$. Importantly, the diffusion model is better at capturing the high-frequency structure of NO$_2$ and O$_3$, while the deterministic models perform better for particulate matter. This behaviour contrasts with point-wise metrics (Table~\ref{tab:all_simulation_results} in Supplementary material), where diffusion achieves better performance for PM$_{2.5}$. The discrepancy arises because particulate matter fields are smoother and dominated by low-frequency components, reducing the advantage of diffusion models in reconstructing fine-scale details.

\begin{figure}[htbp]
    \centering
    \includegraphics[width=0.8\linewidth]{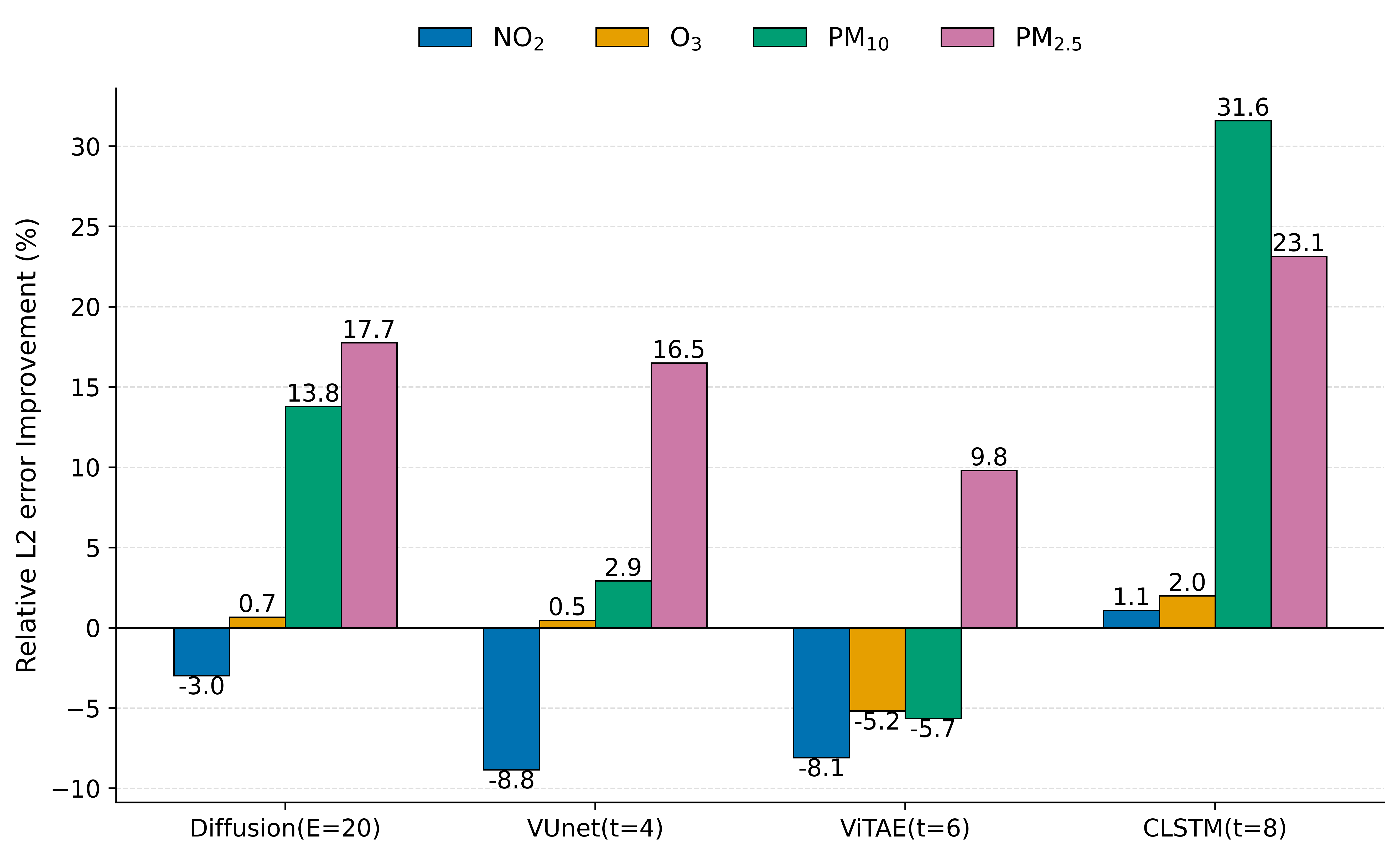}
    \caption{\an{\textbf{Improvement of model due to joint training.} Percentage improvement in relative L2 error by joint training of pollutants compared to single model training. Results are on the test simulation dataset. Bars are coloured blue, yellow, green and pink for NO$_2$, O$_3$, PM$_{10}$, PM$_{2.5}$ respectively. We select only the best performing time window for VUNet, ViTAE and CLSTM, and the optimal ensemble size for the diffusion model.}}
    \label{fig:mre_improvement}
\end{figure}

\an{To assess the benefit of multi-pollutant modelling, we compare the joint pollutant training with independent training for each pollutant on the test dataset. We use the best performing time window for VUNet, ViTAE and CLSTM (Table \ref{tab:simulation_results}) and the optimal ensemble size for the diffusion model. Table~\ref{tab:all_simulation_results} in Supplementary material reports all the metrics for each pollutant and Figure \ref{fig:mre_improvement} shows the percentage improvement in the relative L2 error achieved by joint modelling. The trend is noticeable in diffusion and VUNet models, where the improvement due to joint modelling is more pronounced for the pollutants which have fewer sensors. For PM$_{2.5}$ which has the fewest sensors, joint modelling shows significant improvement across all models, whereas for NO$_2$ with highest active sensors, the individual model often performs better (except CLSTM). NO$_2$ is strongly influenced by traffic whereas the sources of PM$_{2.5}$ are more diverse, potentially making cross pollutant information more beneficial. Thus, joint modelling compensates for limited PM$_{2.5}$ observation data by sharing information across channels. This may provide less benefit for NO$_2$, resulting in its slightly worse performance. \ans{The sensitivity analysis of information exchange between pollutants is conducted in Supplementary subsection "Sensitivity to Cross-Pollutant Information".} Crucially, joint modelling reduces computational and memory requirements since only a single model is required instead of four.}

\subsection*{Validation on real-world dataset} \label{result_realworld}
To validate the performance of trained models under realistic conditions, we investigate the model's robustness using real-world observation data obtained from the Central Air Quality Monitoring Laboratory (LCSQA) through the Geod’air (Management of Air Quality Observation Data) platform~\cite{lscqa}. From the experiments presented in the "Performance of deterministic models on simulation data" subsection and the "Generative model and the ensemble effect" subsection, we select the best-performing hyperparameters for each model and compare their performance on the observation dataset using multiple metrics.

\an{Since the full-field values for real data are unknown, model accuracy is determined with cross validation on test hold-out points using the relative L2 error (see the "Data processing and masking techniques" subsection and the "Evaluation Metrics" Supplementary section). As shown in Figure~\ref{fig:active_sensor} of Supplementary material, the available set of sensors varies considerably between consecutive steps. To evaluate the model's robustness to moving sensors, random stations are removed from each time step of the time window.}

 As detailed in the "Data processing and masking techniques" subsection, we apply augmentations to the simulation dataset to reduce the distribution mismatch with real-world observations, allowing the synthetic data to mimic the characteristics of the real-world dataset. We compare the performance on real-world data across different models with various augmentation methods such as Gaussian, Perlin, Correlated, and Time-aware Gaussian noise (details in "Data augmentation methods" section in Supplementary material).

 \begin{figure}[htbp]
    \centering
    \includegraphics[width=1.0\linewidth]{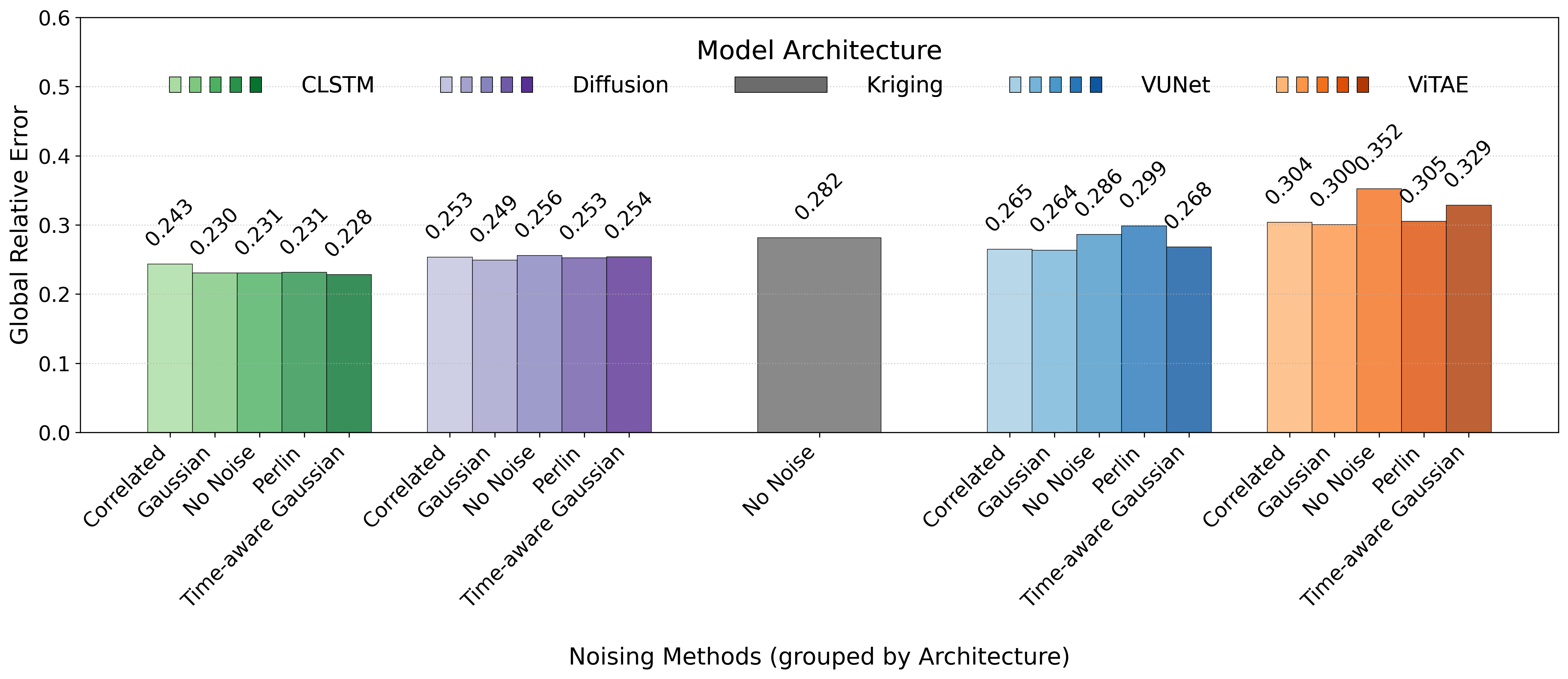}
    \caption{\textbf{Results on real observation dataset with various augmentations.} Comparison of various models with different noising augmentations of (a) Gaussian, (b) Perlin, (c) Correlated, (d) Time-aware Gaussian, and (e) No noise. From Table \ref{tab:simulation_results}, we select the best-performing time step for each model. The \an{relative L2 Error} metric is used on the real observation dataset. Bars are coloured green, purple, grey, blue and orange for CLSTM, diffusion, Kriging, VUNet and ViTAE models respectively. The gradient in the colour indicates different augmentation types as listed in the chart.}
    \label{fig:real-data-results}
\end{figure}

Figure \ref{fig:real-data-results} shows that ViTAE, Kriging, and VUNet achieve \an{relative L2 error} of $0.300$, $0.282$, and $0.264$, respectively, while the diffusion model has a \an{relative L2 error} of $0.249$, and the lowest error is achieved by the CLSTM model with a measure of $0.228$. The introduction of data augmentation during training consistently reduces the error for all models, indicating improved robustness to noise. The largest improvements are observed for ViTAE ($14.77$\%) and VUNet ($7.69$\%), suggesting that these models strongly benefit from exposure to more diverse training data. By contrast, the smaller gains for diffusion ($2.73$\%) and CLSTM ($1.29$\%) indicate that their internal mechanisms of stochastic sampling and temporal modelling, respectively, enable robust performance on noisy out-of-distribution data and thus reduce the reliance on explicit augmentations.

\an{The sensitivity to longer sensor unavailability is provided in the "Sensitivity to prolonged sensor unavailability" section in Supplementary material. Across all models, removing the same station throughout the time window results in slightly higher relative L2 error, likely due to a distribution shift, as the model is trained on dynamically moving sensors and does not explicitly encounter prolonged unavailability of the same sensor. In addition, zone- and pollutant-specific metrics are also included (Figure~\ref{fig:inner_vs_outer_same_window} in Supplementary material). O$_3$ has higher errors than other pollutants, consistent with the strongly left-skewed real-world data distribution (Figure \ref{fig:noise_clean_distribution_comparison}). PM$_{2.5}$ shows higher errors at the outer-city locations where only 3 sensors are available, whereas PM$_{10}$ displays higher errors at inner-city locations due to greater spatial variability in central urban areas.}

\begin{figure}[htbp]
    \centering
    \includegraphics[width=\linewidth]{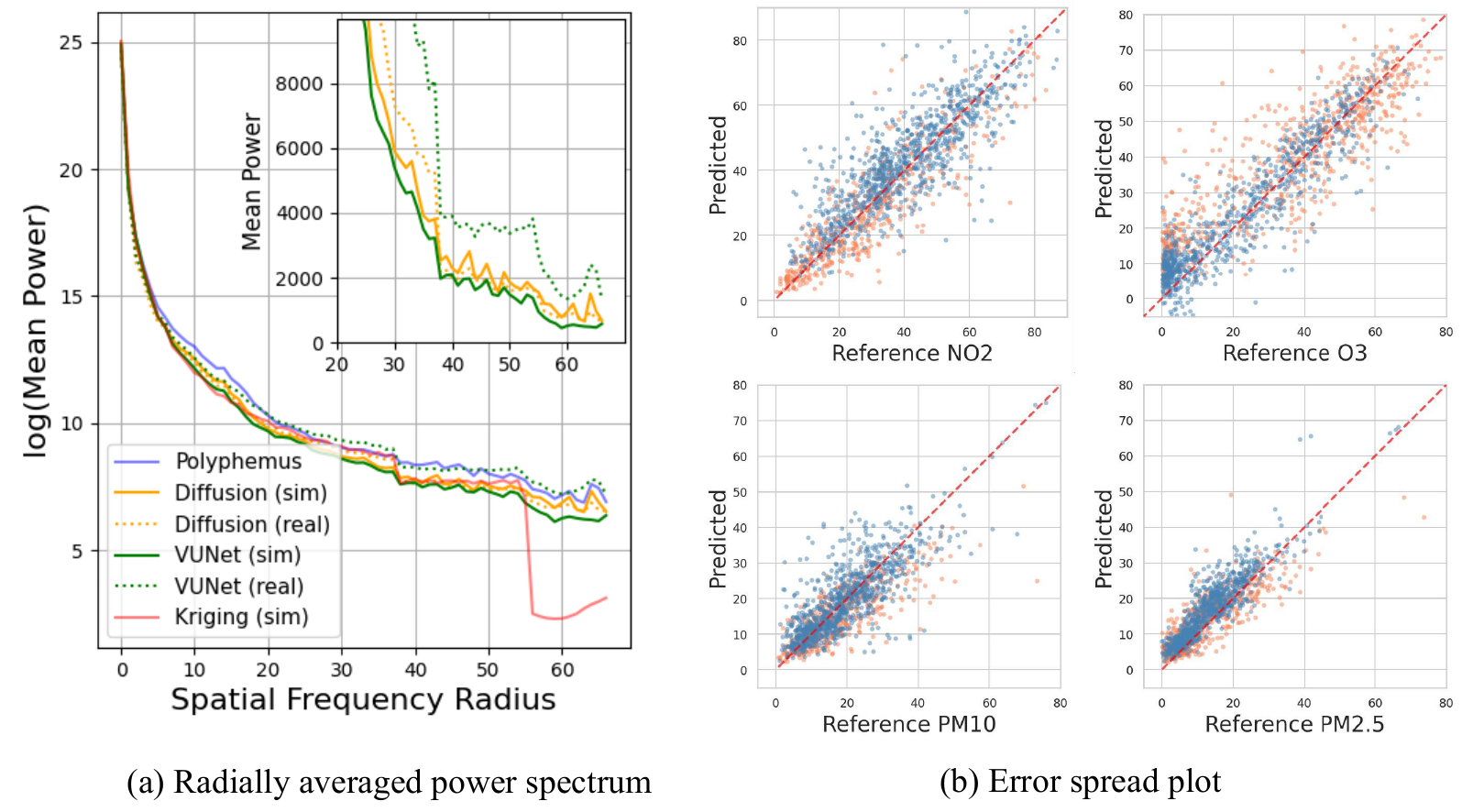}

    \caption{\textbf{Spectral plot and error spread plot on the real dataset.} Panel (a) compares the log of power values of reference simulation data (purple line) and predicted results from the diffusion model ($E=20$) (yellow line), VUNet (green line) and the Kriging model (red line). The solid lines indicate simulation data and the dotted line indicate real data. The values are averaged over all four pollutants. Panel (b) is plotted for the real-world data at the held-out test points. Orange points indicate sensor locations in the outer parts of the city and blue points indicate locations in the inner part of the city. The dashed red line indicates reference for perfect prediction}
    \label{fig:gt_vs_pred}
\end{figure}

While the relative L2 error provides a point-wise accuracy measure, we also compare the structure of predicted pollution fields using the power spectrum plot to analyse their frequency content (Metric details in the "Evaluation Metrics" section in Supplementary material). Figure \ref{fig:gt_vs_pred} a) shows the log mean power, averaged across samples and pollutants as a function of frequency. We plot the reference simulation data (considered as the ground truth when using synthetic observations) against the outputs of the generative and deterministic models evaluated on synthetic and real observations.

On the simulation data (solid colour lines) the diffusion model closely matches the reference spectrum (Polyphemus model) across frequencies. This implies that both large-scale structure and fine-scale details are preserved by the diffusion model. By contrast, the Kriging model decays in power values, reflecting oversmoothing of the output. VUNet (deterministic) relies on a compressed latent representation, which introduces an information bottleneck and leads to a loss of finer details, as reflected by reduced high-frequency components.

On real-world data (dashed lines), the deterministic model deviates from its simulation spectrum and shows increased high-frequency power, suggesting sensitivity to input noise and hallucination of fine textures. By contrast, the diffusion model remains closely aligned with the simulation spectrum, highlighting its greater robustness to noisy observations and its ability to \ans{generate spatial structure consistent with the simulations.}

Additionally, Figure \ref{fig:gt_vs_pred} b) plots the held-out sensor observations against the predicted values of the diffusion model for the real dataset. As described in the "Data processing and masking techniques" subsection, the sensor coverage is denser in inner-city regions than in outer-city locations. \ans{Despite this, a similar error spread of inner- and outer-city points highlights the generalisation ability of the ML models to various spatial locations.}

\begin{figure}[htbp]
    \centering
    \includegraphics[width=0.9\linewidth]{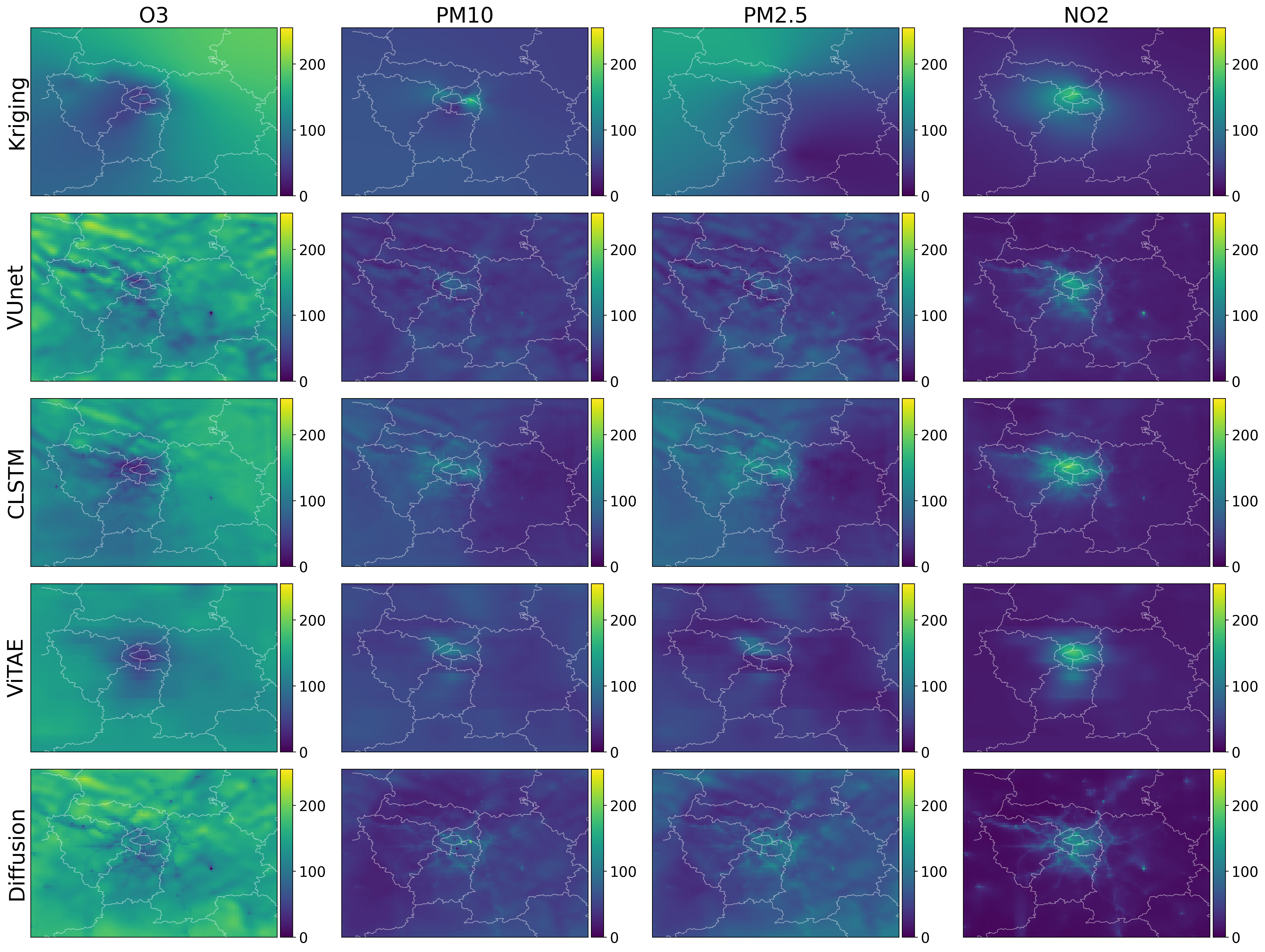}
    \caption{\textbf{Prediction output on real observations.} Visual comparison of (a) Kriging (b) VUNet (with Gaussian noise) (c) CLSTM (with Time-aware Gaussian noise) (d) ViTAE (with Gaussian noise) (e) Ensemble Diffusion (with Gaussian noise) models on O$_3$, PM$_{10}$, PM$_{2.5}$, NO$_2$ pollutants on December 13 12:00:00 2014 }
    \label{fig:real-visual}
\end{figure}

Alongside the quantitative results, we assess the quality of reconstruction through visual comparisons (Figure \ref{fig:real-visual}). From Figure \ref{fig:real-data-results}, we select the best-performing augmentation setting for each model and visualise the results in Figure \ref{fig:real-visual}. Kriging and ViTAE produce distorted fields, whereas VUNet, CLSTM, and diffusion models produce outputs with clear structural patterns. \an{This observation is validated by Figure \ref{fig:real-data-results}, as Kriging and ViTAE have the highest relative L2 error.} The poor performance of ViTAE is likely due to overfitting, as its larger model capacity limits the generalisation to noisy real-world data, while the other ML models demonstrate robustness to noise.

\subsection*{\an{Uncertainty analysis of the Diffusion model}}

\an{We perform uncertainty analysis of the ensemble result of the diffusion model on the held-out stations of the real observation dataset. We report three complementary metrics - absolute error vs spread, rank histogram and CRPS distribution.}

\begin{figure}[htbp]
    \centering
    \includegraphics[width=1.0\linewidth]{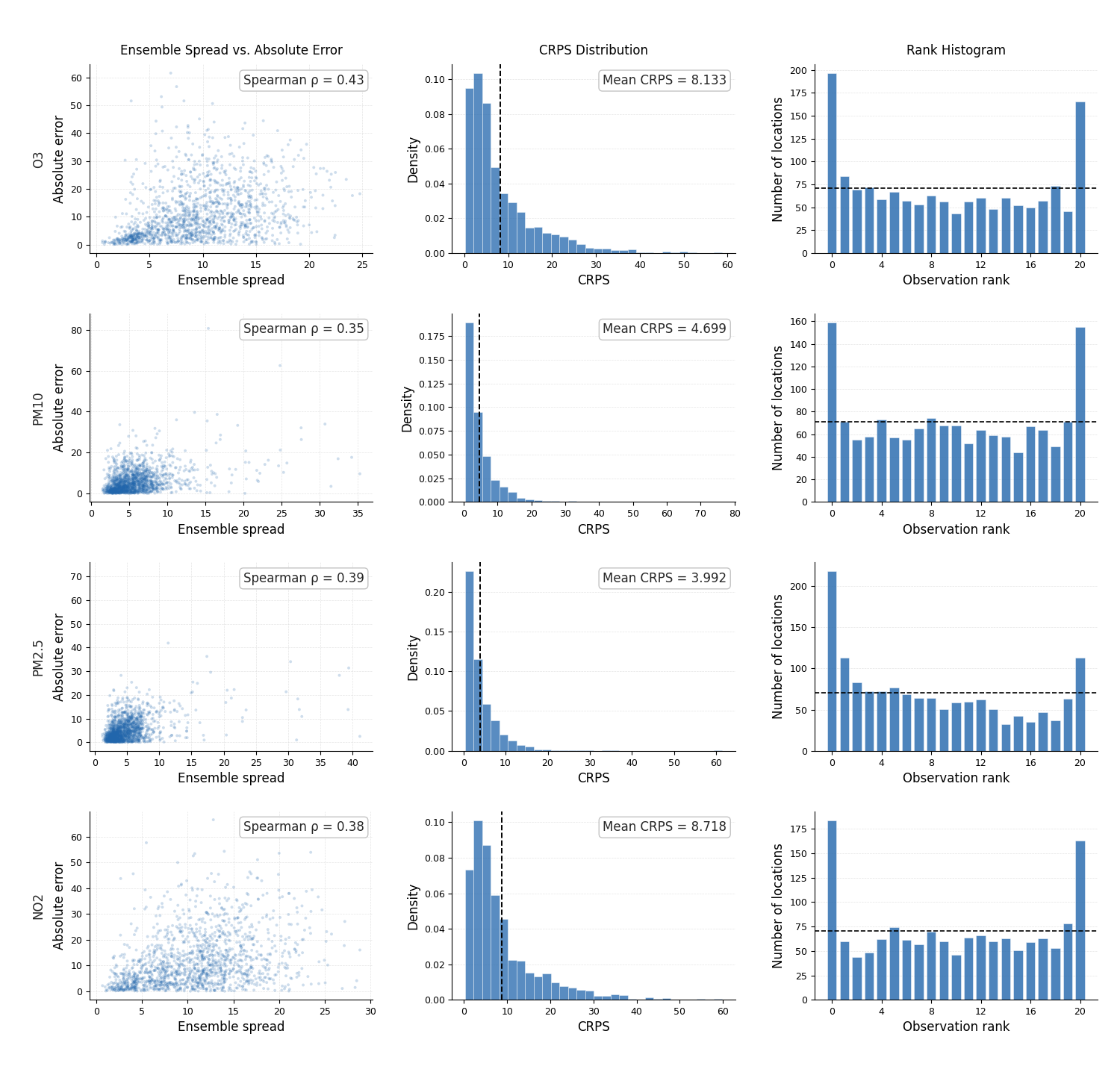}
    \caption{\an{\textbf{Uncertainty analysis of the diffusion model.} Uncertainty analysis on ensemble predictions of the diffusion model at held-out stations of real observation data}}
    \label{fig:uq_diffusion}
\end{figure}

\an{Figure \ref{fig:uq_diffusion} plots these measures for each pollutant. The ensemble spread shows a moderately positive association with absolute error for all pollutants, with the Spearman coefficient ranging between $0.35$ and $0.43$. This indicates that the larger errors can be associated with higher uncertainty. However, the rank histograms have a U-shape which implies that the model is underdispersed and overconfident. Approximately $23\%$ of the observed values fall outside the ensemble range of predicted values for all pollutants (precisely $23.21\%$, $21.13\%$, $22.3\%$, $23.21\%$ for O$_3$, PM$_{10}$, PM$_{2.5}$, NO$_2$ respectively). PM$_{2.5}$ has a higher tendency of overestimation as the histogram is left-skewed.}

\an{We compare these results with the 20-member ensemble prediction from the Kriging model (Figure~\ref{fig:uq_kriging} in Supplementary material). The Kriging model has a Spearman coefficient between $0.05$ and $0.16$, indicating negligible to weak positive correlation between mean error and ensemble spread. The rank histogram is strongly U-shaped, where over $60\%$ of the observed values fall outside the predicted ensemble range. Although Kriging achieves slightly lower CRPS values for some pollutants, complementary metrics indicate that the diffusion model provides better \ans{ensemble variability}. The slightly higher CRPS of the diffusion model can be attributed to its larger predictive spread, which improves coverage but can increase the CRPS.}

\an{Deterministic neural network models can also generate ensemble predictions by introducing variations in the weight initialisations. As discussed in~\cite{kendall2017uncertainties}, a deterministic model trained on MSE loss to produce a single output primarily approximates the conditional mean rather than the conditional distribution. Consequently, ensemble constructed by these methods do not provide a calibrated representation of the predictive distribution. This highlights the advantage of diffusion models which explicitly aim to learn the conditional distribution of the target field. The experiment details are provided in the "Uncertainty analysis of Kriging and VUNet" subsection in Supplementary material to further support this conclusion.}

\an{The diffusion model provides a direct method for representing variability because the stochastic sampling is incorporated into the generation process. The uncertainty estimates of the diffusion model are informative but require further calibration. \ans{Table~\ref{tab:coverage} in Supplementary material confirms the underdispersion in the diffusion model.} Potential sources of this include ineffective sampling, point-wise loss functions, less variability in the training dataset or strong conditioning. Further investigation is required to identify the underlying cause. Previous studies have reported calibrating rank histograms through various techniques such as utilising a CRPS loss function~\cite{lang2026aifs} or using a noise scaling coefficient for the diffusion model~\cite{couairon2026archesweathergen}.}

\section*{Discussion}
\label{discusion}
This study demonstrates the effectiveness of machine learning models for the reconstruction of pollution fields. Despite being trained for a relatively short period of 10 months, the ML-based models outperform the Kriging baseline and produce realistic reconstructions. The maximum number of active sensors (when no station is under maintenance) for each pollutant is 28, 24, 15, 9, respectively, for NO$_2$, O$_3$, PM$_{10}$, and PM$_{2.5}$, which corresponds to a sparsity of $99.66$\%, $99.71$\%, $99.82$\%, and $99.89$\% against the number of grid points in the field. Our method highlights the ability of ML models to generate full fields while observing highly sparse data.

Integration of previous time steps into deterministic models (Table \ref{tab:simulation_results}) reveals the influence of historical data, as accuracy improves. In general, increasing the size of the time window leads to improved performance with diminishing returns after a certain length.  The optimal window length is likely influenced by the variability of weather patterns in the reconstructed region. Under highly volatile weather conditions, long temporal windows may encompass multiple distinct weather events, some of which may be unrelated to the target period. Although deterministic models perform better than Kriging, they are limited by their formulation, as they are trained to predict a single outcome. \an{The uncertainty analysis highlights that a deterministic model trained on MSE loss is optimised to estimate the conditional mean and does not learn the conditional distribution.} This one-to-one mapping restricts their ability to be deployed in complex physical environments where uncertainty quantification and robustness to variability are required.

We construct one of the first spatially-aware generative AI models for the reconstruction of air pollution. The diffusion model provides two unique advantages over deterministic models. First, it models the distribution of the pollution fields rather than predicting a single output. This enables uncertainty exploration by allowing us to sample multiple outputs from the predicted distribution. Due to an increase in \textit{extreme pollution} events, these fields often exhibit anomalies and highly dynamic spatio-temporal behaviour~\cite{bashan2024dynamics}. Diffusion models help capture the uncertainty that is usually observed in atmospheric modelling.

\an{Second, diffusion models exhibit a frequency-structured reconstruction which allows them to \ans{generate spatial structures that are consistent with the simulated pollution fields.} Although the simulation data is validated with real observation fields~\cite{lugon2022}, the spectral agreement is a measure of plausibility of realistic reconstruction rather than independent validation of real-world data.} For pollutants with shorter atmospheric lifetimes (NO$_2$, O$_3$), we observe pollution fields that are sharper with high variability. The gradual denoising process of the diffusion model facilitates the recovery of fine-scale details (Figure \ref{fig:psd_diff_sampling}), consistent with our analysis in the spectral domain. More broadly, the results suggest that sparse-monitor reconstruction should not be viewed only as an interpolation problem, but as a probabilistic inference problem constrained by atmospheric structure.

\an{Although the diffusion model outperforms the deterministic models on the simulation dataset, its performance varies for real-world observations. On the held-out real-world test points, CLSTM has the least relative L2 error, followed by the diffusion model. Since CLSTM uses historical observations as inputs, whereas the diffusion model does not, this comparison highlights the potential benefit of integrating temporal information into the model. Our uncertainty analysis also indicates that the diffusion model is underdispersed. Given the high variability of pollution fields, improving uncertainty calibration represents an important direction for future work. }

\an{This study provides findings that advance the application of ML techniques in environmental science where data is usually sparse, noisy and highly variable. These results provide a foundation for the development of real-world air pollution reconstruction.} First, ML algorithms show a remarkable ability to leverage inter-pollutant relationships under extremely sparse monitoring conditions to improve output quality. Despite the relatively low monitoring coverage for PM$_{10}$ and PM$_{2.5}$, we achieve high reconstruction quality. Second, the transition to real-world observations is an important aspect for assessing the model's generalisation capability. By incorporating noise-based augmentations during training, these models can be deployed in real-world environments without requiring \an{supervised fine-tuning}. This represents an important result, offering a cost-effective and scalable solution for reconstruction tasks while enabling the extension of this research to other weather and climate modelling applications. In particular, the construction of precise air quality concentration fields is of significant importance for analysing the spatial relationships between healthcare outcomes and air quality data.

The main limitations of the current study are its dependence on training with simulation data and its limited generalisability to different cities. The model has to be retrained to adapt to a particular region. The limited availability of simulation data constrains the use of high-capacity ML models. To address this limitation, future research could incorporate auxiliary data sources—such as topographic information, emission inventories, climate variables, and satellite observations—as model inputs or conditioning variables for generative models. By leveraging the ability of ML models to process multimodal data, the interdependencies among these factors could provide a more detailed analysis. The effect of previous time step data on the reconstruction should be studied in more detail, especially for generative models.

\section*{Methods}\label{method}

\subsection*{Air-quality simulations}
\label{method_airquality}

The air quality of the Paris metropolitan region is simulated using the three-dimensional Eulerian chemistry–transport model Polyphemus/Polair3D \citep{sartelet2007} with the configuration described in \citet{lugon2022}. The model accounts for the main atmospheric processes, including emissions from anthropogenic and biogenic sources, transport (advection and diffusion), chemical transformations, and removal by dry and wet deposition. The formation of secondary gases and aerosols is represented through coupling with the SSH-aerosol module \citep{sartelet2020}. As detailed in \citet{lugon2022}, the simulations are performed over a one-year period in 2014 on a 2 km × 2 km horizontal grid with 14 vertical levels. The initial and boundary conditions are prescribed from larger-scale Polair3D simulations over Europe and metropolitan France, following the setup described in \citet{andre2020}.

\subsection*{Data processing and masking techniques} \label{method_dataprocess}

This study uses two different datasets for training and evaluating the models. The \textit{simulation} dataset of all four pollutants (NO$_2$, O$_3$, PM$_{2.5}$, PM$_{10}$) is obtained as described in the "Air-quality simulations" subsection. \an{To ensure a fair training process, this dataset is split into training, validation and test sets. This split is based on calendar months with November and December representing validation and test sets respectively. The ML models are trained on data from January to October.} To evaluate real-world performance, we use the \textit{observation} dataset obtained from the Central Air Quality Monitoring Laboratory (LCSQA) through the Geod’air (Management of Air Quality Observation Data) platform. This dataset contains hourly recordings of the concentrations of the four regulated pollutants mentioned above in the Paris metro area from January 1 to December 30, 2014.

\begin{figure}[htbp]
    \centering
    \includegraphics[width=1\linewidth]{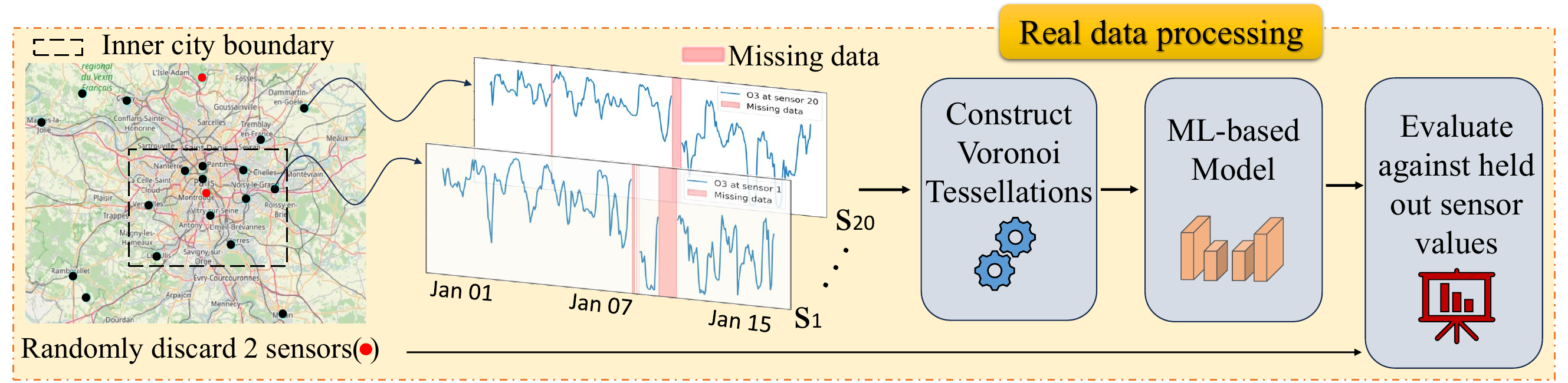}
    \caption{\textbf{Flow chart of real data processing.} Figure illustrates the processing of real-world data. First, randomly withhold one active sensor inside and one active sensor outside the city boundary. Next, record observation values from station number $S_1 \dots S_{20}$. Then, construct Voronoi tessellations for each time step. Later, reconstruct pollution fields with the trained ML model and finally evaluate model performance against held-out sensor points. The map of Paris is reproduced from Google Maps.}
    \label{fig:real_data_process}
\end{figure}

For all pollutants, \an{the Paris city is discretised on 2km × 2km map which becomes a} grid with a resolution of $n_x = 75$ and $n_y = 110$ pixels. \an{The latitude-longitude of every sensor location is mapped to the nearest grid point.} We define the binary mask $\mathbf{\Omega}_t \in \{ 0, 1\}^{4\times n_x \times n_y}$, where 1 indicates the presence of a monitoring sensor at time $t$. Let $\mathbf{x}_t \in \mathbb{R}^{4\times n_x \times n_y}$ denote the true concentration fields at time $t$, which are inaccessible in real-world applications. We define the tessellated fields $\mathbf{z}_t = f^v(\mathbf{\Omega}_t, \mathbf{x}_t ) \in \mathbb{R}^{4\times n_x \times n_y}$ where $f^v$ is the two-dimensional tessellation function. We concatenate the four pollutants to facilitate multi-pollutant analysis (as illustrated in Figure \ref{fig:train_flow}).

\an{During reconstruction, each pollutant channel is scaled independently using training-set statistics to account for differences in concentration ranges. For the deterministic models (VUNet, ViTAE and CLSTM), we apply min-max normalisation, whereas the diffusion model uses standardisation based on training mean and variance. While the simulation data generated by Polyphemus is clean, the observation data contains missing values when sensors are unavailable. The sensors which have negative or missing data are excluded from the Voronoi tessellation formulation. The proposed framework adapts to the missing data without requiring any additional processing.}

\an{At any given time the sensor which records a positive value is defined as an active sensor.} We develop a cross-validation masking technique that randomly selects active sensor locations, as shown in Figure \ref{fig:real_data_process}. At each time step, one sensor from the inner-city and one from the outer-city regions are classified as evaluation locations: $\mathbf{\Omega}_t^{\textrm{eval}}$. During training, these points are removed from the model's input mask: $\mathbf{\Omega}_t^{\textrm{train}} = \mathbf{\Omega}_t \odot (1-\mathbf{\Omega}_t^{\textrm{eval}})$. \an{For models that take previous time step data as input, random sensor points are removed from each of the previous time steps to encourage the model to learn inconsistent spatial patterns.} The inner-city boundary is defined by the area of the administrative city of Paris. This distinction helps to evaluate the model's performance in both densely and sparsely monitored regions. The variation in masking encourages the models to learn different configurations that could be used to obtain richer reconstructions for a sparsely observed field.

\begin{figure}[htbp]
    \centering
    \includegraphics[width=1\linewidth]{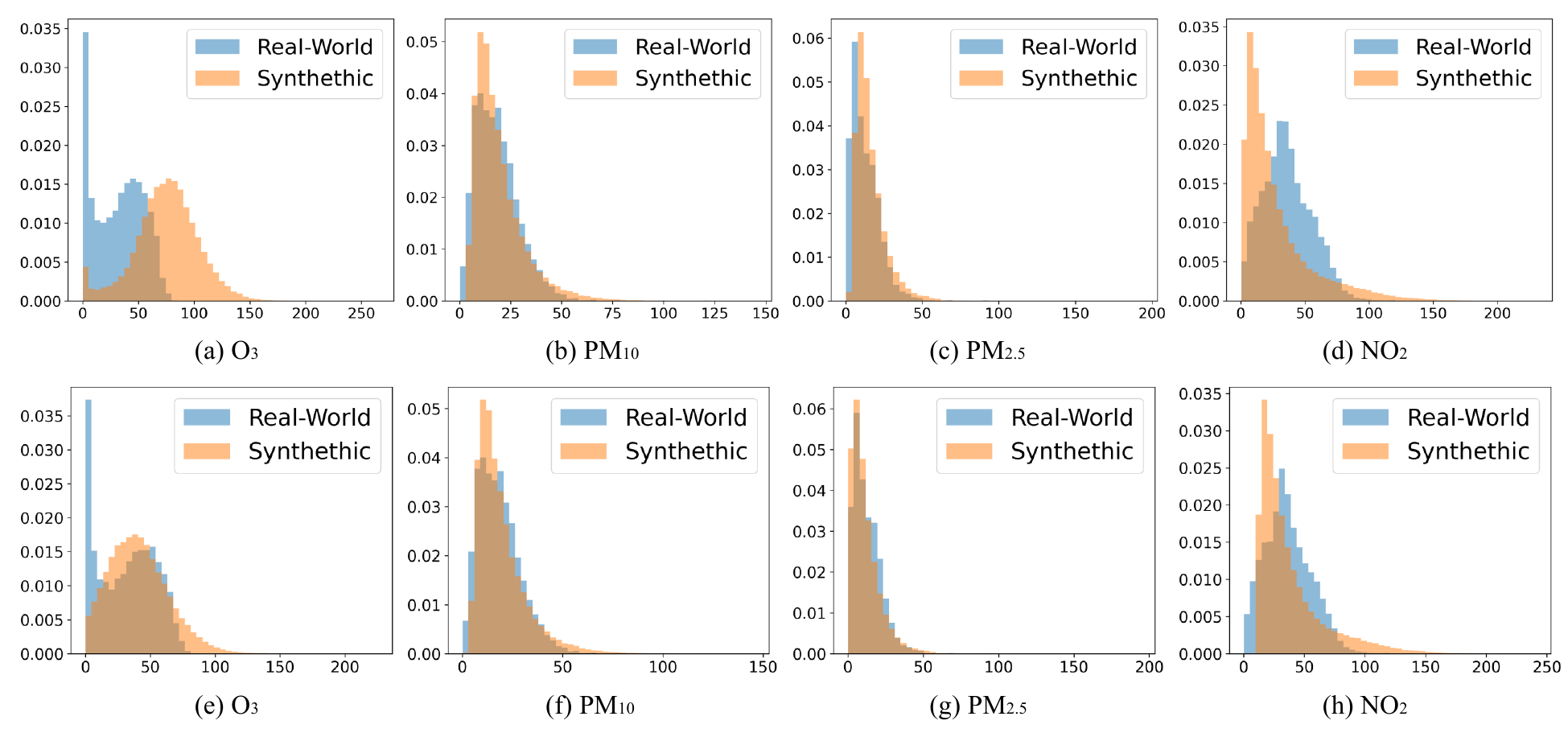}
    \caption{\textbf{Data distribution before and after augmentations.} Comparison of the distribution of the masked field for each pollutant between real and synthetic data. Figures (a) to (d) are before augmentations and (e) to (h) are the transformed distributions after augmentation. The blue histogram indicates real-world data, and the orange histogram indicates synthetic data.}
    \label{fig:noise_clean_distribution_comparison}
\end{figure}

We apply data augmentation to simulation data during training to diversify the model's input and reduce the distribution gap between observed and simulated data. Figure \ref{fig:noise_clean_distribution_comparison} panels (a) to (d) compares the distribution of simulation and real data before augmentations and (e) to (h) highlights the shift after augmentations. \an{During training, the perturbations are applied to the input and not the target data.} Experiments are performed using various perturbation techniques such as Gaussian, Time-aware Gaussian, Perlin, and Cross-correlation noise (details can be found in the "Data augmentation methods" section in Supplementary material). Each method is parameterised by a mean and standard deviation that determine the magnitude of the applied perturbation. \an{The perturbations are applied only on the training dataset and not the validation or test dataset.} To better align the pollutant-specific distributions, the mean is determined by $\mu_{\textrm{noise}} = \mu_{\textrm{obs}} - \mu_{\textrm{sim}}$, where $\mu_{\textrm{obs}},  \mu_{\textrm{sim}}$ denote the average observation and simulation values, respectively, computed across all sensors and time steps used in the \an{training dataset}. The standard deviation controls the strength of the perturbation and is selected through a randomised search on the VUNet model \an{using the metrics on the validation set}.

\subsection*{Deterministic models}\label{method_model_determ}
In this section, we discuss in detail the deterministic ML models employed in this study, namely VUNet, ViTAE, and CLSTM

\textbf{VUNet} (short for Voronoi UNet) is inspired by Fukami et al.~\cite{RefWorks:RefID:4-fukami2021global}. We adopt the UNet architecture, consisting of 4 downsampling and upsampling steps along with self-attention injected into the bottom layer. The input is a concatenation ($\oplus$ operator) of Voronoi tessellation and its corresponding masks ($\mathbf{z}_t$ and $\mathbf{\Omega}_t$). To observe the effect of historical data, previous time step inputs are concatenated too. The concatenation is performed by stacking along the channel dimension:
\an{
\begin{equation}
    \mathbf{y}_{t} = f^{\textrm{UNET}}( ( \mathbf{z}_{t-k+1} \oplus ( \mathbf{x}_{t-k+1} \odot \mathbf{\Omega}_{t-k+1})) \oplus \dots \oplus (\mathbf{z}_{t} \oplus (\mathbf{x}_t \odot \mathbf{\Omega}_{t}) )),
\end{equation}
}
where $\odot$ is the Schur product  and $\mathbf{y}_{t}$ denotes the output of the neural network, which is an estimation of $\mathbf{x}_{t}$.
This is a simple model design that compresses the data into a latent space, which encourages the model to learn the most important features.

Extending the formulation of Fan et al.~\cite{RefWorks:RefID:3-fan2025vitae-sl:}, we modify \textbf{ViTAE} by replacing its original decoder (CNN) with a UNet model. The encoder is a Vision Transformer that divides the input image into patches, flattens them, and linearly projects them into a fixed-dimensional embedding space. The resulting embeddings are then rearranged into a multi-channel 2D format and passed through a UNet decoder. In the present work, the ViTAE model receives only sparse observations as input, similar to Fan et al.~\cite{RefWorks:RefID:3-fan2025vitae-sl:},

\an{
\begin{equation}
    \mathbf{y}_{t} = f^{\textrm{ViTAE}}( (\mathbf{x}_{t-k+1} \odot \mathbf{\Omega}_{t-k+1} )\oplus \dots \oplus (\mathbf{x}_t \odot \mathbf{\Omega}_{t} )).
\end{equation}
}

 Unlike VUNet, which operates only on the local neighbourhood regions, ViTAE is capable of establishing long-distance interactions and finding meaningful patterns.

Originally proposed by Shi et al.~\cite{shi2015convolutional} for precipitation nowcasting, the Convolutional LSTM extends the traditional LSTM~\cite{lstm} by integrating convolutional operations into its temporal architecture. This hybrid design allows the model to capture the spatio-temporal dependencies inherent in air quality data. This study extends this implementation by building on the improved Peephole LSTM architecture~\cite{peehole_lstm}. This model receives the previous $k$ time steps' Voronoi tessellations as input to estimate the concentration fields,

\an{
\begin{equation}
    \mathbf{y}_{t} = f^{\textrm{CLSTM}} (\mathbf{z}_{t-k+1}, \dots, \mathbf{z}_t).
\end{equation}
}

All deterministic models are trained using the mean squared error (MSE) between the model output $\mathbf{y}_t$ and the reference simulations $\mathbf{x}_t$,
\begin{equation}
L_{\text{MSE}} = \frac{1}{4 \times n_x \times n_y \times N_{\text{train}}} \sum_{t=1}^{N_{\text{train}}} \left\| \mathbf{y}_t - \mathbf{x}_t \right\|^2,
\end{equation}
where $N_{\text{train}}$ denotes the number of training samples, corresponding to all hourly data from January through October 2014.

\subsection*{Generative model}\label{method_model_generat}

Conditional diffusion models learn the distribution of the dataset rather than learning a one-to-one mapping. The output is sampled from a learned distribution, which allows the model to generate different outputs for the same inputs. These models are conditioned on sparse observations for guided and controllable generation. We work with a class of diffusion models based on the \textit{score-based stochastic differential equation}~\cite{song2020score} and use the Elucidating Diffusion Model (EDM) framework~\cite{karras2022elucidating} for its implementation.
Instead of directly optimising the MSE loss as in deterministic models, diffusion models aim to generate realistic samples that are consistent with the input conditions by progressively denoising (through back sampling) randomly generated Gaussian noise.
For back sampling, which is a technique that solves reverse SDEs, we use DPM Solver++~\cite{lu2025dpm} for faster and high-quality sample generation.

\begin{figure}[htpb]
    \centering
    \includegraphics[width=\linewidth]{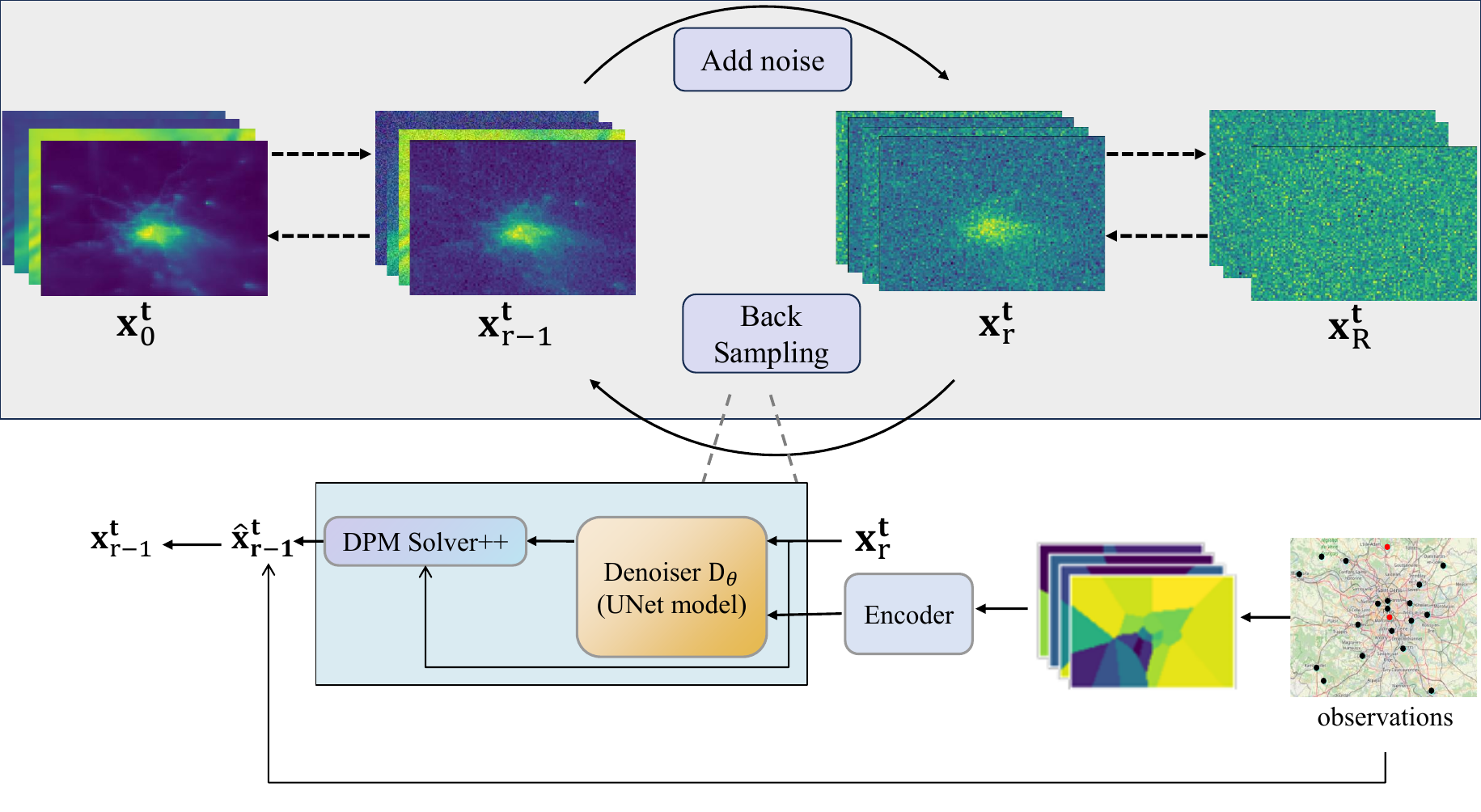}
    \caption{\textbf{Diffusion process of NO$_2$ pollutant showing the forward and reverse processes.} Figure shows adding of Gaussian noise and the mechanism of denoising to predict the output. The map of Paris is reproduced from Google Maps.}
    \label{fig:diffusion}
\end{figure}

The diffusion models operate in a pseudo time space defined by $r \in [0,R]$ such that $\{\mathbf{x}^t_r\}_{r=0}^R$ represents the noise added to the original data at time $r$ with $\mathbf{x}^t_0 = \mathbf{x}^t$ denoting the original data and $\mathbf{x}^t_R$ as pure Gaussian noise. The denoiser model $D_{\theta}$ is trained to predict the \textit{score function} (derivative of log probability), which is used in the sampling process (details in the "Diffusion-based generative model" section in Supplementary material).

The denoiser ($D_{\theta}$) is a UNet model conditioned on sparse observations. These are represented as Voronoi tessellations and projected into latent space using an encoder. These embeddings are injected into UNet layers through cross-attention. The Voronoi embedding is defined as
\begin{equation}
    \mathbf{z}^t_{\textrm{emb}} = f^{\textrm{EMB}} (\mathbf{z}^t),
\end{equation}
 where $f^{\textrm{EMB}}$ is a Transformer encoder model.

As shown in Figure \ref{fig:diffusion}, we start from Gaussian noise at diffusion step $R$ and iteratively denoise using DPM-Solver++ until the final reconstruction is obtained,

\begin{equation}
    \hat{\mathbf{x}}^t_{r-1} = \textrm{DPMSolver}(D_{\theta}(\mathbf{x}^t_r, \mathbf{z}^t_{\textrm{emb}};r)) \quad \textrm{with} \quad \hat{\mathbf{x}}^t_R = \mathbf{x}^t_R, \quad \mathbf{y}^t = \hat{\mathbf{x}}^t_0.
\end{equation}

Although conditioning the denoiser provides consistency with the sparse observations, the generated samples may deviate from the observed values during iterative denoising. To prevent this drift, we employ masked back-sampling, which explicitly enforces the observed measurements at every reverse diffusion step. Rather than inserting the clean observations directly into the noisy samples, the observations are first re-noised to match the current noise level,

\begin{equation}
    \mathbf{\tilde{x}}_{r-1}^t = \mathbf{x}_0^t + \sigma_{r-1}\mathbf{\epsilon}, \quad \mathbf{\epsilon} \sim \mathcal{N}(0, \mathbf{I})
\end{equation}

and are then reinserted into the generated sample at the observed locations

\begin{equation}
    \mathbf{x}^t_{r-1} = \hat{\mathbf{x}}^t_{r-1} \odot (1 - \mathbf{\Omega}_t) + \mathbf{\tilde{x}}_{r-1}^t \odot \mathbf{\Omega}_t
\end{equation}

This update not only ensures that the reconstructed field matches the observations at sensor locations, but also guides the generation of neighbouring regions, as the corrected sample is subsequently passed through the denoiser in the next reverse diffusion step.

For multiple random noise initialisations $\{\mathbf{x}^{t,(e)}_{R}\}$ with $e=1,...,E$, an ensemble of output samples is produced $\{\mathbf{y}^{t,(e)}\}$ and is aggregated to produce the average reconstruction,

\begin{equation}
    \mathbf{y}^t_{en} = \frac{1}{E} \sum_{e=1}^{E} \mathbf{y}^{t,(e)}.
\end{equation}

Ensemble predictions can also be used to estimate predictive uncertainty, which is particularly valuable for data assimilation and risk assessment, as shown in recent studies~\cite{zhuang2025spatially,leinonen2023latent}.

\section*{Data availability}\label{sec5}

The simulation data generated and analysed during the current study are available at \url{https://huggingface.co/datasets/potatotopatoo/Air_pollution_simulation}. The real-world observation data were obtained from the Geod'air platform operated by LCSQA. These data are open-access at \url{https://www.geodair.fr/donnees/consultation} and are subject to the terms of use of the data provider.

\section*{Code availability}
The custom code used to train and evaluate the models and reproduce the results is available at \url{https://github.com/Miha5092/Air-Quality-Field-Reconstruction}.

\section*{Acknowledgements}\label{sec7}
Not applicable

\section*{Funding}
A.Sabnis, M.Bocquet and S.Cheng acknowledge the support of the French Agence Nationale de la Recherche (ANR) under grant reference ANR-25-CE56-0198-01. This work also benefited from State funding managed by ANR under the France 2030 program, reference ANR-23-IACL-0005.
This work has received funding from the PEPR Villes Durables (French
National Research Agency, France, URBHEALTH project ANR-24-PVD0007).
CEREA is a member of Institut Pierre-Simon Laplace (IPSL).

\section*{Author contributions}\label{sec8}

AS and MM performed the formal data analysis and implemented the machine learning models. LL and KS carried out the physics-based model simulations to generate the training data. AS, MM, and SC prepared the manuscript with contributions from all co-authors. MB and SC provided the financial support for the project that led to this publication. SC coordinated the research activities. XC and SZ provided technical support. All co-authors reviewed and edited the manuscript.

\section*{Competing Interests}\label{sec9}

The authors declare no competing financial or non-financial interests. The other authors do not have a competing interest

\bibliography{references}%

\clearpage
\phantomsection
\pdfbookmark[0]{Supplementary material}{supplementary-material}
\begin{center}
  {\Large\bfseries Supplementary material\par}
  \medskip
  {\large From Deterministic to Generative Deep Learning for Urban Air Quality
  Reconstruction from Sparse Observations\par}
\end{center}
\bigskip
\setcounter{section}{0}
\setcounter{subsection}{0}
\setcounter{figure}{0}
\setcounter{table}{0}
\setcounter{equation}{0}
\renewcommand{\thesection}{S\arabic{section}}
\renewcommand{\thefigure}{S\arabic{figure}}
\renewcommand{\thetable}{S\arabic{table}}
\renewcommand{\theequation}{S\arabic{equation}}
\renewcommand{\theHsection}{supp.\arabic{section}}
\renewcommand{\theHsubsection}{\theHsection.\arabic{subsection}}
\renewcommand{\theHsubsubsection}{\theHsubsection.\arabic{subsubsection}}
\renewcommand{\theHfigure}{supp.\arabic{figure}}
\renewcommand{\theHtable}{supp.\arabic{table}}
\renewcommand{\theHequation}{supp.\arabic{equation}}

\section{Notation Table}\label{appendix_notation}
\vspace{-5mm}
\begin{table}[htbp]
\centering
\begin{adjustbox}{max width=\linewidth,center}
\begin{tabular}{ll}
\hline
\textbf{Notation} & \textbf{Description} \\[0.5ex]
\hline
$\mathbf{s}^d$ & Set of $n$ distinct locations of sensor points for pollutant $d$ \\
$\mathbf{z}_t$ & Voronoi tessellation at time $t$ \\
$\mathbf{\Omega}_t$ & Binary mask of sensor locations \\
$\mathbf{\Omega}_t^{train}$ & Binary mask used for training model \\
$\mathbf{\Omega}_t^{eval}$ & Binary mask used for evaluating model \\
$\mathbf{x}_t$ & Ground truth of pollutant fields \\
$\mathbf{y}_t$ & Predicted reconstruction from models \\
$R$ & Number of back sampling steps in the diffusion model \\
$E$ & Ensemble size in diffusion model \\
$\mathbf{x}_r^t$ & Noise added to ground truth image at diffusion time $r$ and pollutant time $t$ \\
$\mathbf{y}_{en}^t$ & Ensemble prediction of diffusion model \\

\hline
\end{tabular}
\end{adjustbox}
\caption{Notation Table}
\label{tab:notation}
\end{table}

\section{\an{Sensor placement}}\label{appendix:sensor_analysis}

\begin{figure}[htbp]
    \centering
    \includegraphics[width=\linewidth]{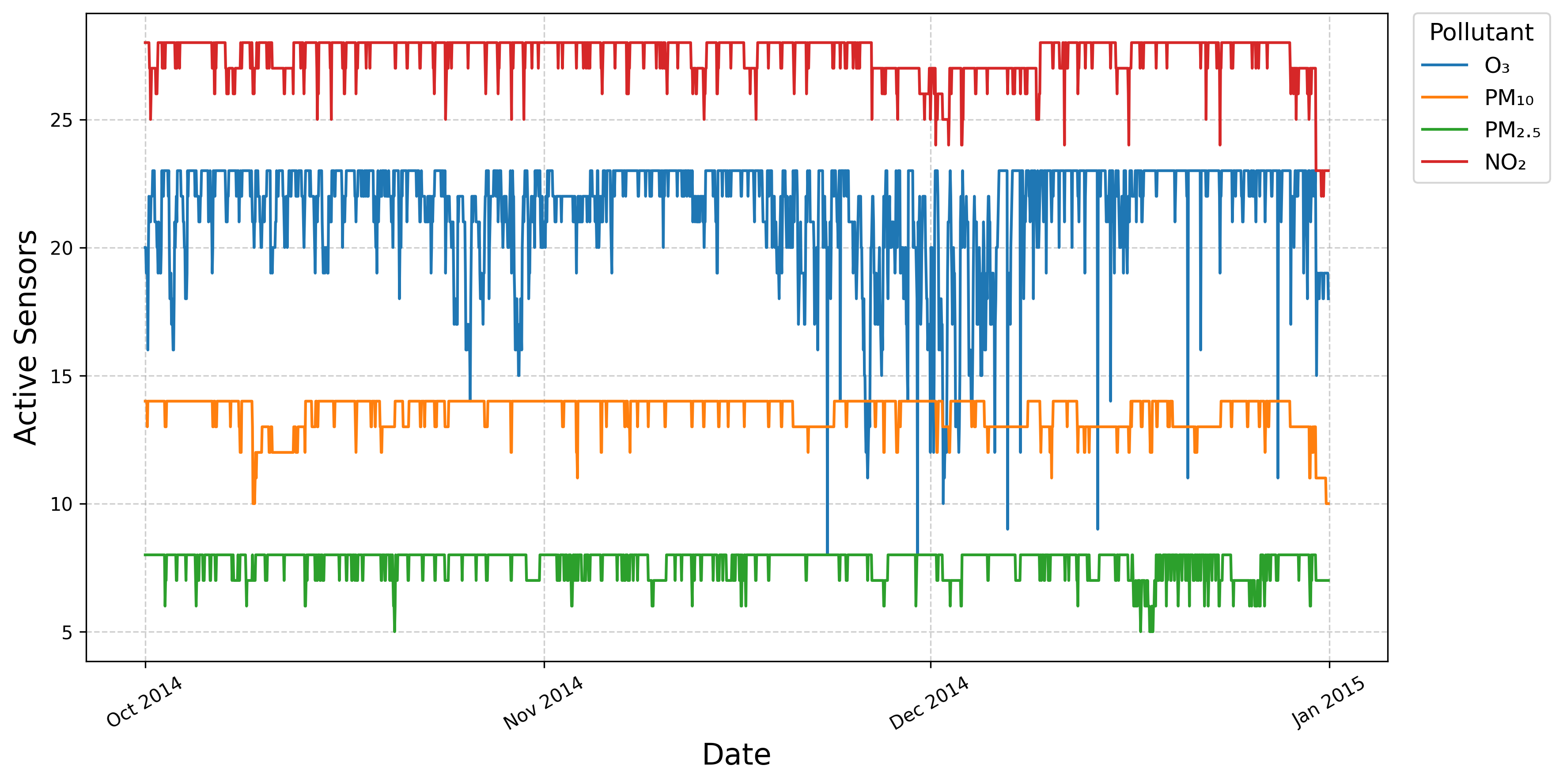}
    \caption{\an{\textbf{Variation of active sensors per pollutant.} Number of active sensors of each pollutant in the real-world observation dataset for the months of October (part of the training set), November (validation set) and December(test set). The blue, orange, green and red lines are for  O$_3$, PM$_{10}$, PM$_{2.5}$, NO$_2$ respectively}}
    \label{fig:active_sensor}
\end{figure}

\an{Figure \ref{fig:active_sensor} plots the number of active sensors for each pollutant at hourly time step for the months of October, November and December. The number of active sensors varies rapidly and they reach their maximum value frequently, suggesting that a station does not remain unmonitored for a prolonged time period. In addition, the set of available sensors can change substantially between consecutive time steps.}

\section{\an{Sensitivity to prolonged sensor unavailability}}\label{real_data_analysis}

\an{In this section we analyse the reconstruction quality on treal-world dataset by measuring the relative L2 error on the test holdout point. During model training, we randomly remove the sensors from previous time steps to improve robustness to constantly changing sensor availability as indicated by Figure \ref{fig:active_sensor}. In this experiment, when including the historical information, we compare two settings - removing the same station from all previous time steps versus randomly removing stations at each time step. }

\begin{figure}[htbp]
    \centering
    \begin{subfigure}[b]{0.33\linewidth}
        \includegraphics[width=\linewidth]{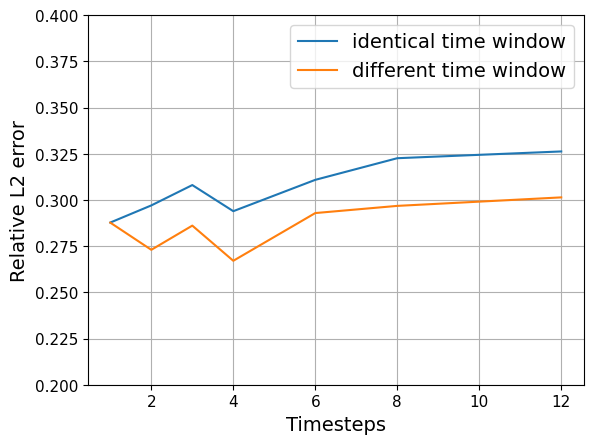}
        \caption{VUNet }
    \end{subfigure}\hfill
    \begin{subfigure}[b]{0.33\linewidth}
        \includegraphics[width=\linewidth]{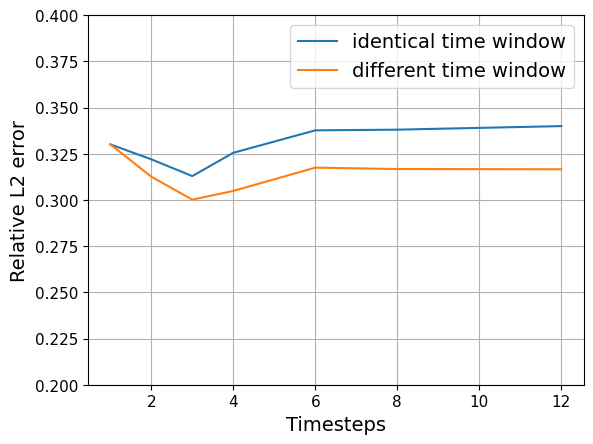}
        \caption{ViTAE}
    \end{subfigure}\hfill
    \begin{subfigure}[b]{0.33\linewidth}
        \includegraphics[width=\linewidth]{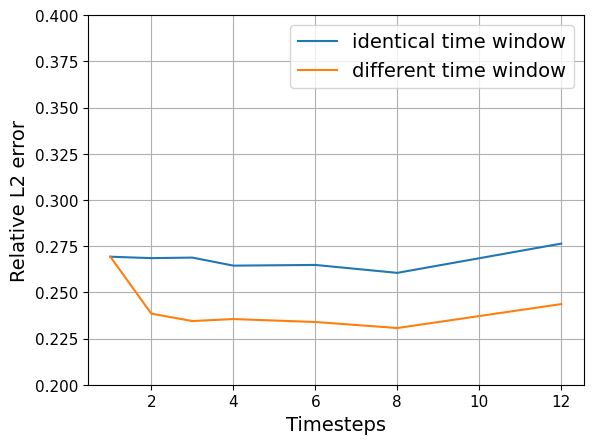}
        \caption{CLSTM}
    \end{subfigure}

    \caption{\an{\textbf{Sensitivity test of prolonged unavailability of sensors.} Comparison of relative L2 error on test holdout of real observation data. Models include (a) VUNet, (b) ViTAE and (c) CLSTM. Orange lines indicate that different stations are randomly excluded at each time step, whereas blue lines indicate that the same stations are excluded. }}
    \label{fig:different_vs_constant_time_window}
\end{figure}

\begin{figure}[htbp]
    \centering
    \includegraphics[width=0.9\linewidth]{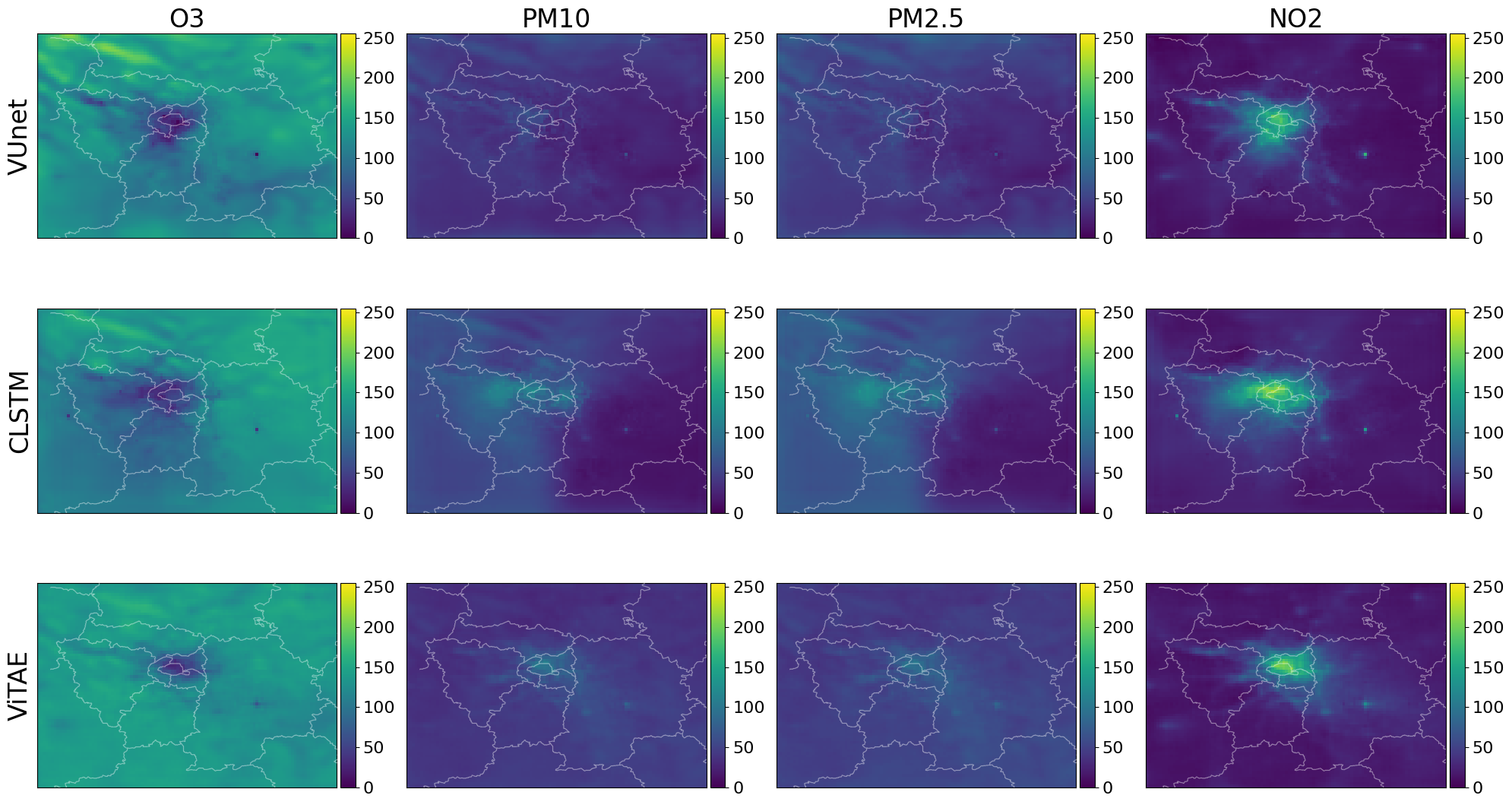}
    \caption{\an{\textbf{Predicted output when the same sensor is removed from time window.} Visual comparison of (a) VUNet (b) CLSTM (c) ViTAE models on O$_3$, PM$_{10}$, PM$_{2.5}$, NO$_2$ pollutants on December 13 12:00:00 2014. Same sensors are withheld from the entire historical time window to make predictions}}
    \label{fig:real-visual-same-window}
\end{figure}

\an{Figure \ref{fig:different_vs_constant_time_window} plots the reconstruction error as the time window length changes. This comparison allows us to verify the model's capacity to reconstruct when there is an unmonitored station for a longer time period. We visualise the reconstructions when the same sensor is removed from the entire time window. Figure \ref{fig:real-visual-same-window} shows that the prediction quality is similar to that obtained with randomly withheld sensors (Figure~\ref{fig:real-visual} of the main manuscript).}

\begin{figure}[htbp]
    \centering
    \makebox[\textwidth][c]{%
    \begin{minipage}{1.0\textwidth}
    \centering

    \includegraphics[width=\linewidth]{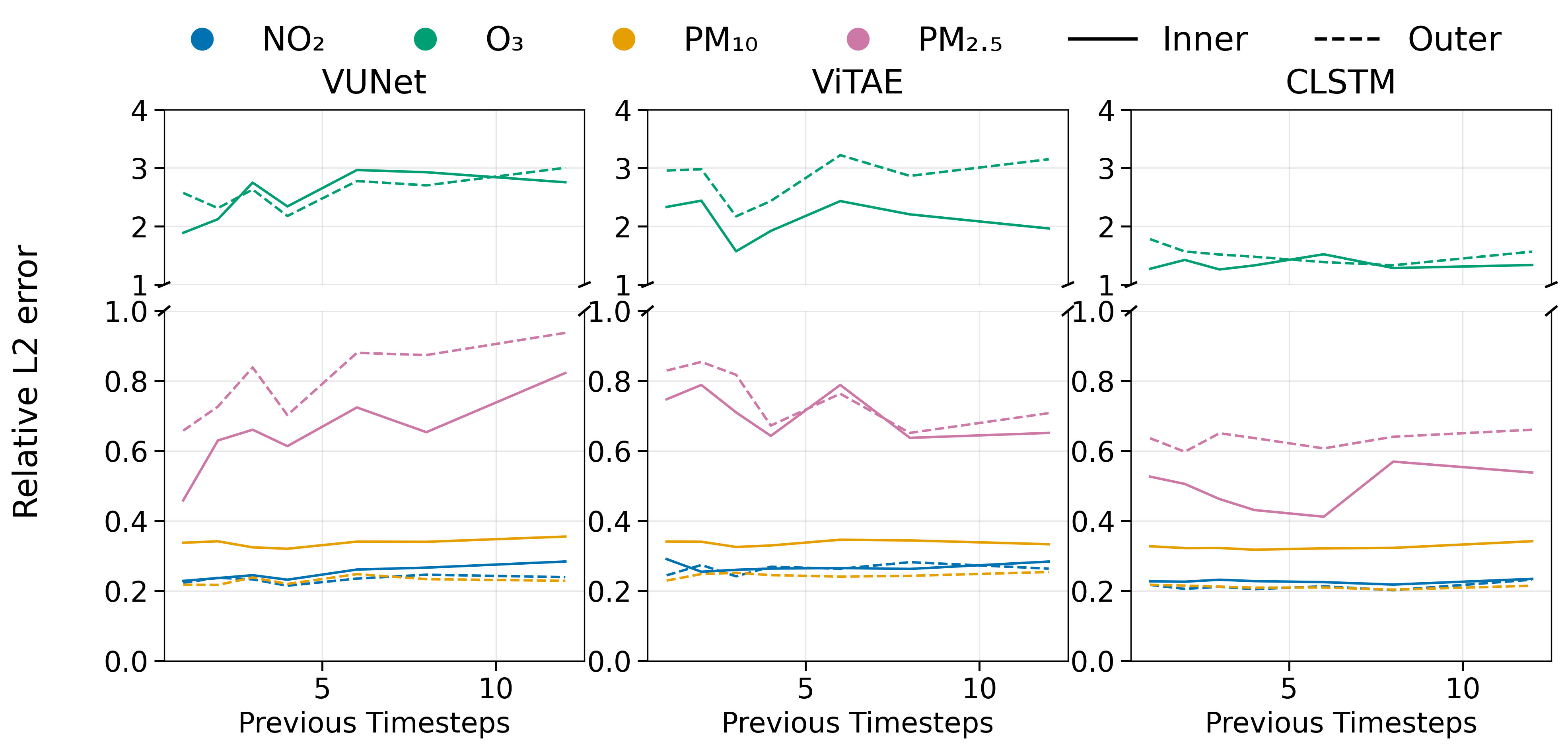}

    \end{minipage}
    }

    \caption{\an{\textbf{Inner- versus outer-city errors for different pollutants on real data.} This figure compares the relative L2 error metric with different lengths of time window for various deterministic models. The solid lines indicate "inner" which implies that one inner-city station is left out from the complete time window and "outer" indicates that one outer-city station is left out as a holdout point for metric evaluation. The blue, green, yellow, pink lines are for NO$_2$, O$_3$, PM$_{10}$, PM$_{2.5}$ respectively}}
    \label{fig:inner_vs_outer_same_window}
\end{figure}

\an{Figure \ref{fig:inner_vs_outer_same_window} compares the performance of various models on each pollutant on the real observation dataset as the time window length changes. To assess the effect of sensor location, we remove the same sensor throughout each historical input window from either an inner- or outer-city location. To reduce potential bias from a single location, the excluded sensor is randomly selected for each historical window.}

\section{Simulation data results}\label{appendix_simu_table}

In this section, we present detailed pollutant-specific results for various metrics discussed in the manuscript for the simulation dataset during the test period.

\subsection{Joint vs individual training}

Table \ref{tab:all_simulation_results} displays the metrics for each pollutant type and training methodology (joint vs single) on the simulation dataset for the CLSTM, VUNet, ViTAE and diffusion models.

\begin{figure}[htbp]
    \centering
    \vspace{-5mm}
    
    \makebox[\linewidth][c]{%
\begin{minipage}{1.4\textwidth}
        \centering
        \begin{subfigure}{0.19\linewidth}
            \includegraphics[width=\linewidth]{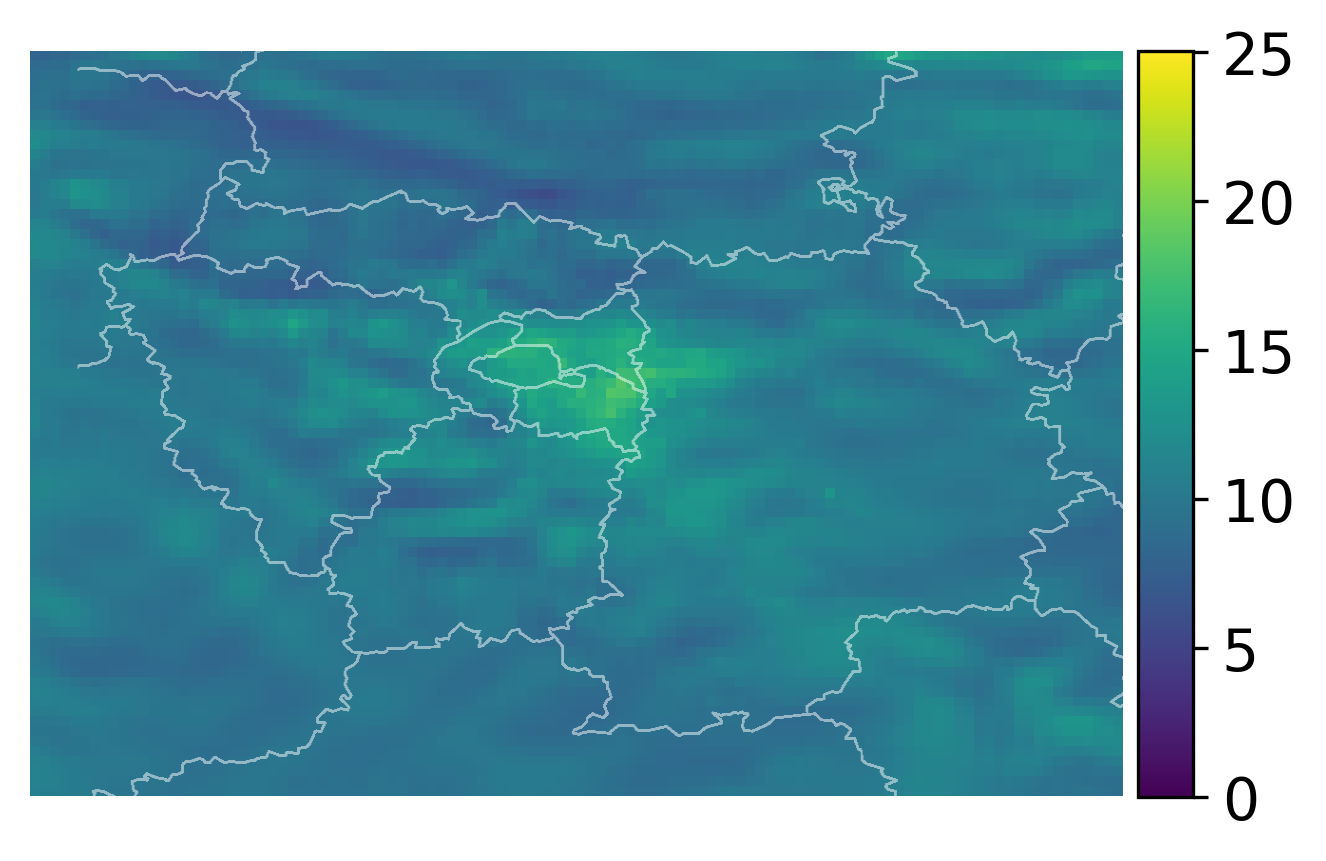}
            \caption*{PM$_{2.5}$ reference}
        \end{subfigure}
        \begin{subfigure}{0.19\linewidth}
            \includegraphics[width=\linewidth]{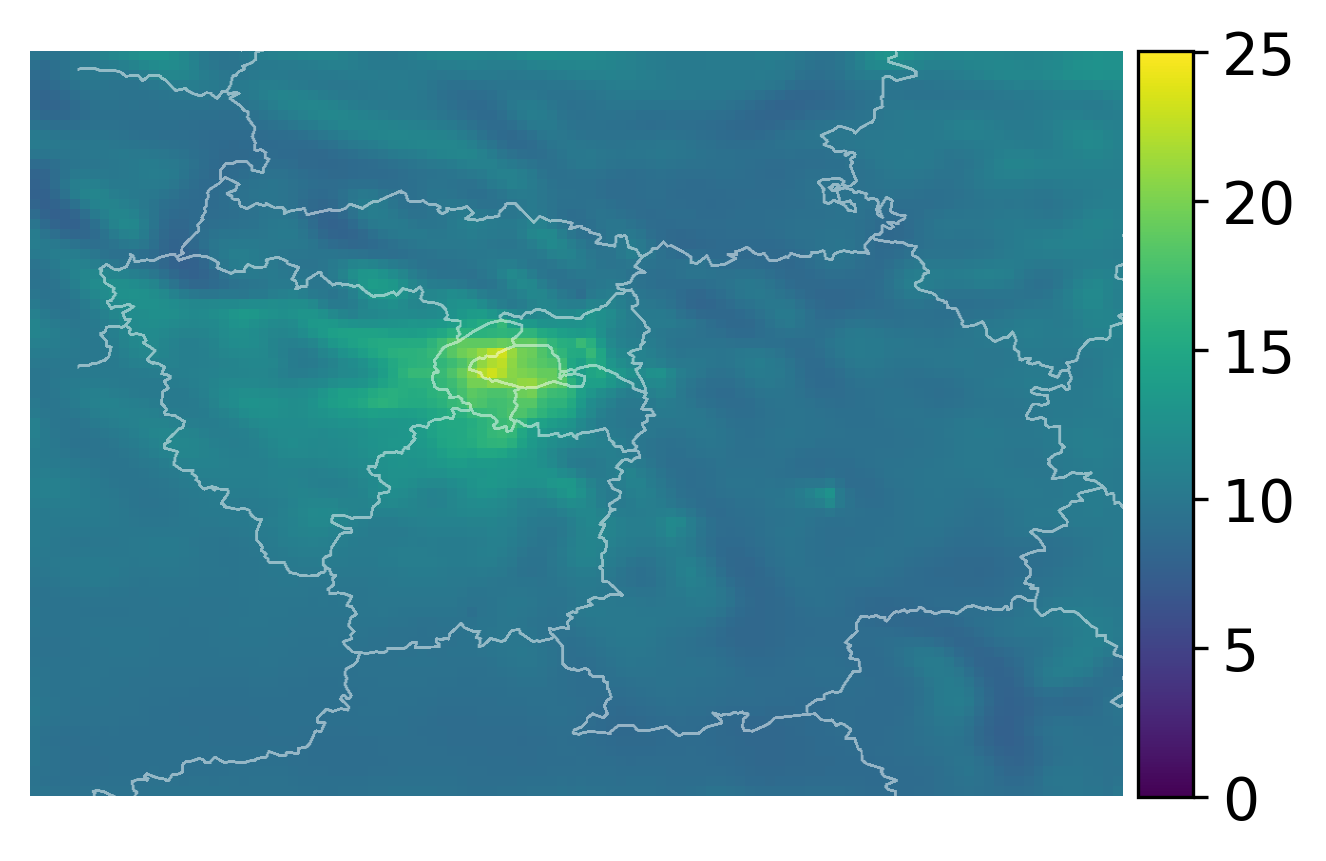}
            \caption*{Diffusion(individual)}
        \end{subfigure}
        \begin{subfigure}{0.19\linewidth}
            \includegraphics[width=\linewidth]{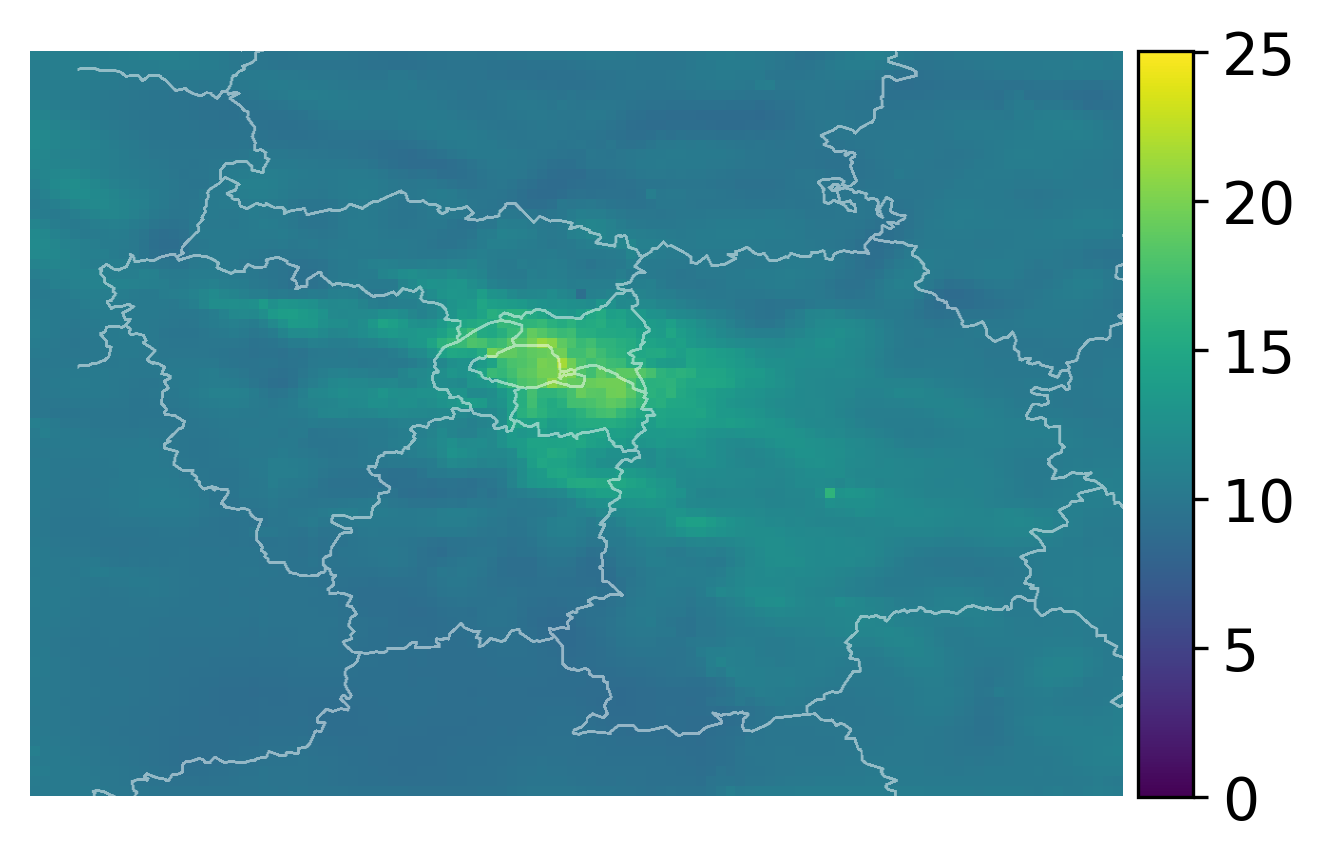}
            \caption*{Diffusion(joint)}
        \end{subfigure}
        \begin{subfigure}{0.19\linewidth}
            \includegraphics[width=\linewidth]{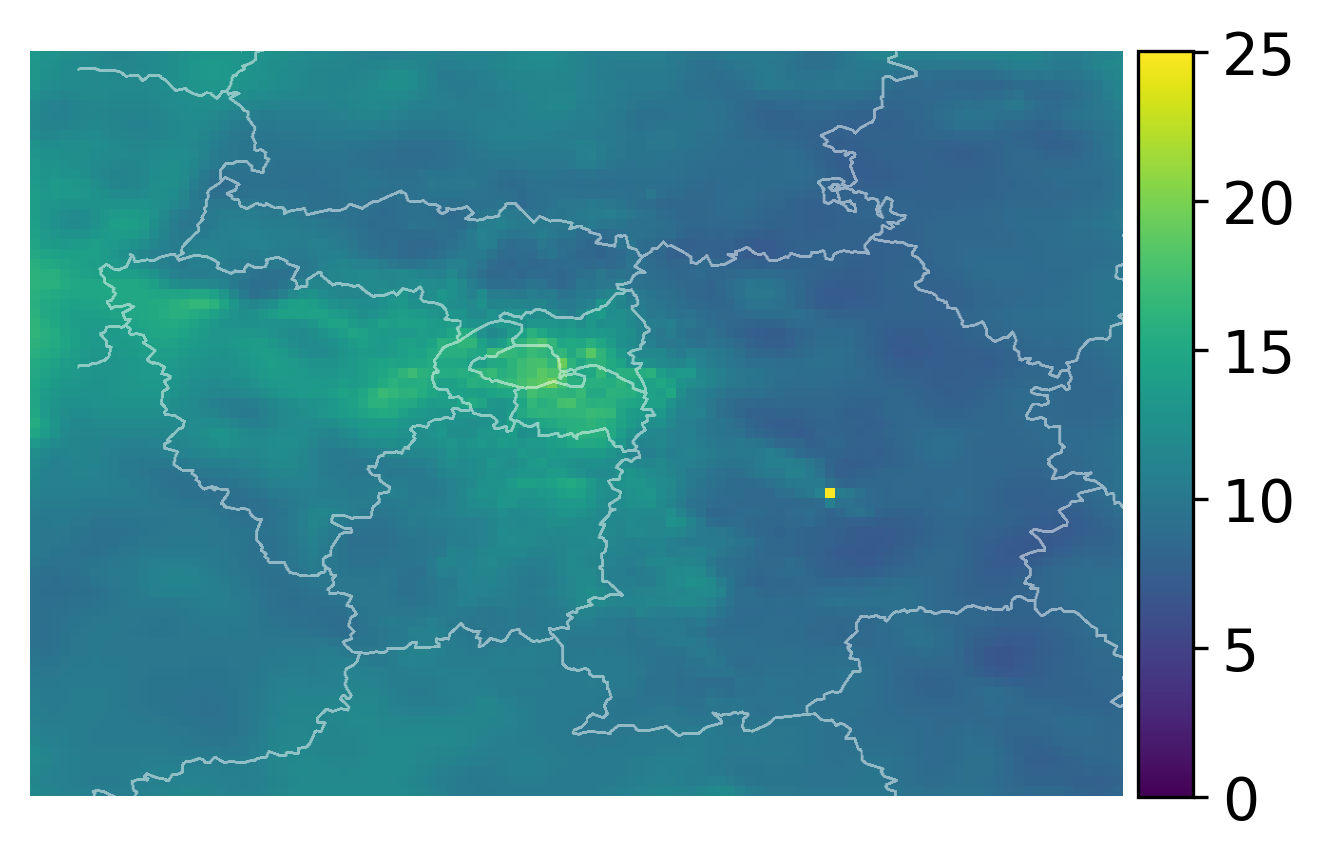}
            \caption*{VUNet(individual)}
        \end{subfigure}
        \begin{subfigure}{0.19\linewidth}
            \includegraphics[width=\linewidth]{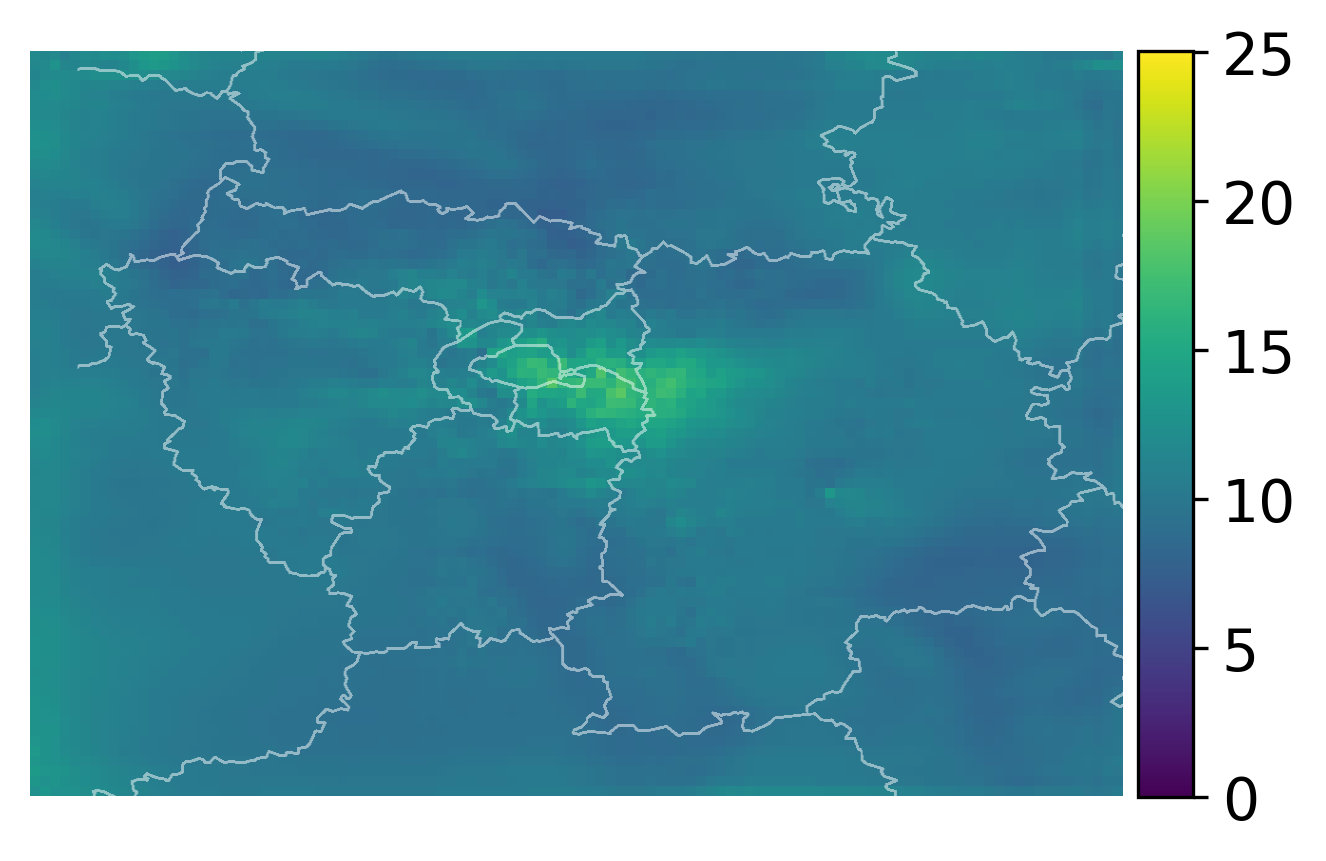}
            \caption*{VUNet(joint)}
        \end{subfigure}
    \end{minipage}%
    }

    \makebox[\linewidth][c]{%
\begin{minipage}{1.4\textwidth}
    \centering
        \hspace{0.19\linewidth}
        \begin{subfigure}{0.19\linewidth}
            \includegraphics[width=\linewidth]{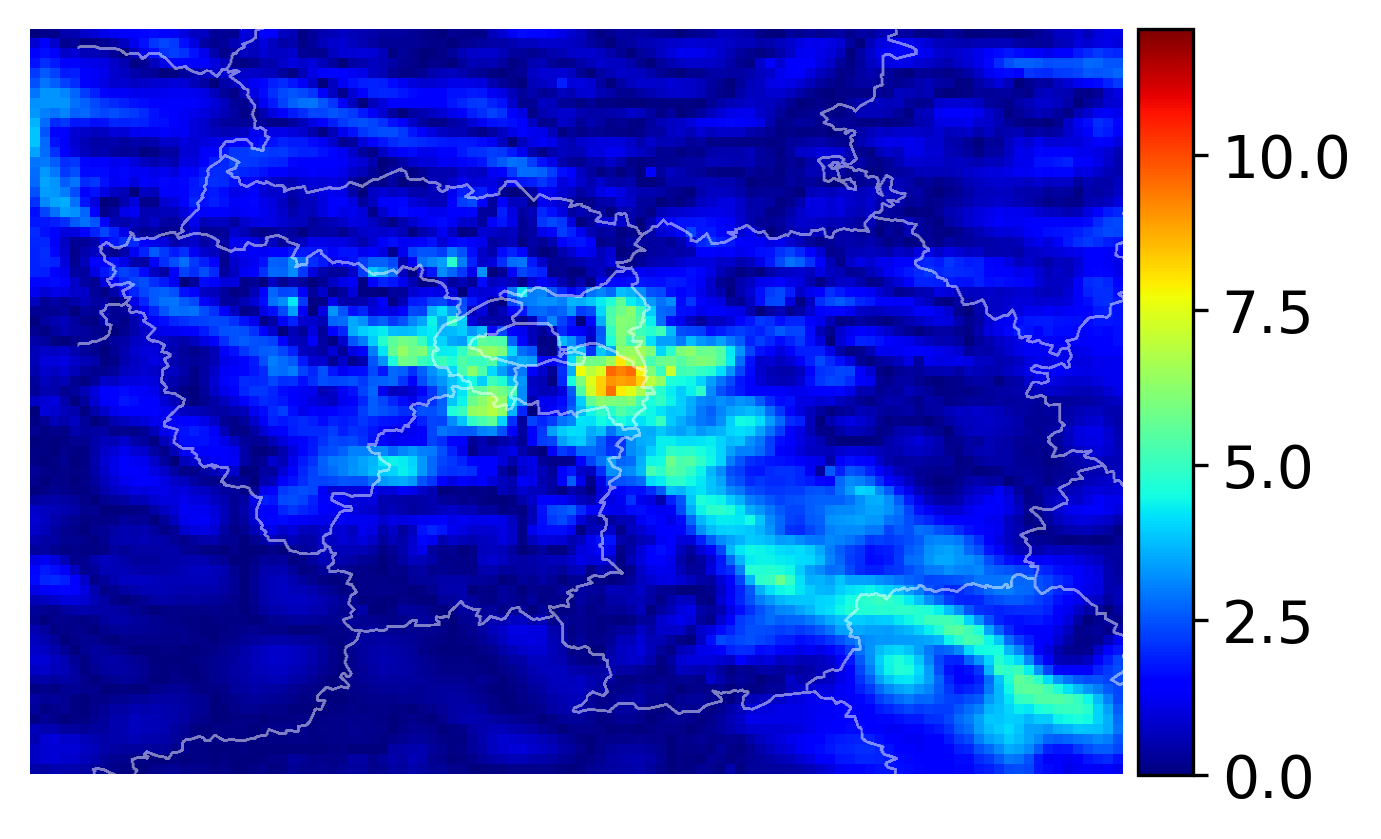}
            \caption*{Error}
        \end{subfigure}
        \begin{subfigure}{0.19\linewidth}
            \includegraphics[width=\linewidth]{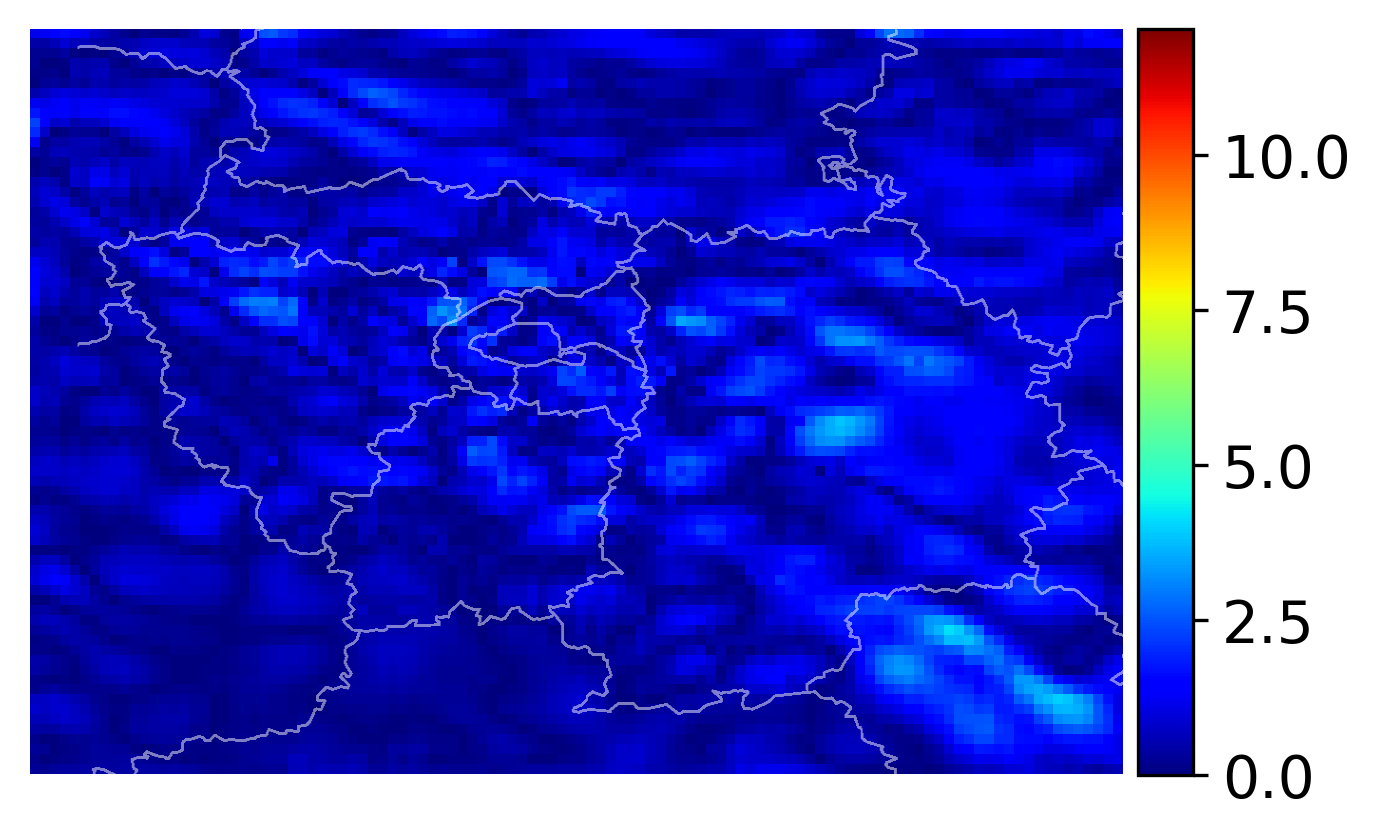}
            \caption*{Error}
        \end{subfigure}
        \begin{subfigure}{0.19\linewidth}
            \includegraphics[width=\linewidth]{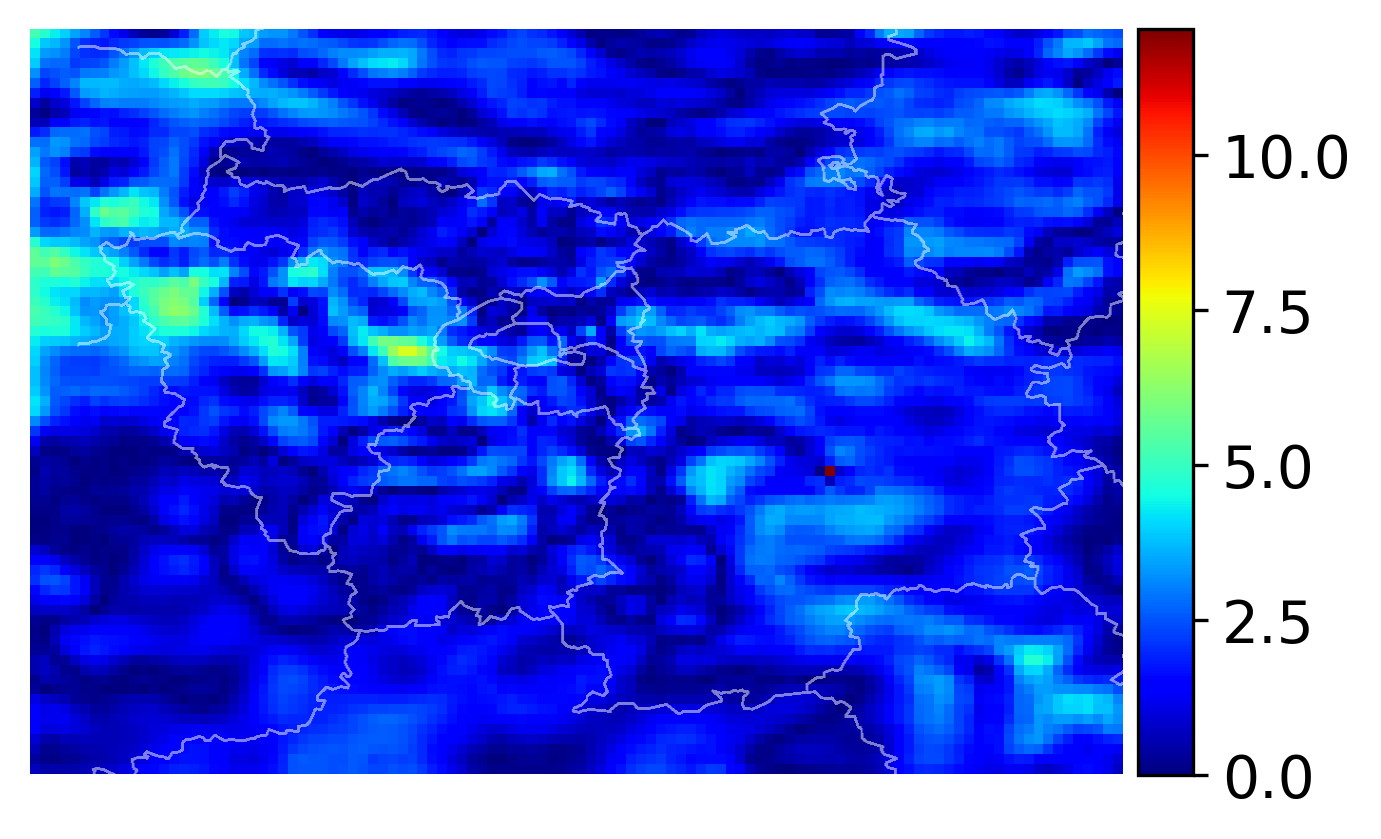}
            \caption*{Error}
        \end{subfigure}
        \begin{subfigure}{0.19\linewidth}
            \includegraphics[width=\linewidth]{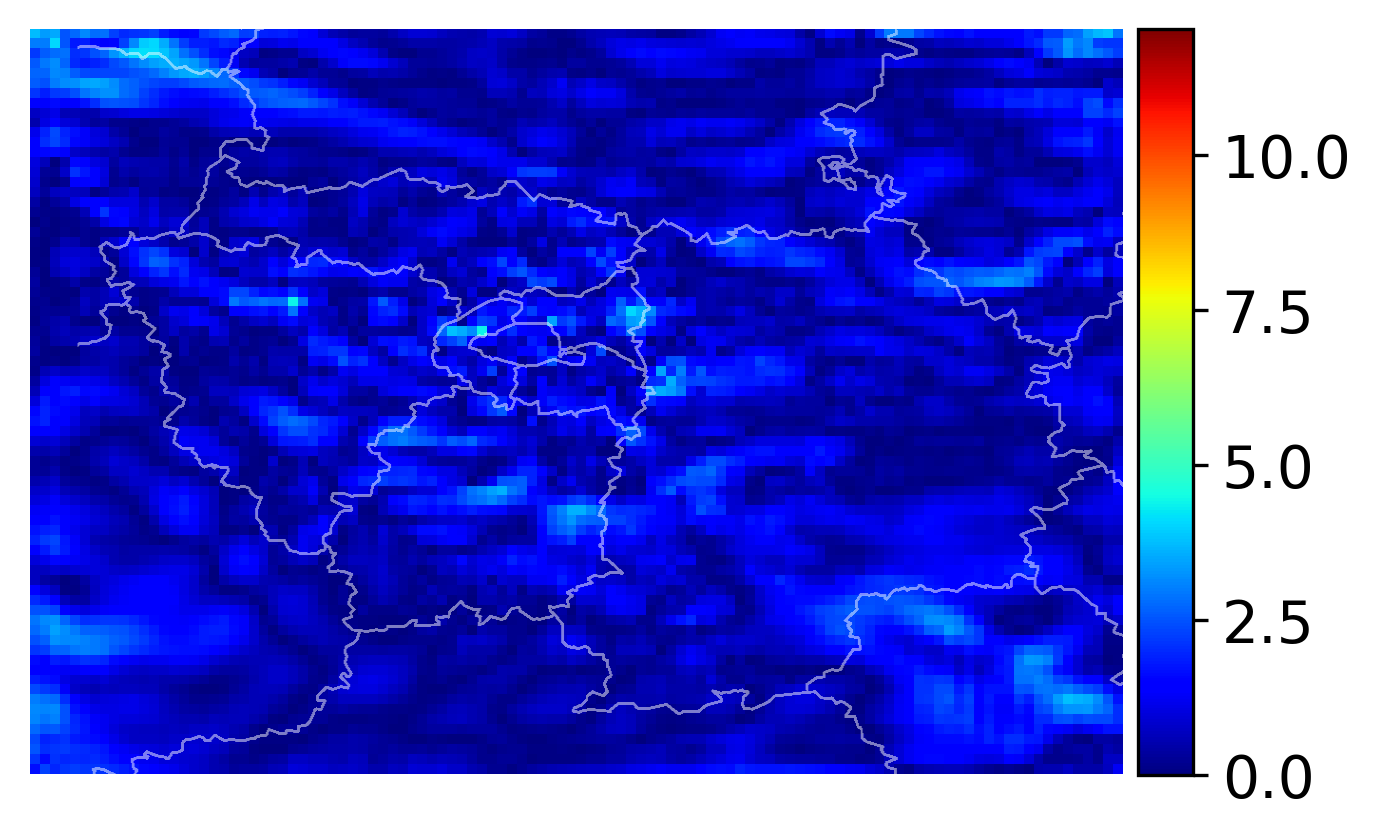}
            \caption*{Error}
        \end{subfigure}

    \end{minipage}%
    }

    \caption{\an{\textbf{Visual comparison of joint and individual model training.} Side-by-side comparison of the PM$_{2.5}$ predictions at 8 PM on the 1st of December for Diffusion and VUNet models trained individually and jointly}}
    \label{fig:pm25_joint_single}
\end{figure}

Figure \ref{fig:pm25_joint_single} compares the predictions of the VUNet and Diffusion model trained on individual pollutants versus joint modelling on the simulation dataset. We observe the absolute error diagram highlighting larger errors for individually trained models when compared to joint modelling.

\begin{table}[htbp]
\centering
\setlength{\tabcolsep}{3pt}
\footnotesize
\renewcommand{\arraystretch}{1.2}

\begin{adjustbox}{max width=1.2\textwidth,center}
\begin{tabular}{ll|cc|cc|cc|cc}
\toprule

\multirow{2}{*}{\textbf{Pollutant}} &
\multirow{2}{*}{\textbf{Model}} &
\multicolumn{2}{c|}{\textbf{Relative L2 $\blacktriangledown$}} &
\multicolumn{2}{c|}{\textbf{SSIM $\vartriangle$}} &
\multicolumn{2}{c|}{\textbf{MFE $\blacktriangledown$}} &
\multicolumn{2}{c}{\textbf{MFB $\approxeq0$}} \\

\cmidrule(lr){3-4}
\cmidrule(lr){5-6}
\cmidrule(lr){7-8}
\cmidrule(lr){9-10}

& & \textbf{v1} & \textbf{v2}
  & \textbf{v1} & \textbf{v2}
  & \textbf{v1} & \textbf{v2}
  & \textbf{v1} & \textbf{v2} \\

\midrule

\multirow{4}{*}{\textbf{NO$_2$}}
& CLSTM(t=8)
& 0.268 & 0.266
& 0.82 & 0.812
& 0.263 & 0.269
& \textbf{0.048} & 0.081 \\

& VUNet(t=4)
& 0.217 & 0.236
& 0.882 & 0.840
& 0.233 & 0.268
& 0.091 & 0.143 \\

& ViTAE(t=6)
& 0.239 & 0.258
& 0.869 & 0.837
& 0.279 & 0.256
& 0.187 & 0.088 \\

& Diffusion(E=20)
& \textbf{0.207} & 0.213
& \textbf{0.898} & 0.890
& \textbf{0.199} & 0.211
& 0.079 & 0.081 \\

\midrule

\multirow{4}{*}{\textbf{O$_3$}}
& CLSTM(t=8)
& 0.092 & 0.09
& 0.565 & 0.551
& 0.080 & \textbf{0.080}
& -0.004 & -0.003 \\

& VUNet(t=4)
& 0.089 & \textbf{0.089}
& 0.586 & 0.622
& 0.082 & 0.232
& \textbf{-0.001} & 0.148 \\

& ViTAE(t=6)
& 0.105 & 0.11
& 0.442 & 0.566
& 0.091 & 0.099
& 0.012 & -0.029 \\

& Diffusion(E=20)
& 0.091 & 0.091
& 0.619 & \textbf{0.642}
& 0.081 & 0.080
& -0.021 & -0.031 \\

\midrule

\multirow{4}{*}{\textbf{PM$_{10}$}}
& CLSTM(t=8)
& 0.266 & \textbf{0.182}
& 0.544 & 0.614
& 0.190 & \textbf{0.143}
& -0.123 & -0.046 \\

& VUNet(t=4)
& 0.192 & 0.186
& 0.636 & 0.659
& 0.153 & 0.152
& -0.032 & -0.044 \\

& ViTAE(t=6)
& 0.216 & 0.228
& 0.549 & 0.629
& 0.171 & 0.185
& \textbf{-0.029} & -0.056 \\

& Diffusion(E=20)
& 0.213 & 0.183
& 0.667 & \textbf{0.699}
& 0.372 & 0.144
& -0.169 & -0.037 \\

\midrule

\multirow{4}{*}{\textbf{PM$_{2.5}$}}
& CLSTM(t=8)
& 0.245 & 0.188
& 0.471 & 0.565
& 0.184 & 0.148
& -0.065 & -0.069 \\

& VUNet(t=4)
& 0.231 & 0.193
& 0.545 & 0.619
& 0.182 & 0.157
& -0.077 & -0.07 \\

& ViTAE(t=6)
& 0.255 & 0.229
& 0.482 & 0.586
& 0.197 & 0.188
& -0.074 & -0.079 \\

& Diffusion(E=20)
& 0.222 & \textbf{0.182}
& 0.587 & \textbf{0.659}
& 0.401 & \textbf{0.146}
& -0.212 & \textbf{-0.052} \\

\bottomrule

\end{tabular}
\end{adjustbox}

\caption{\an{Evaluation of reconstruction accuracy using relative L2 error, SSIM, MFE, and MFB for each pollutant type. Performance comparison of v1 and v2 across the four model architectures and pollutant types. v1 denotes single pollutant training, whereas v2 denotes joint pollutant training. The $\blacktriangledown$ symbol indicates that a lower value is better, $\vartriangle$ indicates that a higher value is better, and $\approxeq0$ indicates that a value closer to zero is better. The best-performing value for each pollutant and metric is shown in bold.}}

\label{tab:all_simulation_results}
\end{table}

\vfill

\subsection{\ans{Sensitivity to Cross-Pollutant Information}}

\ans{To examine the contribution from other pollutants to joint reconstruction, we conduct an input-shuffling experiment using simulation data for VUNet and diffusion model. This experiment considers only the current time step as input. For each target pollutant, its observations are unchanged, while the the other pollutants are shuffled temporally. Table \ref{tab:shuffle_results} reports the mean relative L2 error for the regular and shuffled case for all pollutants. The results show that the shuffled scenario has noticeably higher errors, especially for the pollutants with fewest available sensors (PM$_{10}$ and PM$_{2.5}$). Together with the results from Table~\ref{tab:all_simulation_results}, these findings show that the benefits of joint modelling are partly due to the information carried by other pollutants.}

\begin{table}[ht]
\centering

\begin{adjustbox}{max width=\linewidth,center}
\begin{tabular}{lcccccccc}
\toprule
& \multicolumn{2}{c}{\textbf{O$_3$}}
& \multicolumn{2}{c}{\textbf{PM$_{10}$}}
& \multicolumn{2}{c}{\textbf{PM$_{2.5}$}}
& \multicolumn{2}{c}{\textbf{NO$_2$}} \\
\cmidrule(lr){2-3}
\cmidrule(lr){4-5}
\cmidrule(lr){6-7}
\cmidrule(lr){8-9}
\textbf{Model}
& \textbf{Regular} & \textbf{Shuffle}
& \textbf{Regular} & \textbf{Shuffle}
& \textbf{Regular} & \textbf{Shuffle}
& \textbf{Regular} & \textbf{Shuffle} \\
\midrule
VUNet
& 0.0956 & 0.1187
& 0.2066 & 0.6050
& 0.2130 & 0.5367
& 0.2619 & 0.4112 \\

Diffusion
& 0.0906 & 0.1296
& 0.1835 & 0.4465
& 0.1827 & 0.5504
& 0.2131 & 0.3559 \\
\bottomrule
\end{tabular}
\end{adjustbox}
\caption{\ans{Comparison of regular and shuffled prediction results (mean relative L2 error) across models and pollutants on simulation dataset. In the "shuffled" scenario, the target pollutant is not changed whereas the other 3 pollutants are shuffled temporally. Both VUNet and diffusion model utilise only the current time step to reconstruct the field.}}
\label{tab:shuffle_results}
\end{table}

\subsection{Power spectrum plot}

\begin{figure}[htbp]
    \centering
    \vspace{-5mm}
    \begin{subfigure}[b]{0.45\linewidth}
        \includegraphics[width=\linewidth]{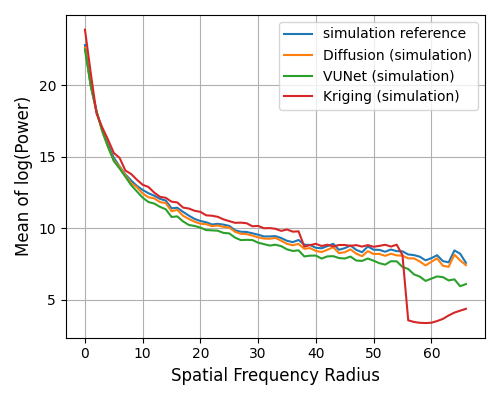}
        \caption{NO$_2$ }
    \end{subfigure}\hfill
    \begin{subfigure}[b]{0.45\linewidth}
        \includegraphics[width=\linewidth]{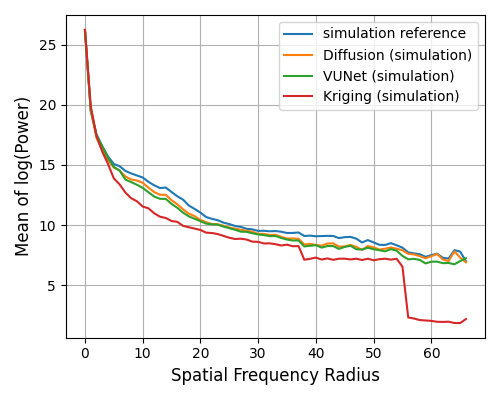}
        \caption{O$_3$}
    \end{subfigure}

    \begin{subfigure}[b]{0.45\linewidth}
        \includegraphics[width=\linewidth]{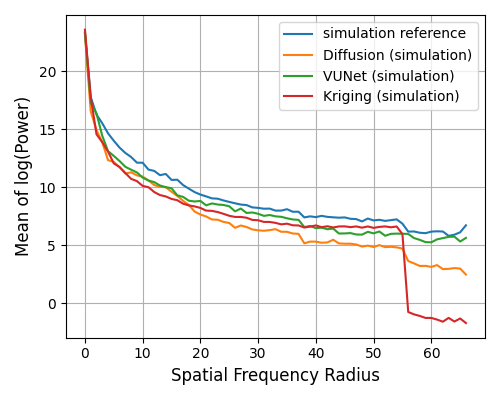}
        \caption{PM$_{10}$}
    \end{subfigure}\hfill
    \begin{subfigure}[b]{0.45\linewidth}
        \includegraphics[width=\linewidth]{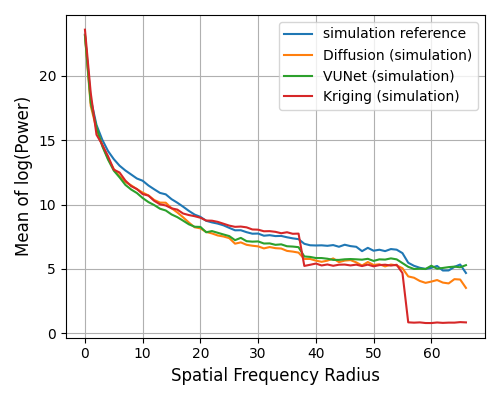}
        \caption{PM$_{2.5}$}
    \end{subfigure}

    \caption{\textbf{Spectral plot of the predictions of simulation data for each pollutant.} Comparison of the logarithm of the radially averaged power spectrum for ground truth of simulation data (blue line) and predicted results from diffusion (orange line), VUNet (green line) and the Kriging model (red line). We show results for each pollutant (a) NO$_2$, (b) O$_3$, (c) PM$_{10}$, (d) PM$_{2.5}$.  }
    \label{fig:psd_pollutants}
\end{figure}

Figure \ref{fig:psd_pollutants} displays the power spectrum plot for individual pollutants on the predicted results in the test dataset of various models.

\subsection{Prediction visualisations}\label{appendix_simu_visual}
\begin{figure}[htbp]
    \centering
    
    \makebox[\linewidth][c]{%
\begin{minipage}{1.4\textwidth}
        \centering
        \begin{subfigure}{0.19\linewidth}
            \includegraphics[width=\linewidth]{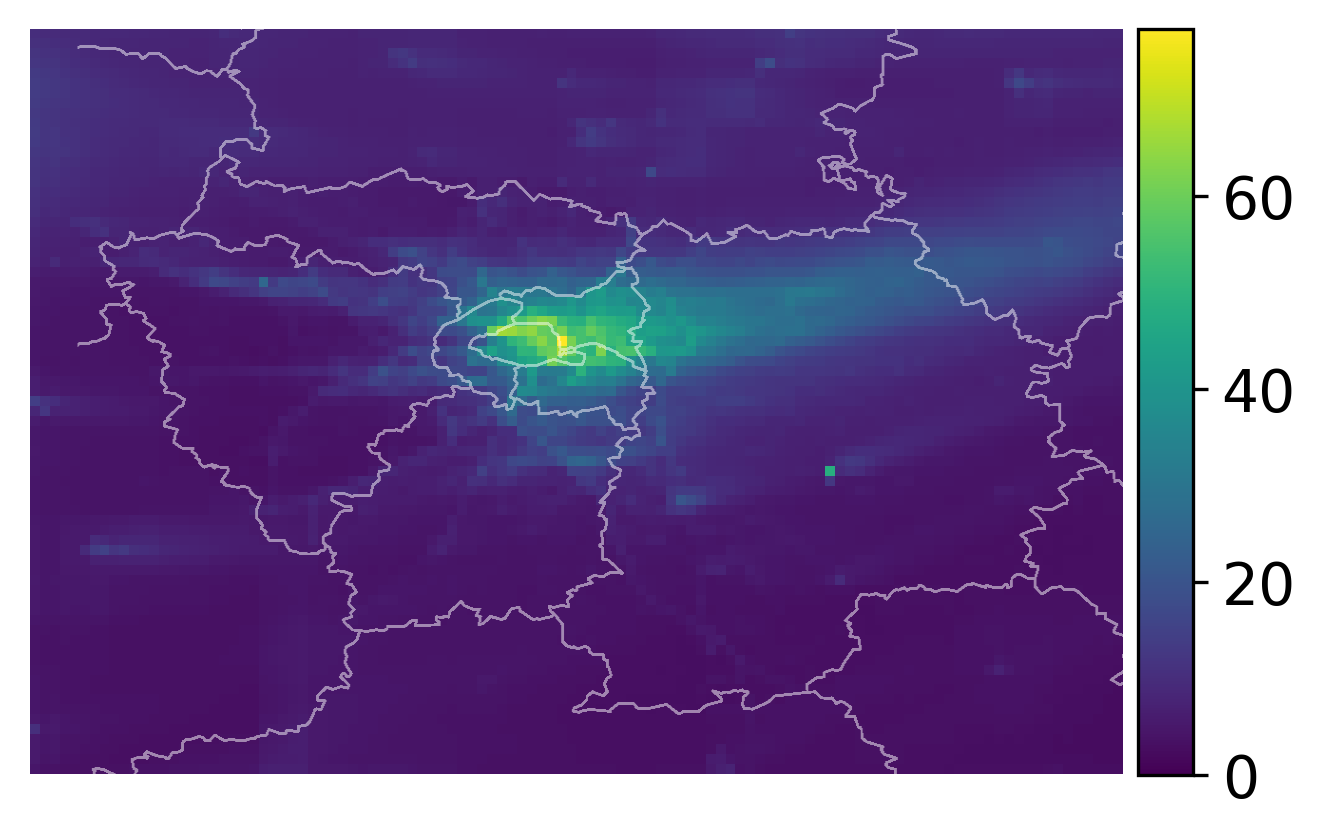}
            \caption*{NO$_2$ reference}
        \end{subfigure}
        \begin{subfigure}{0.19\linewidth}
            \includegraphics[width=\linewidth]{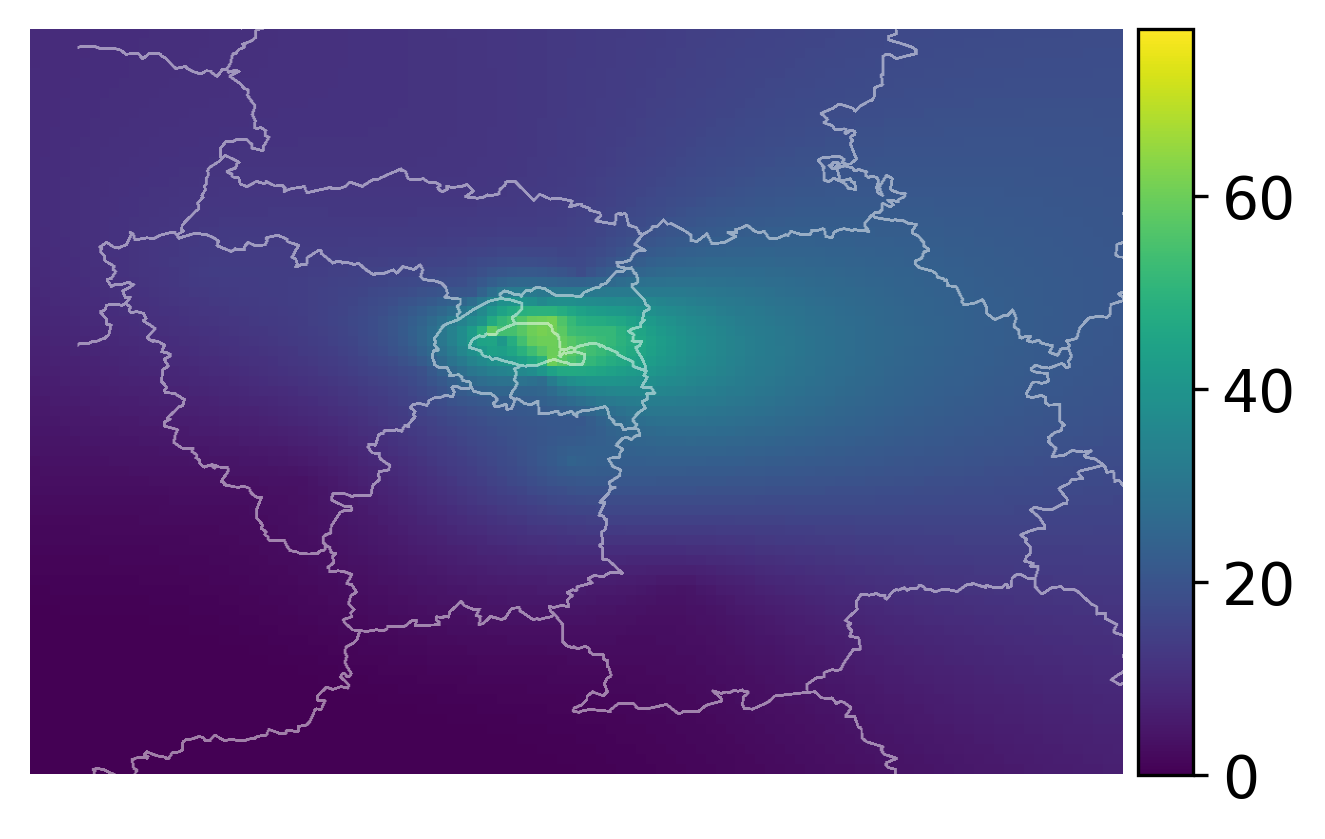}
            \caption*{Kriging}
        \end{subfigure}
        \begin{subfigure}{0.19\linewidth}
            \includegraphics[width=\linewidth]{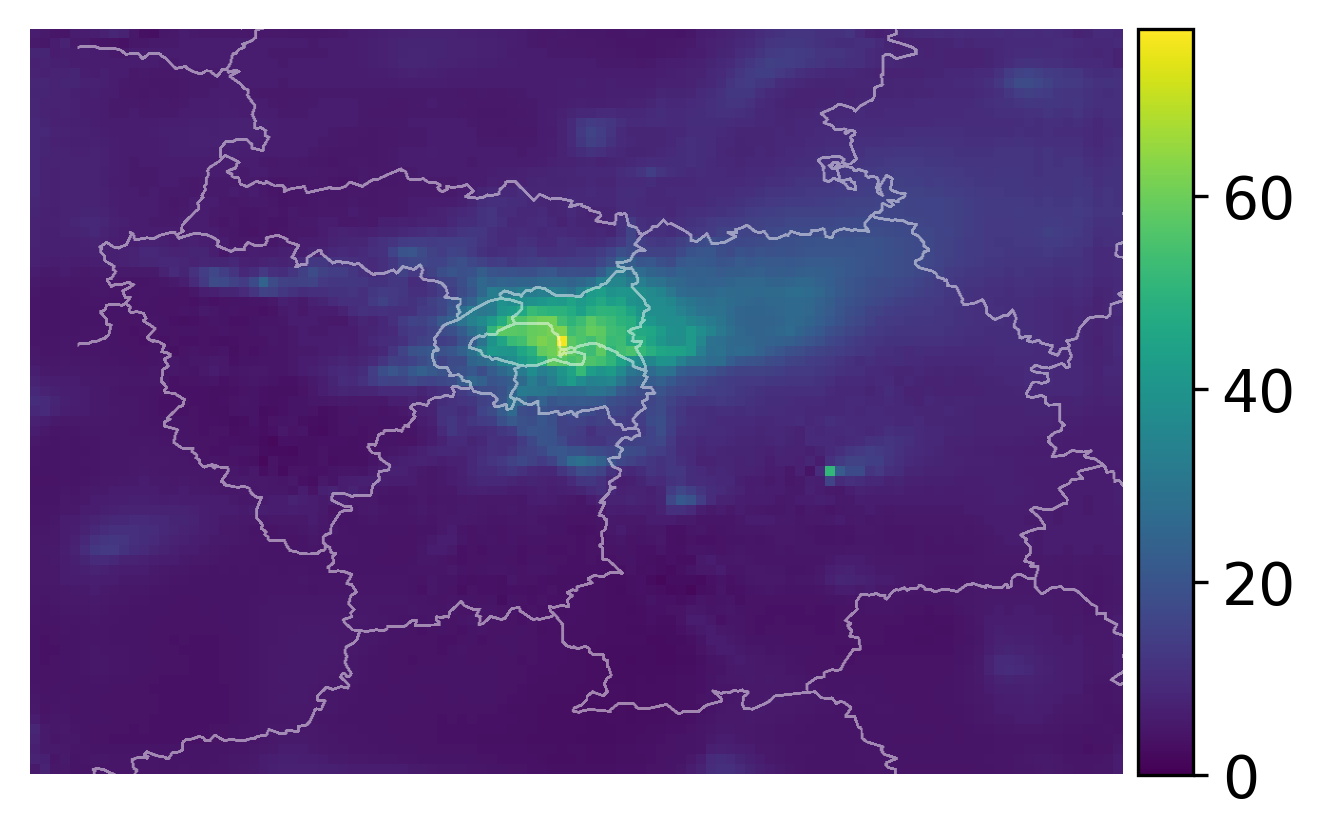}
            \caption*{VUNet}
        \end{subfigure}
        \begin{subfigure}{0.19\linewidth}
            \includegraphics[width=\linewidth]{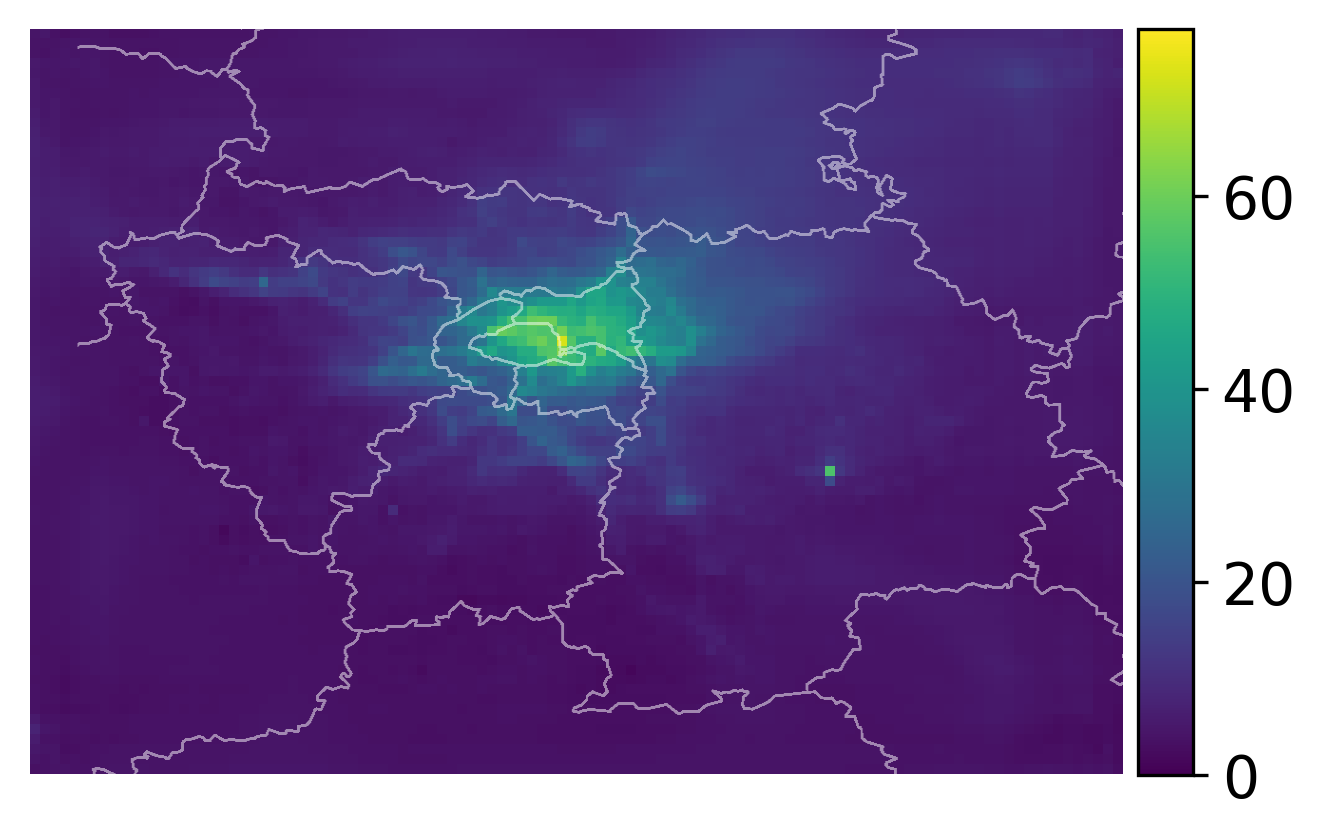}
            \caption*{CLSTM}
        \end{subfigure}
        \begin{subfigure}{0.19\linewidth}
            \includegraphics[width=\linewidth]{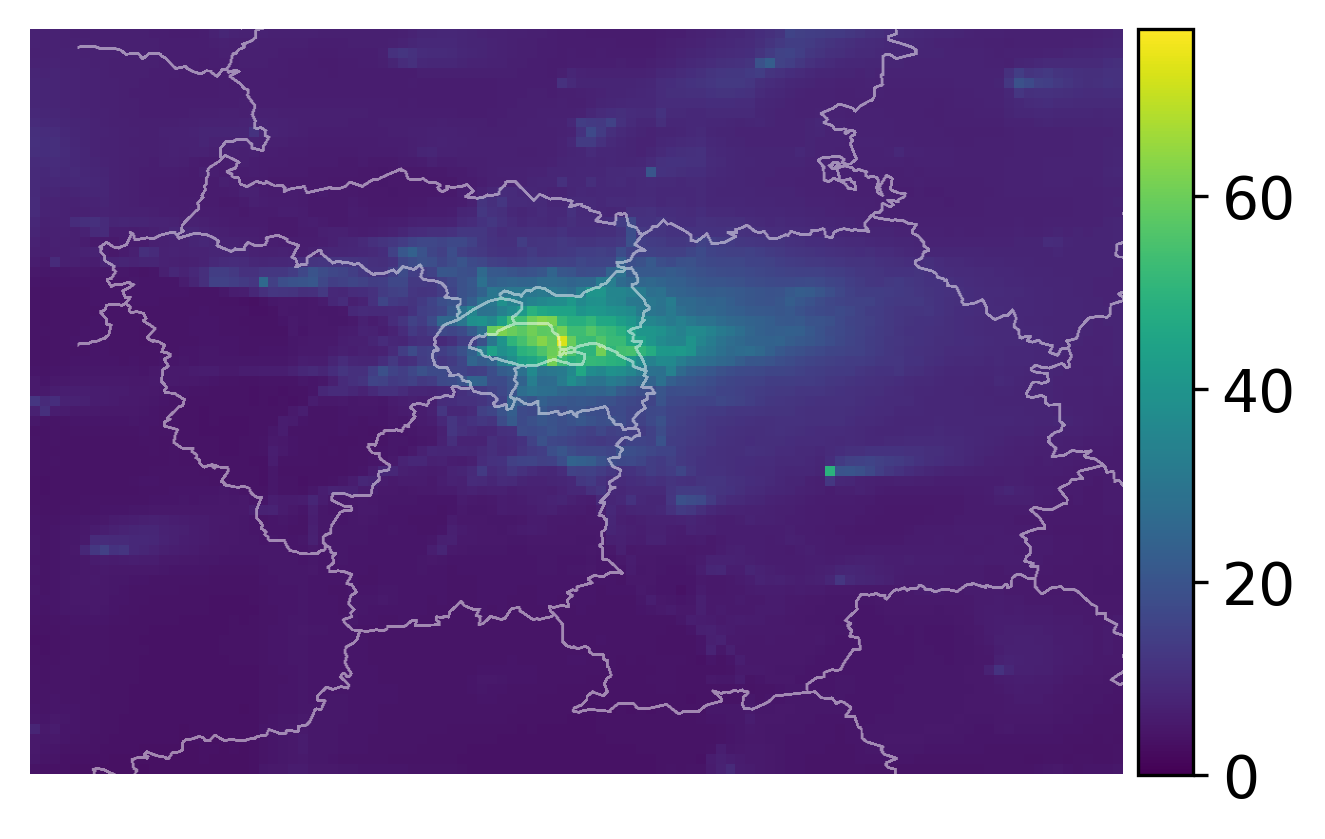}
            \caption*{Diffusion}
        \end{subfigure}

    \end{minipage}%
    }

    \makebox[\linewidth][c]{%
\begin{minipage}{1.4\textwidth}
    \centering
        \hspace{0.19\linewidth}
        \begin{subfigure}{0.19\linewidth}
            \includegraphics[width=\linewidth]{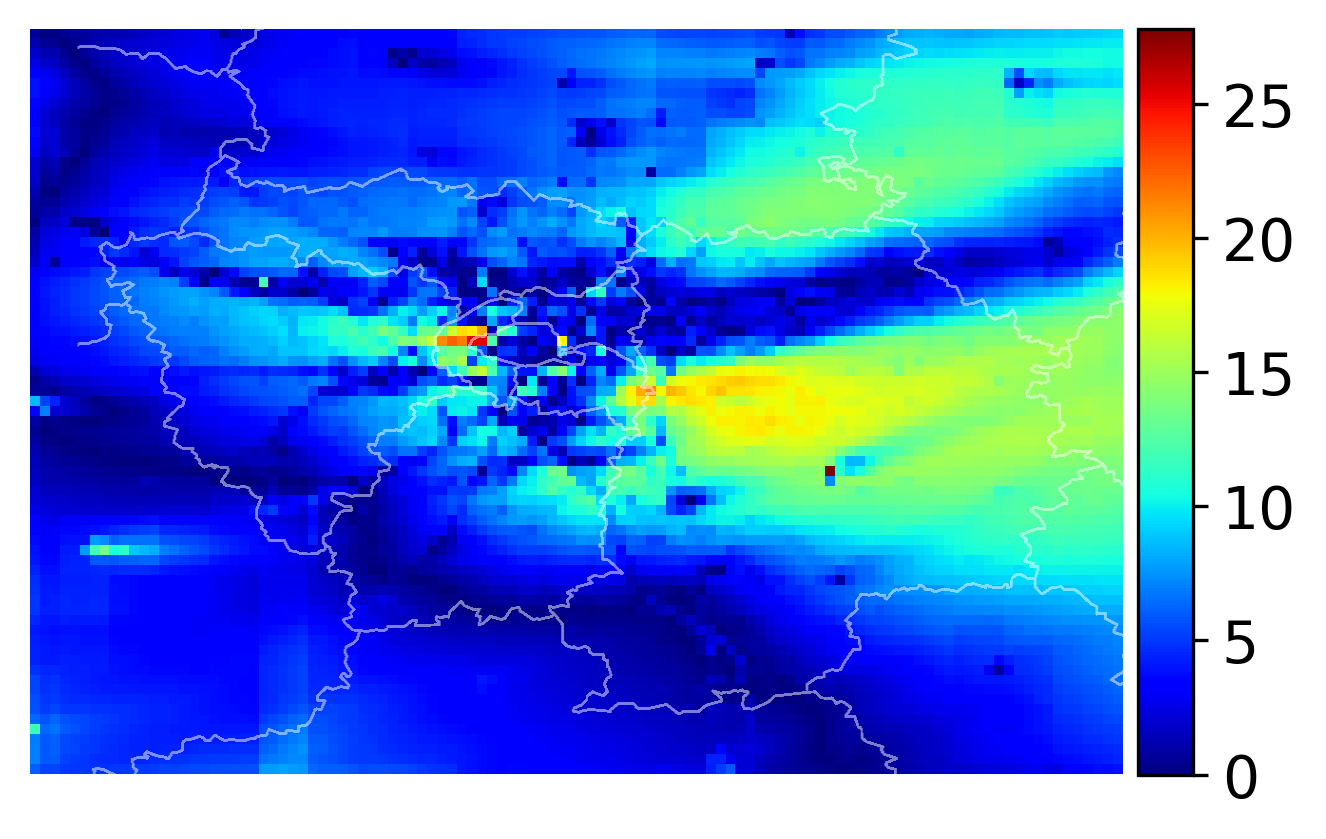}
            \caption*{Kriging Error}
        \end{subfigure}
        \begin{subfigure}{0.19\linewidth}
            \includegraphics[width=\linewidth]{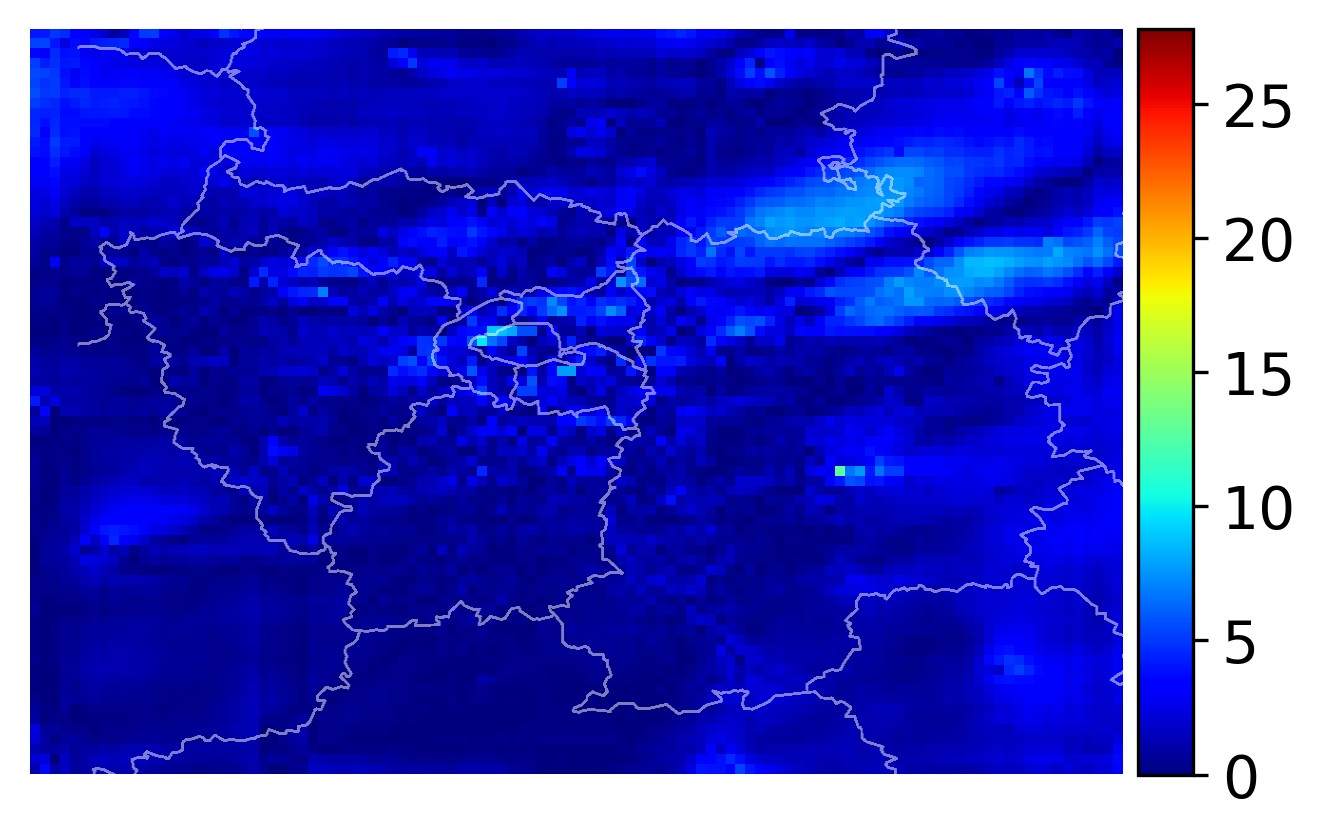}
            \caption*{VUNet Error}
        \end{subfigure}
        \begin{subfigure}{0.19\linewidth}
            \includegraphics[width=\linewidth]{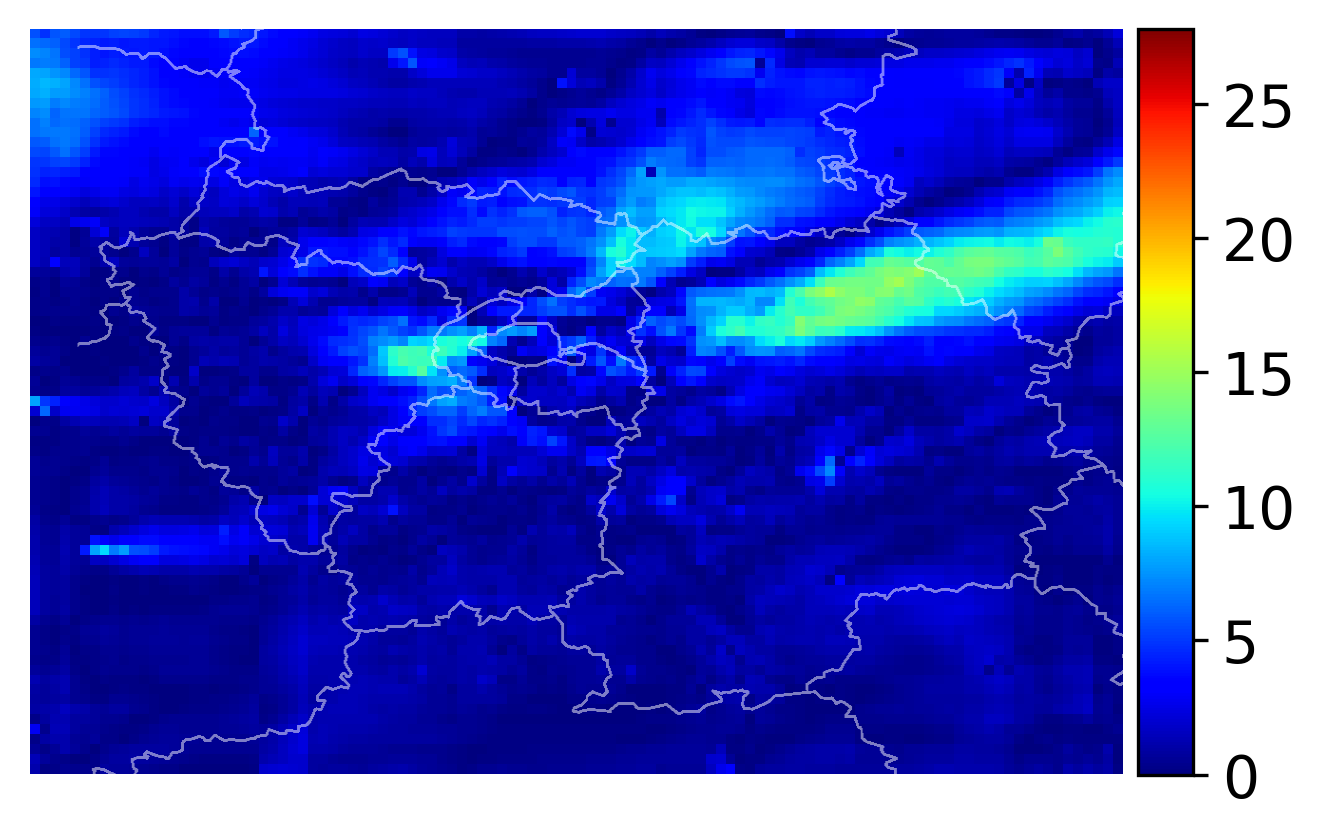}
            \caption*{CLSTM error}
        \end{subfigure}
        \begin{subfigure}{0.19\linewidth}
            \includegraphics[width=\linewidth]{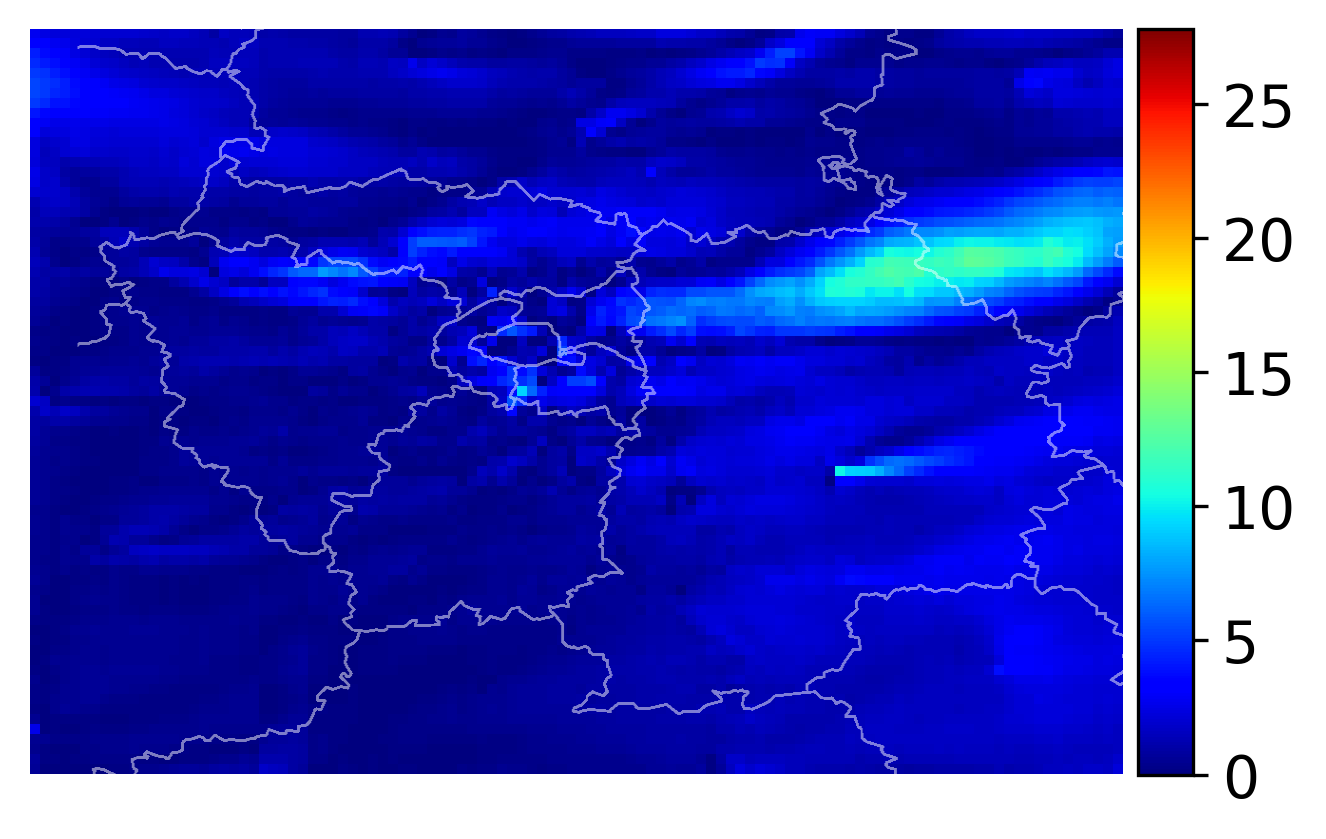}
            \caption*{Diffusion error}
        \end{subfigure}

    \end{minipage}%
    }

    \caption{\textbf{Prediction visualisation of each model on NO$_2$ simulation data.} Side-by-side comparison of the NO$_2$ predictions at 8 AM on the 1st of December for Kriging, VUNet, CLSTM and the diffusion model}
    \label{fig:no2_comparison_random}
\end{figure}

\begin{figure}[htbp]
    \centering
    
    \makebox[\linewidth][c]{%
\begin{minipage}{1.4\textwidth}
        \centering
        \begin{subfigure}{0.19\linewidth}
            \includegraphics[width=\linewidth]{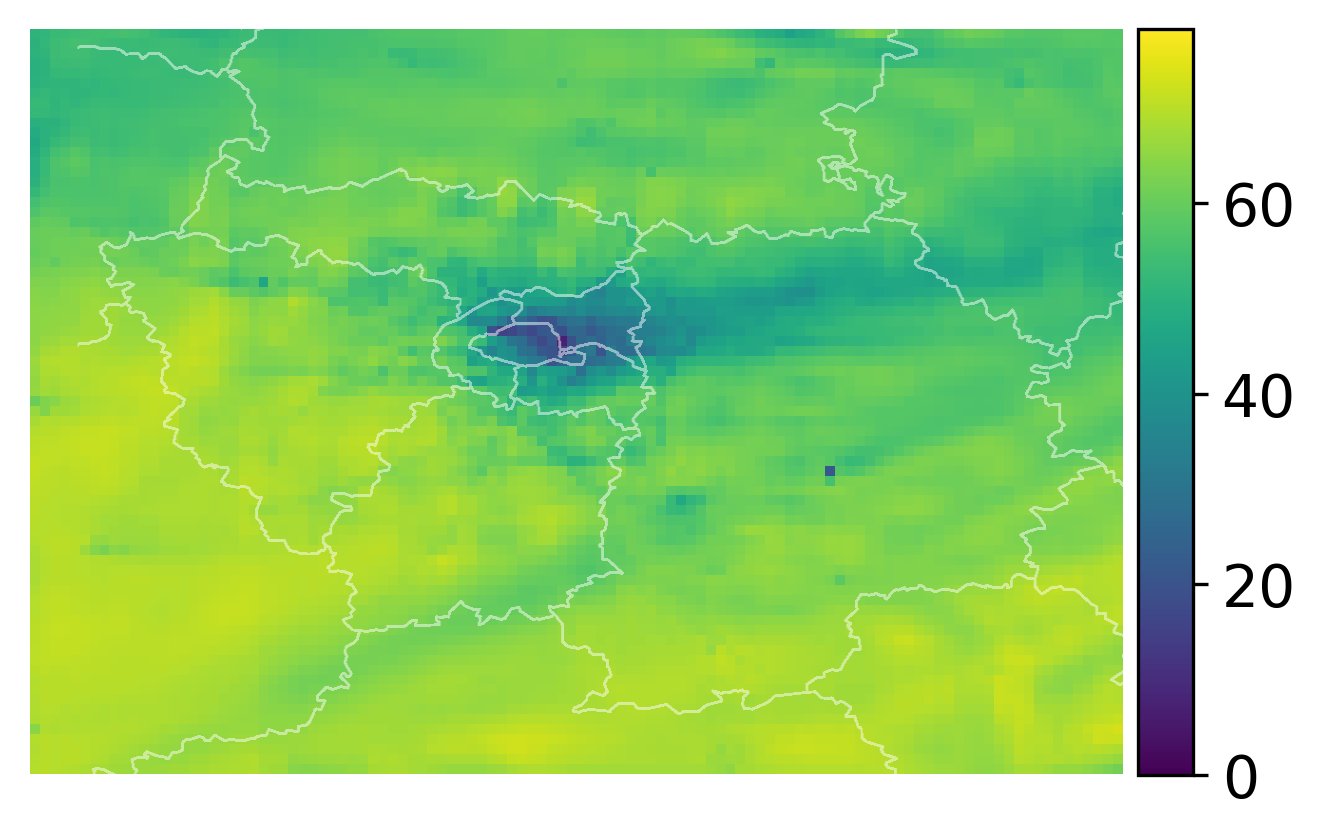}
            \caption*{O$_3$ reference}
        \end{subfigure}
        \begin{subfigure}{0.19\linewidth}
            \includegraphics[width=\linewidth]{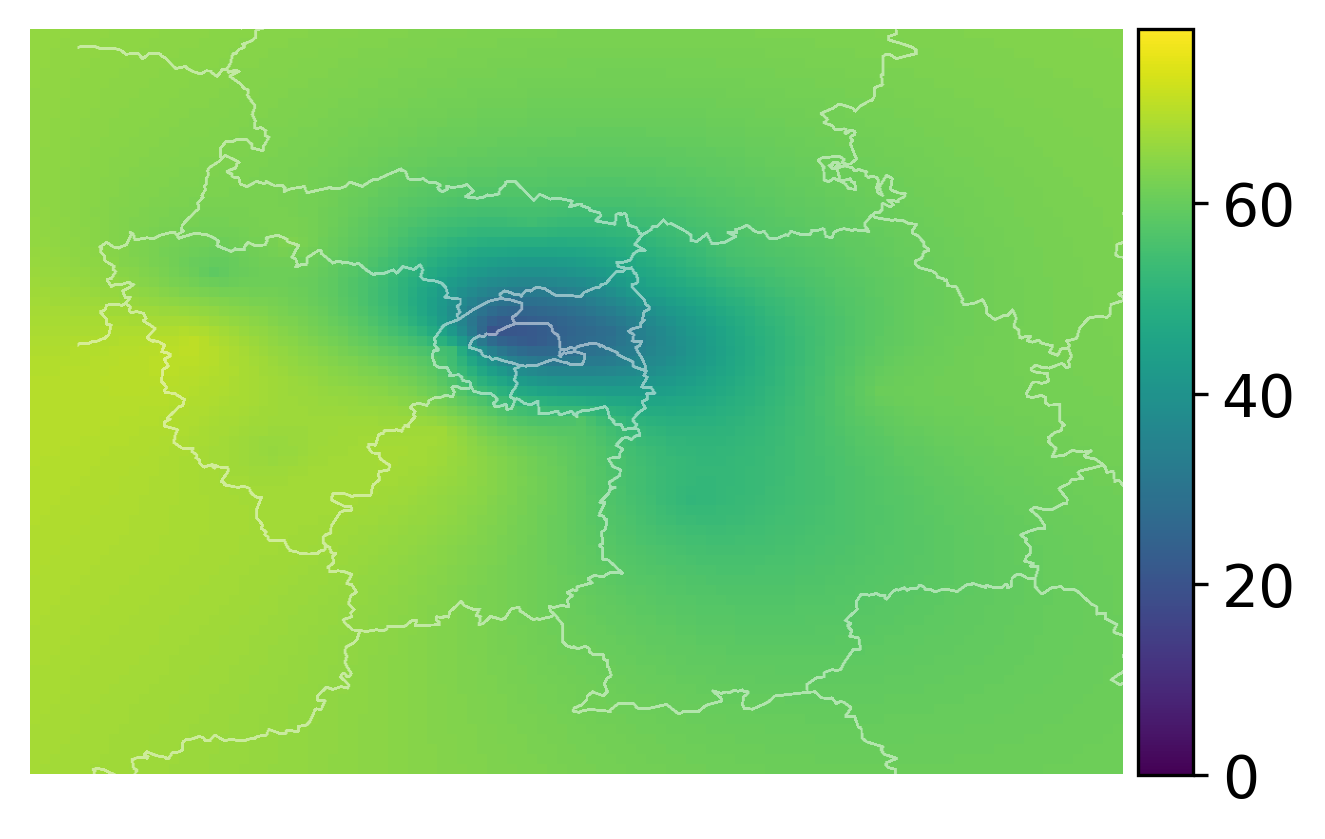}
            \caption*{Kriging}
        \end{subfigure}
        \begin{subfigure}{0.19\linewidth}
            \includegraphics[width=\linewidth]{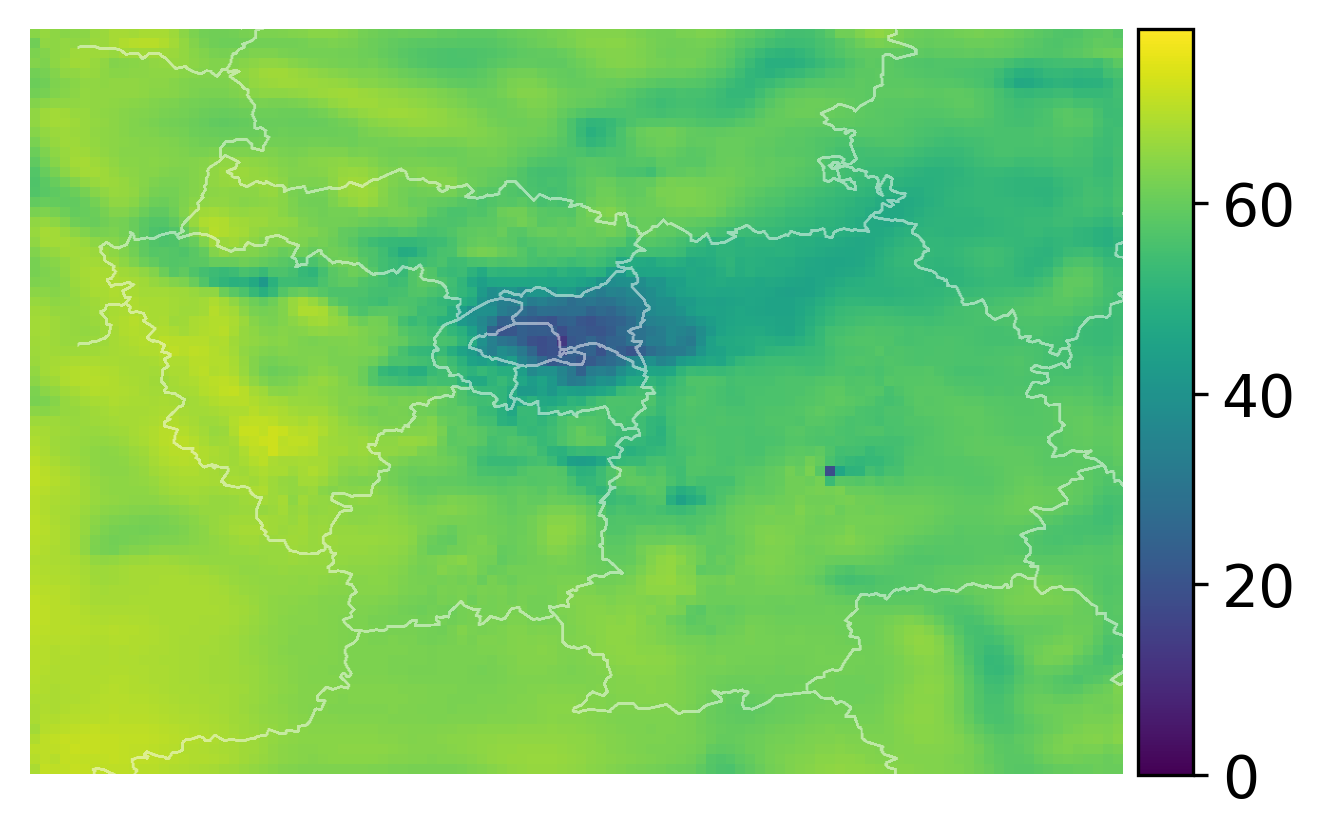}
            \caption*{VUNet}
        \end{subfigure}
        \begin{subfigure}{0.19\linewidth}
            \includegraphics[width=\linewidth]{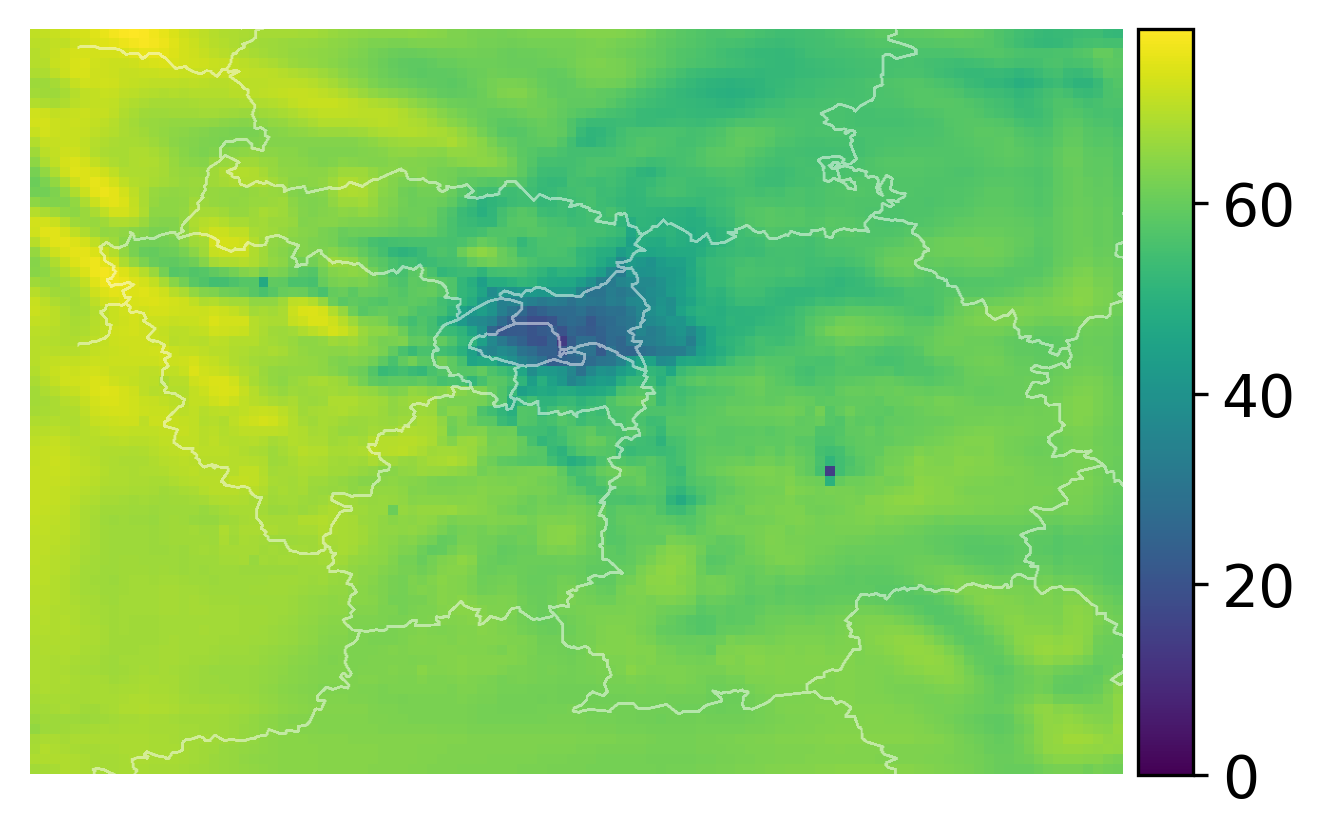}
            \caption*{CLSTM}
        \end{subfigure}
        \begin{subfigure}{0.19\linewidth}
            \includegraphics[width=\linewidth]{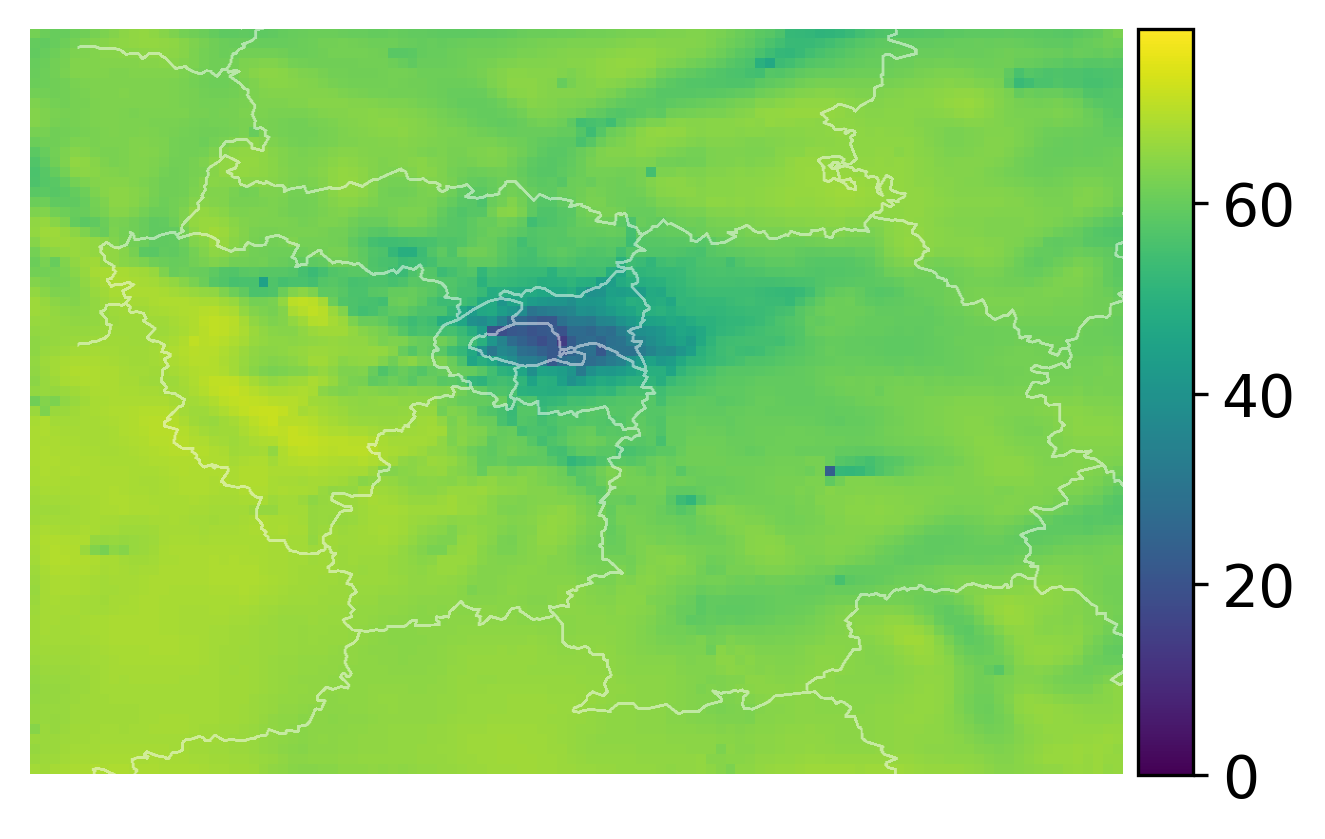}
            \caption*{Diffusion}
        \end{subfigure}
    \end{minipage}%
    }

    \makebox[\linewidth][c]{%
\begin{minipage}{1.4\textwidth}
    \centering
        \hspace{0.19\linewidth}
        \begin{subfigure}{0.19\linewidth}
            \includegraphics[width=\linewidth]{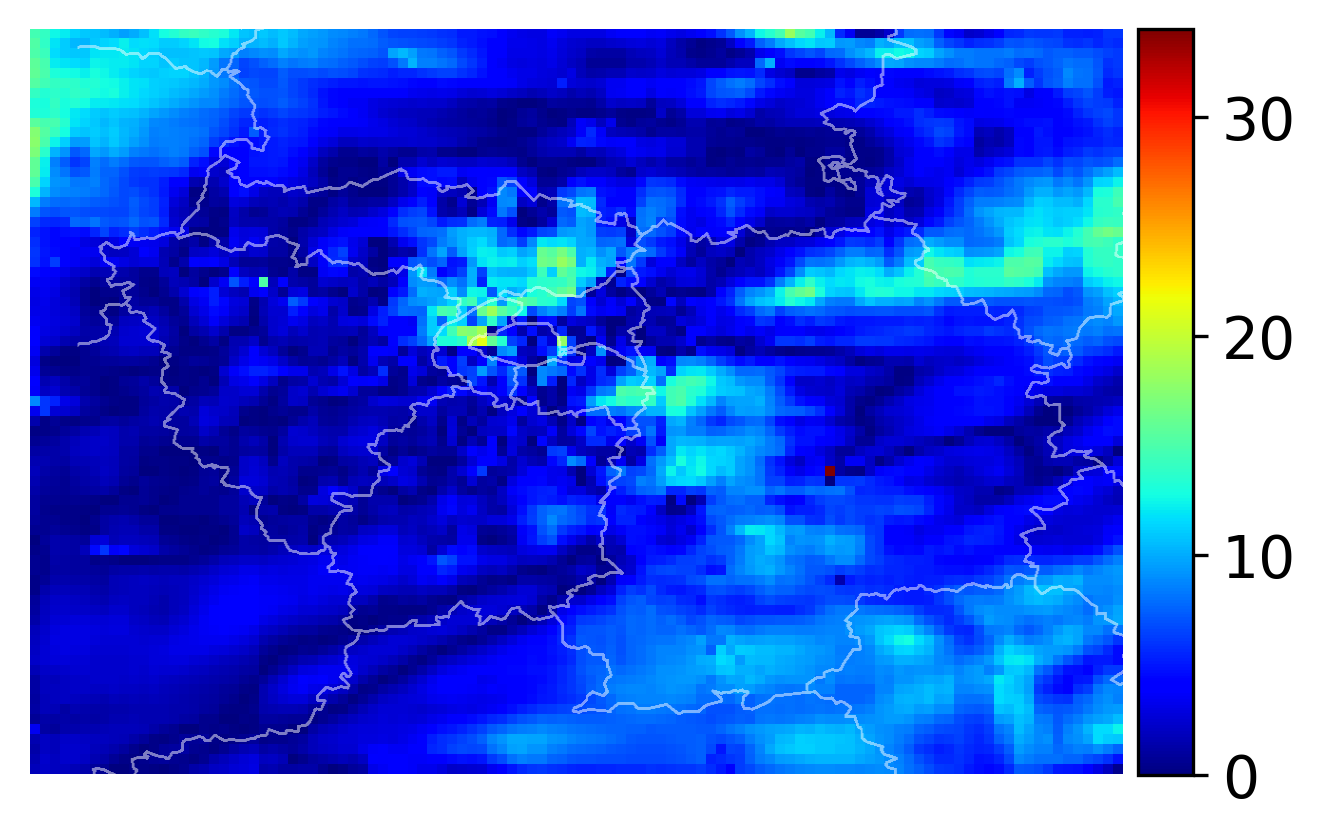}
            \caption*{Kriging Error}
        \end{subfigure}
        \begin{subfigure}{0.19\linewidth}
            \includegraphics[width=\linewidth]{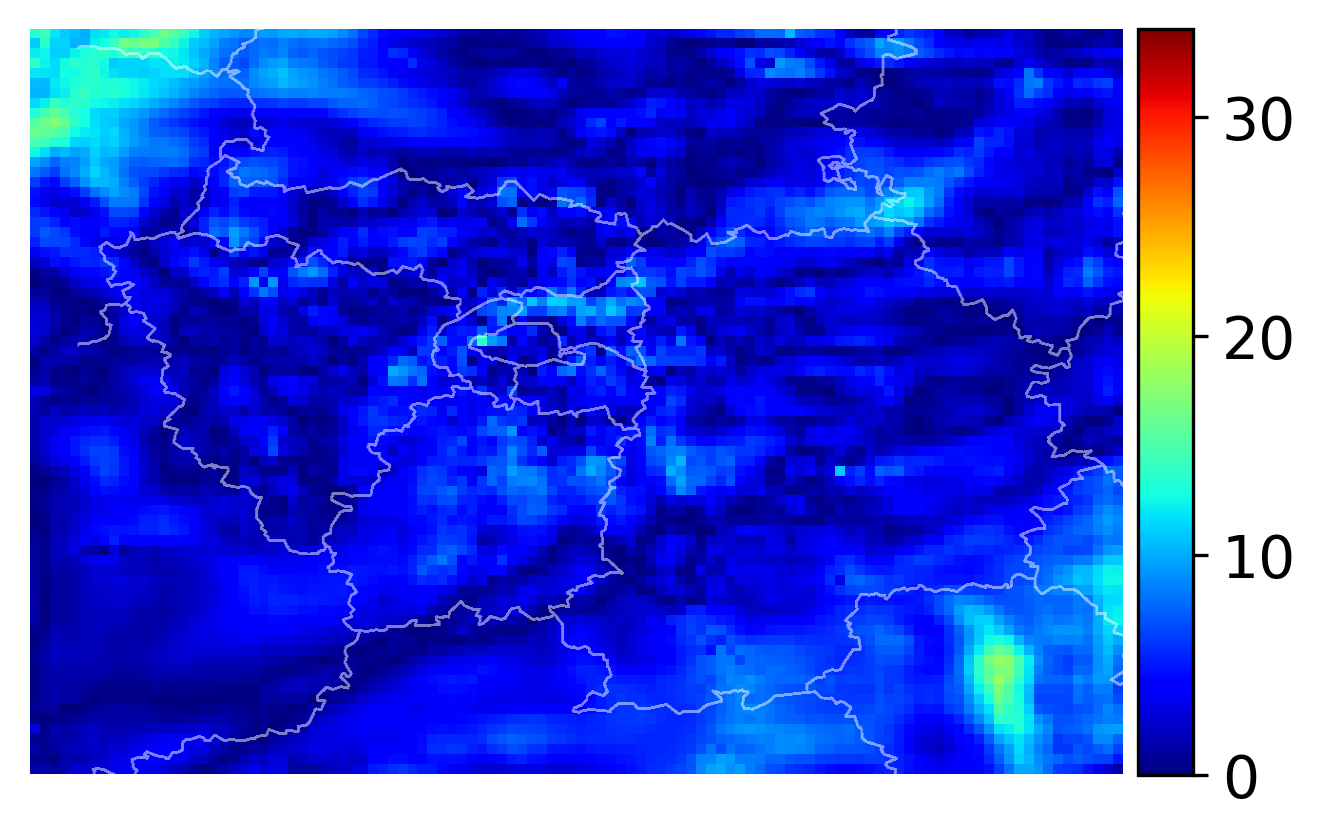}
            \caption*{VUNet Error}
        \end{subfigure}
        \begin{subfigure}{0.19\linewidth}
            \includegraphics[width=\linewidth]{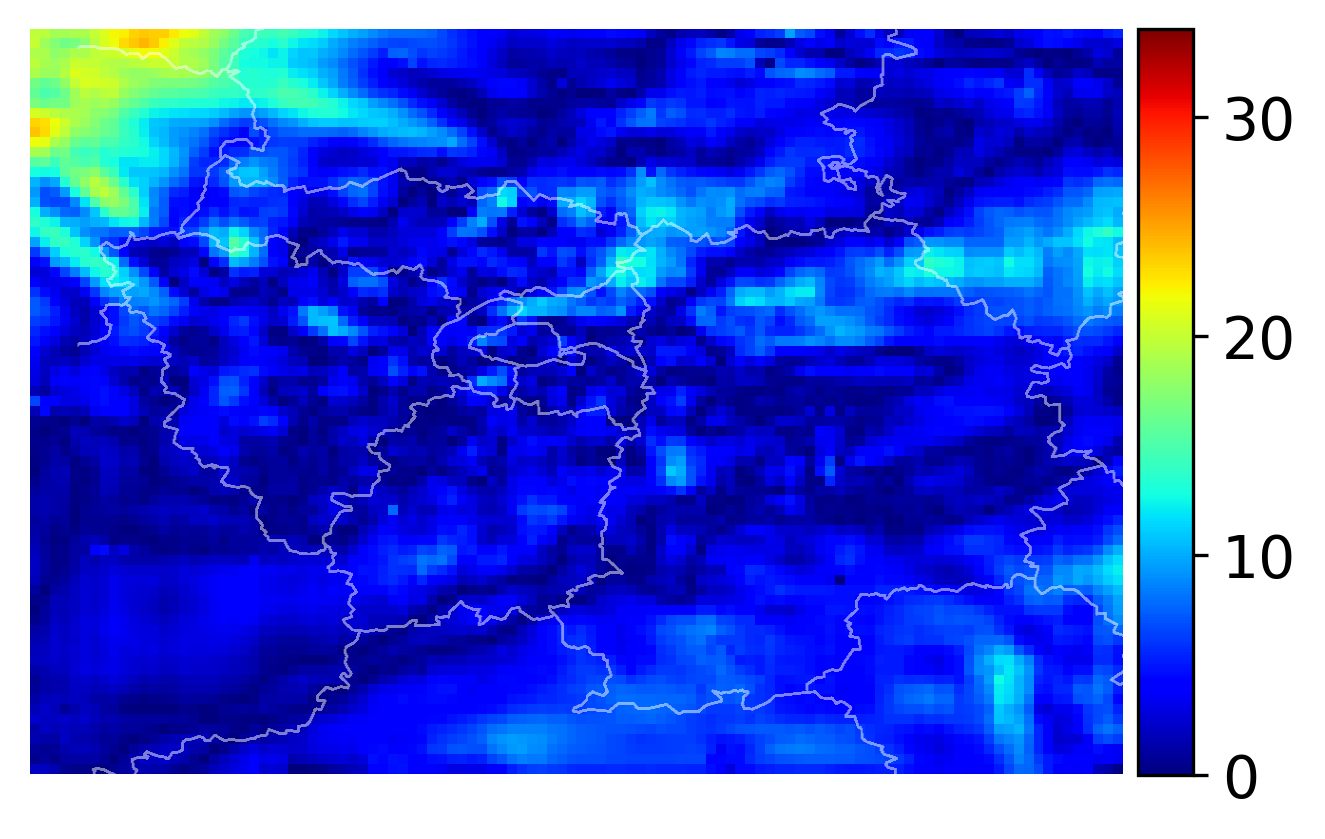}
            \caption*{CLSTM error}
        \end{subfigure}
        \begin{subfigure}{0.19\linewidth}
            \includegraphics[width=\linewidth]{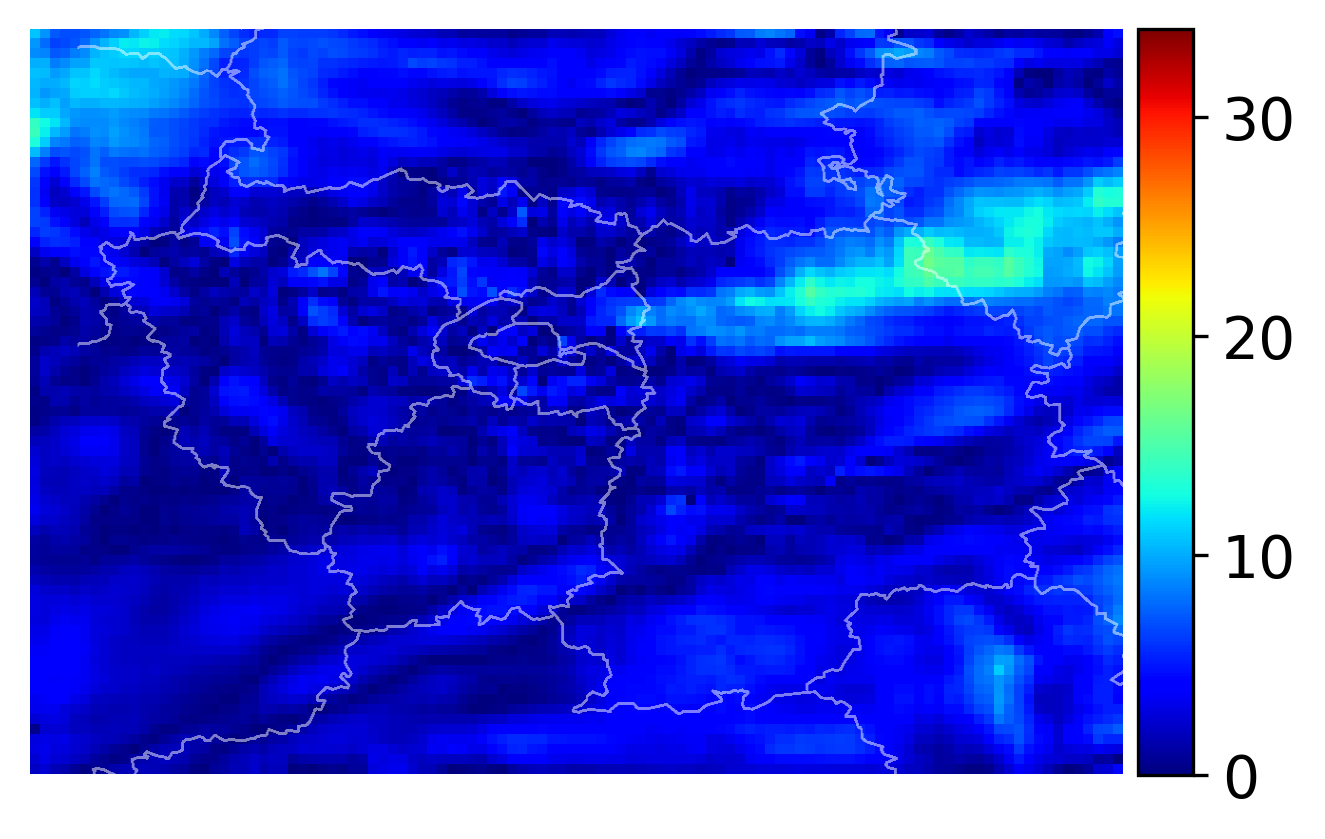}
            \caption*{Diffusion error}
        \end{subfigure}

    \end{minipage}%
    }

    \caption{\textbf{Prediction visualisation of each model on  O$_3$ simulation data.} Side-by-side comparison of the O$_3$ predictions at 8 AM on the 1st of December for Kriging, VUNet, CLSTM and the diffusion model}
    \label{fig:o3_comparison_random}
\end{figure}

\begin{figure}[htbp]
    \centering
    
    \makebox[\linewidth][c]{%
\begin{minipage}{1.4\textwidth}
        \centering
        \begin{subfigure}{0.19\linewidth}
            \includegraphics[width=\linewidth]{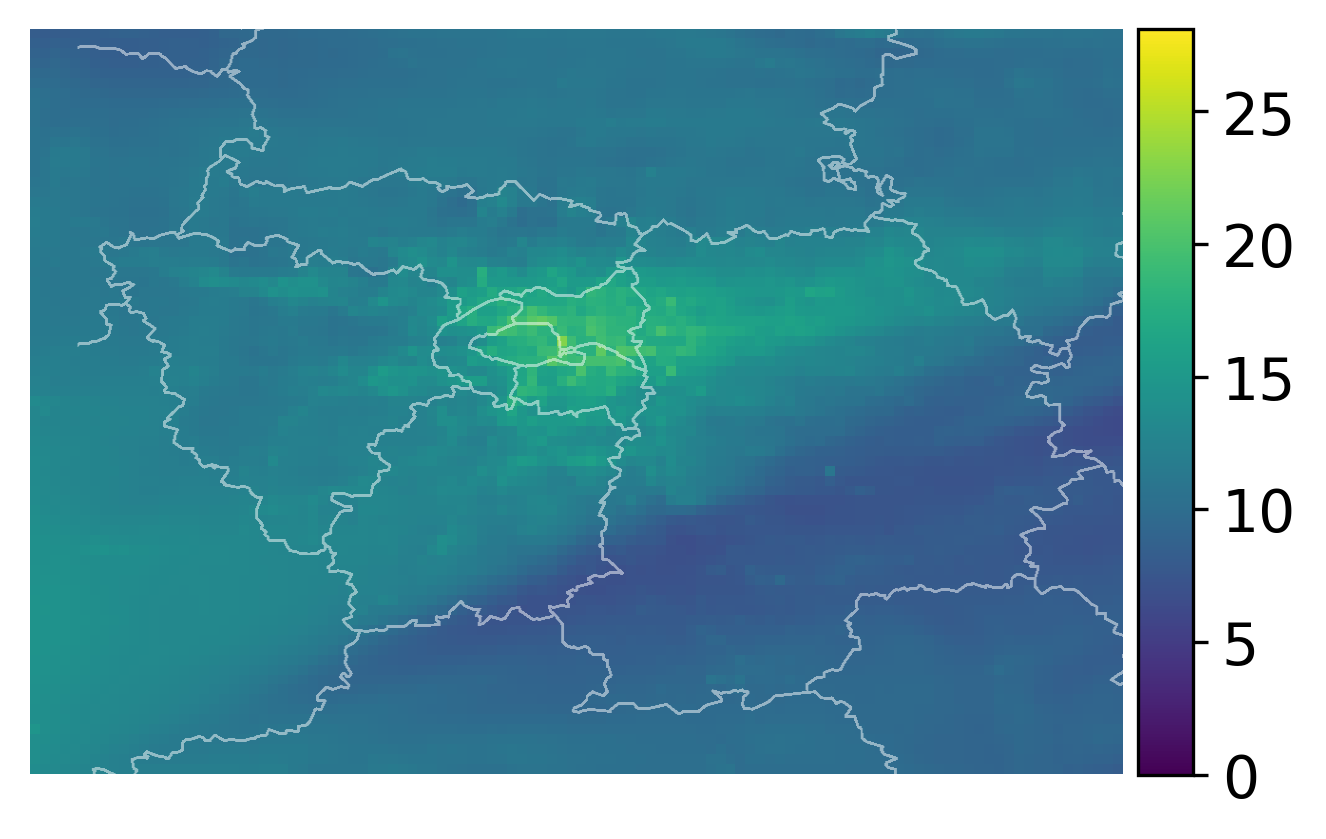}
            \caption*{PM$_{10}$ reference}
        \end{subfigure}
        \begin{subfigure}{0.19\linewidth}
            \includegraphics[width=\linewidth]{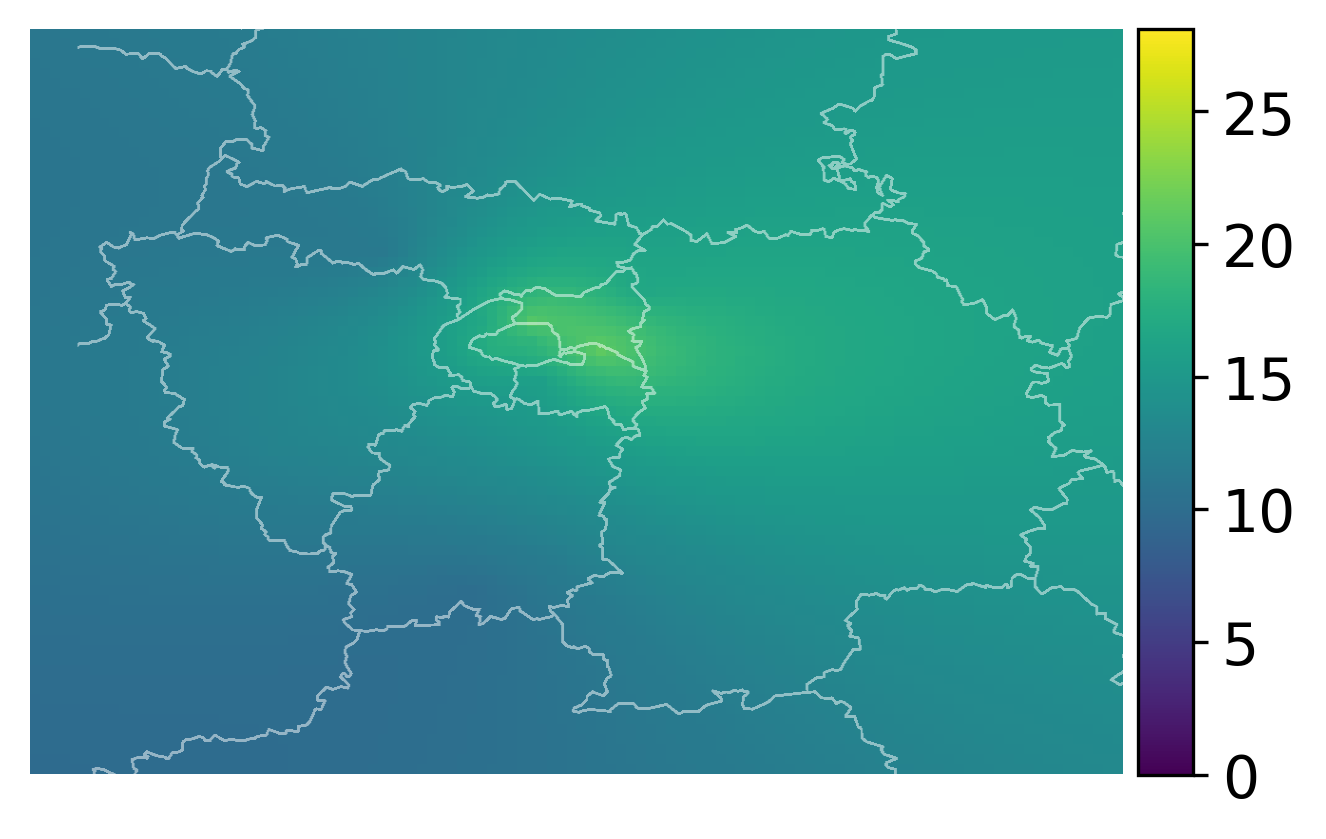}
            \caption*{Kriging}
        \end{subfigure}
        \begin{subfigure}{0.19\linewidth}
            \includegraphics[width=\linewidth]{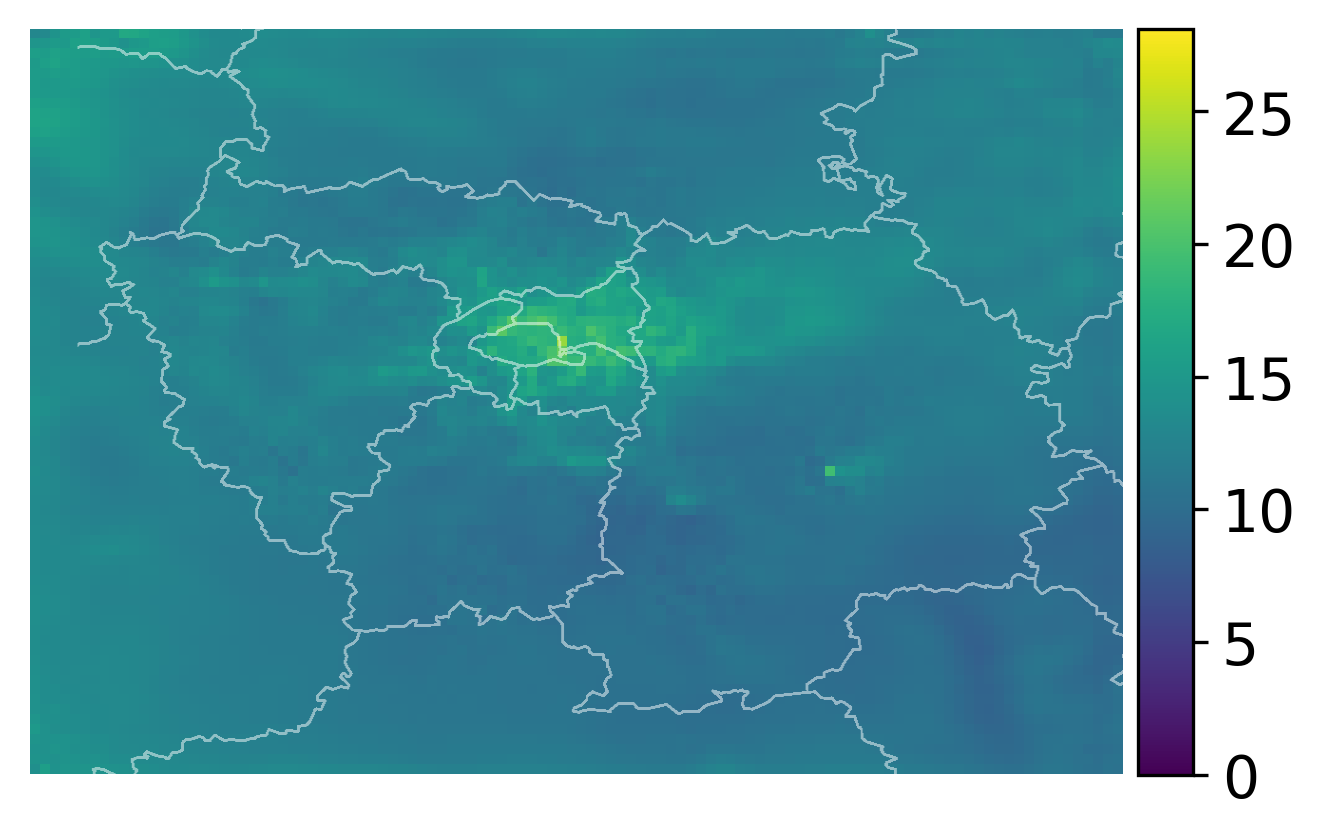}
            \caption*{VUNet}
        \end{subfigure}
        \begin{subfigure}{0.19\linewidth}
            \includegraphics[width=\linewidth]{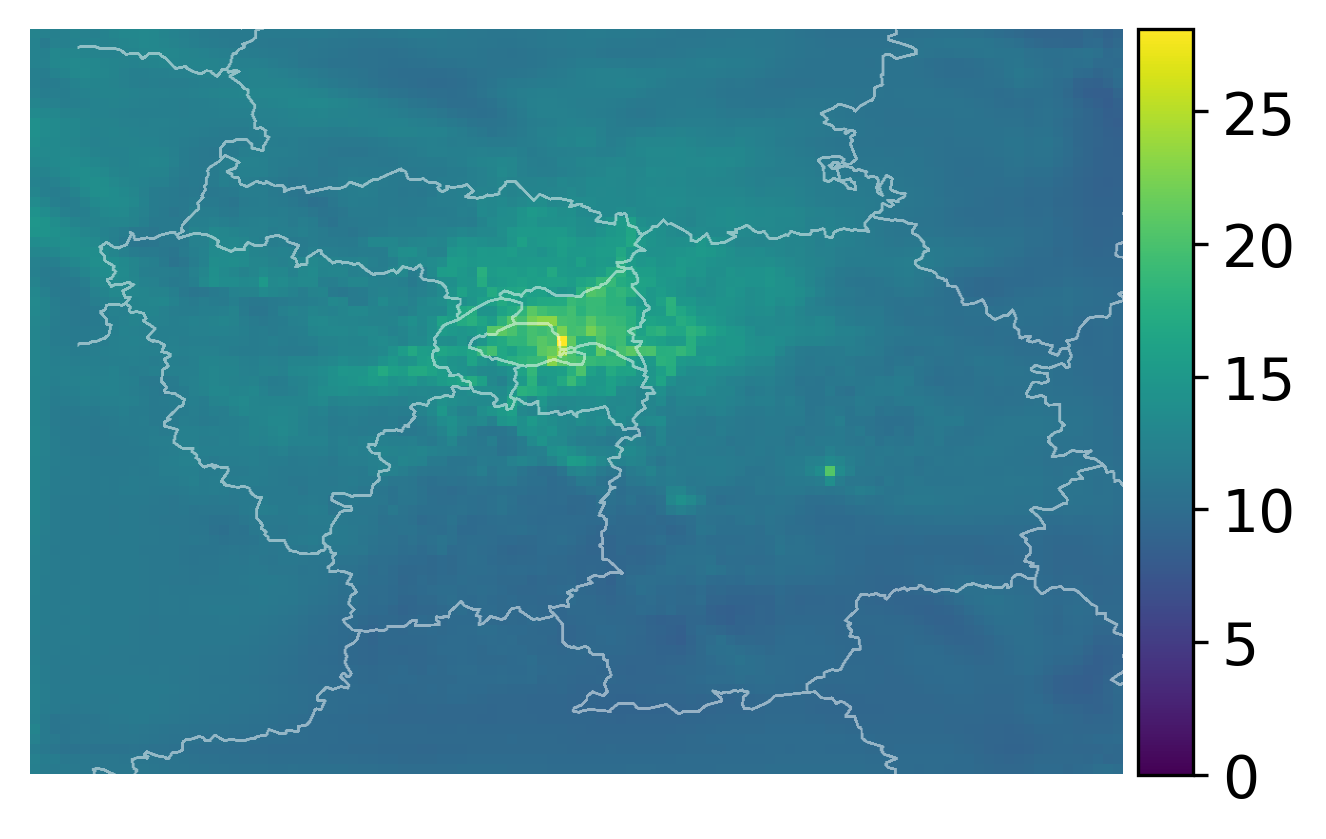}
            \caption*{CLSTM}
        \end{subfigure}
        \begin{subfigure}{0.19\linewidth}
            \includegraphics[width=\linewidth]{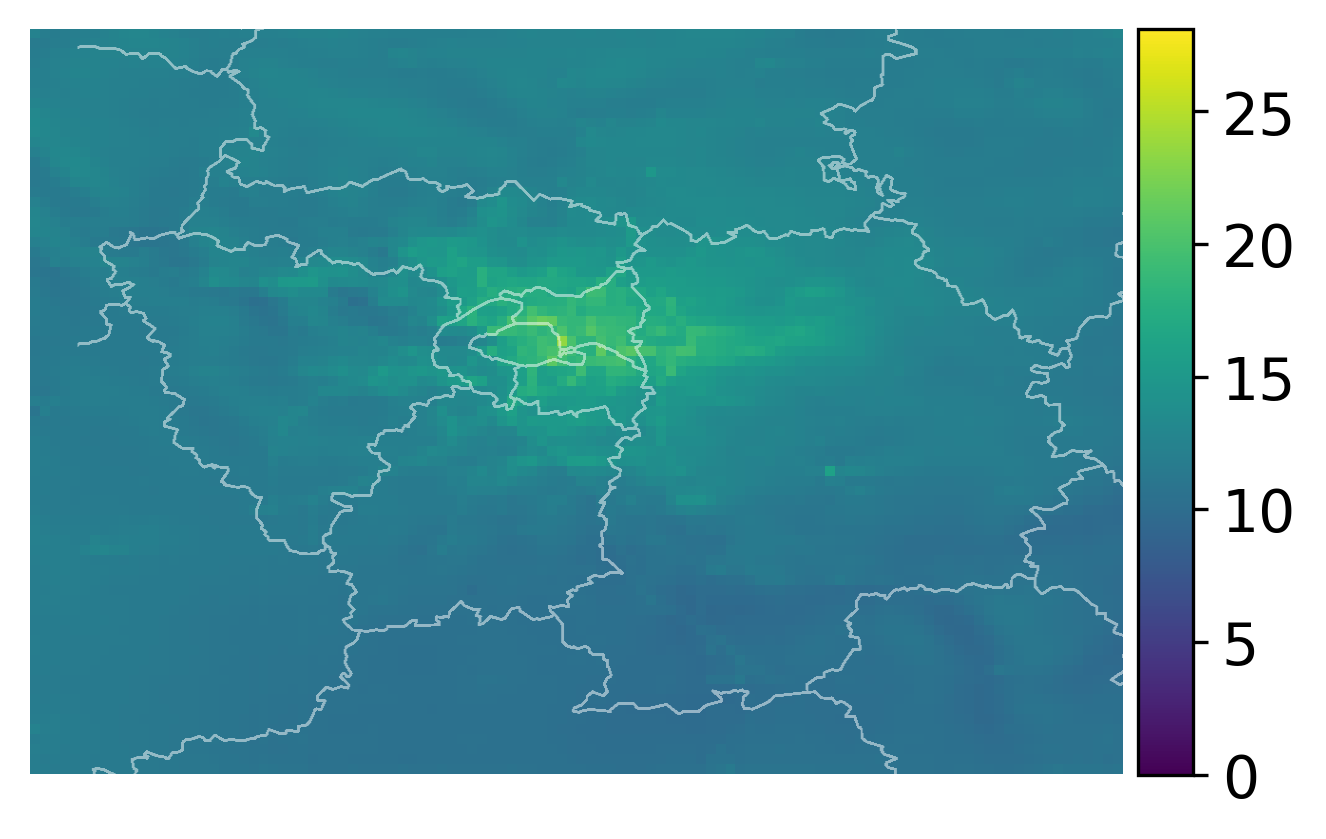}
            \caption*{Diffusion}
        \end{subfigure}
    \end{minipage}%
    }

    \makebox[\linewidth][c]{%
\begin{minipage}{1.4\textwidth}
    \centering
        \hspace{0.19\linewidth}
        \begin{subfigure}{0.19\linewidth}
            \includegraphics[width=\linewidth]{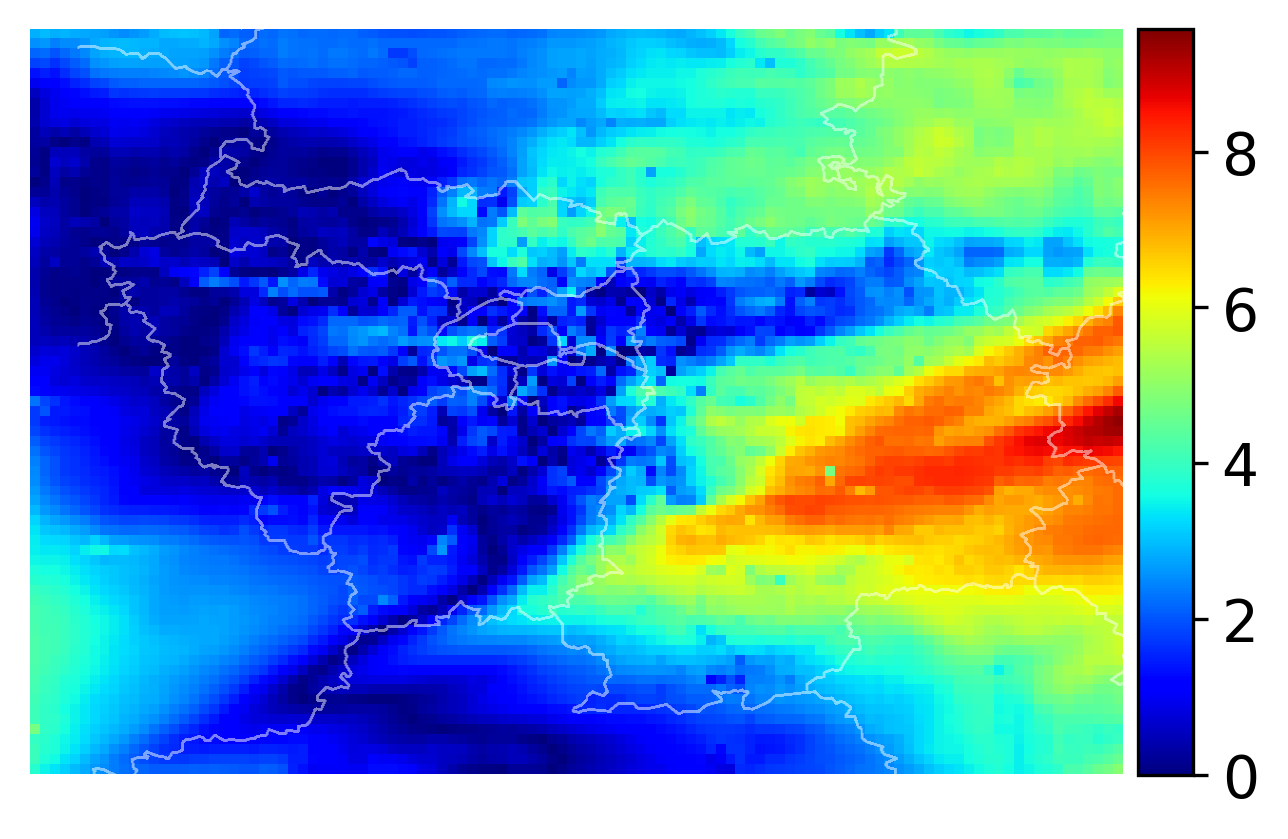}
            \caption*{Kriging Error}
        \end{subfigure}
        \begin{subfigure}{0.19\linewidth}
            \includegraphics[width=\linewidth]{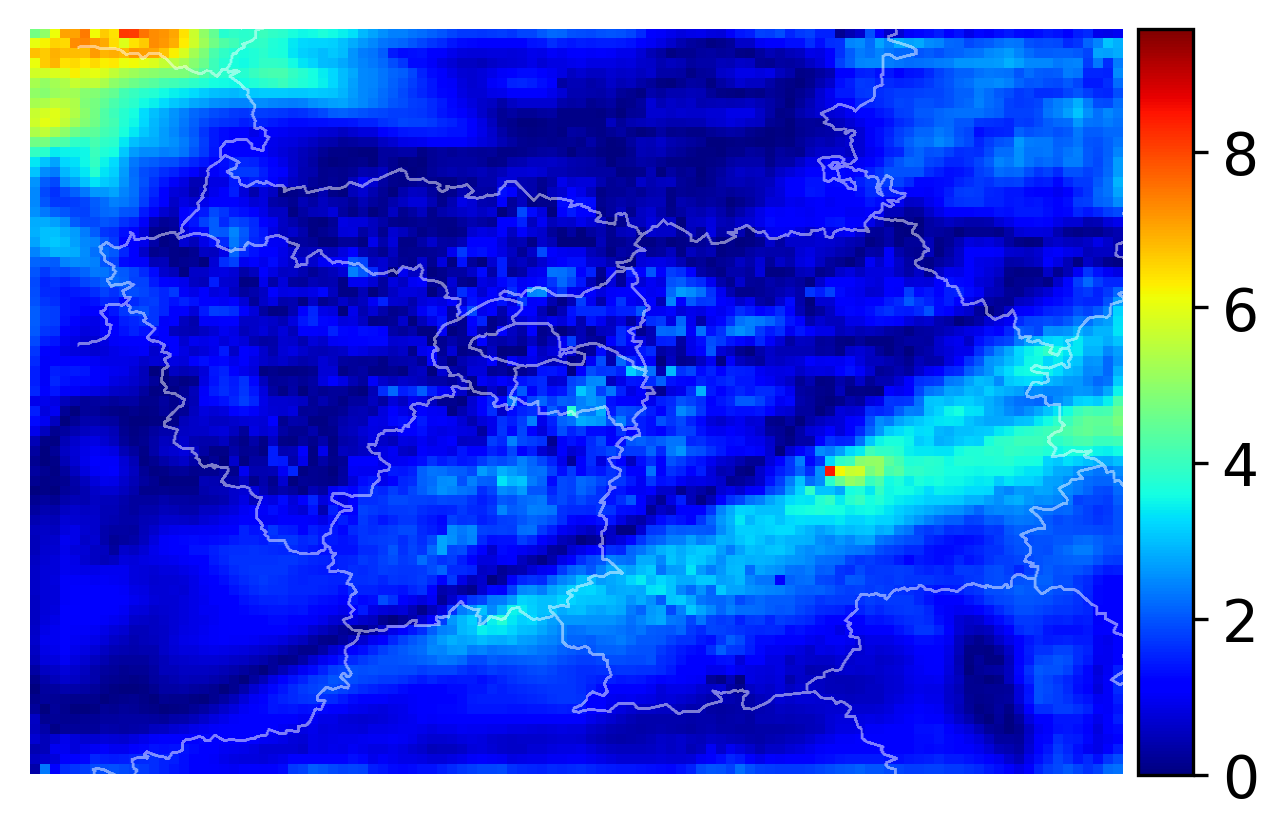}
            \caption*{VUNet Error}
        \end{subfigure}
        \begin{subfigure}{0.19\linewidth}
            \includegraphics[width=\linewidth]{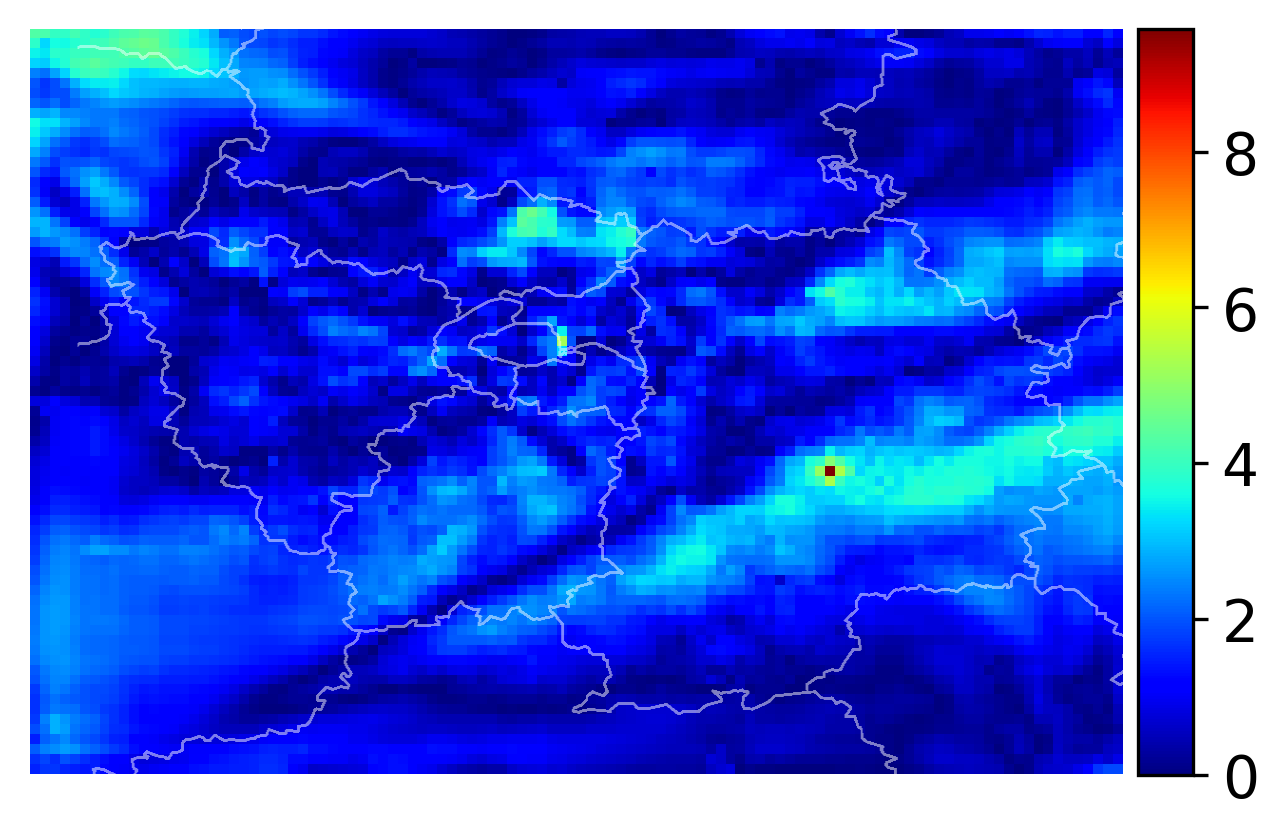}
            \caption*{CLSTM error}
        \end{subfigure}
        \begin{subfigure}{0.19\linewidth}
            \includegraphics[width=\linewidth]{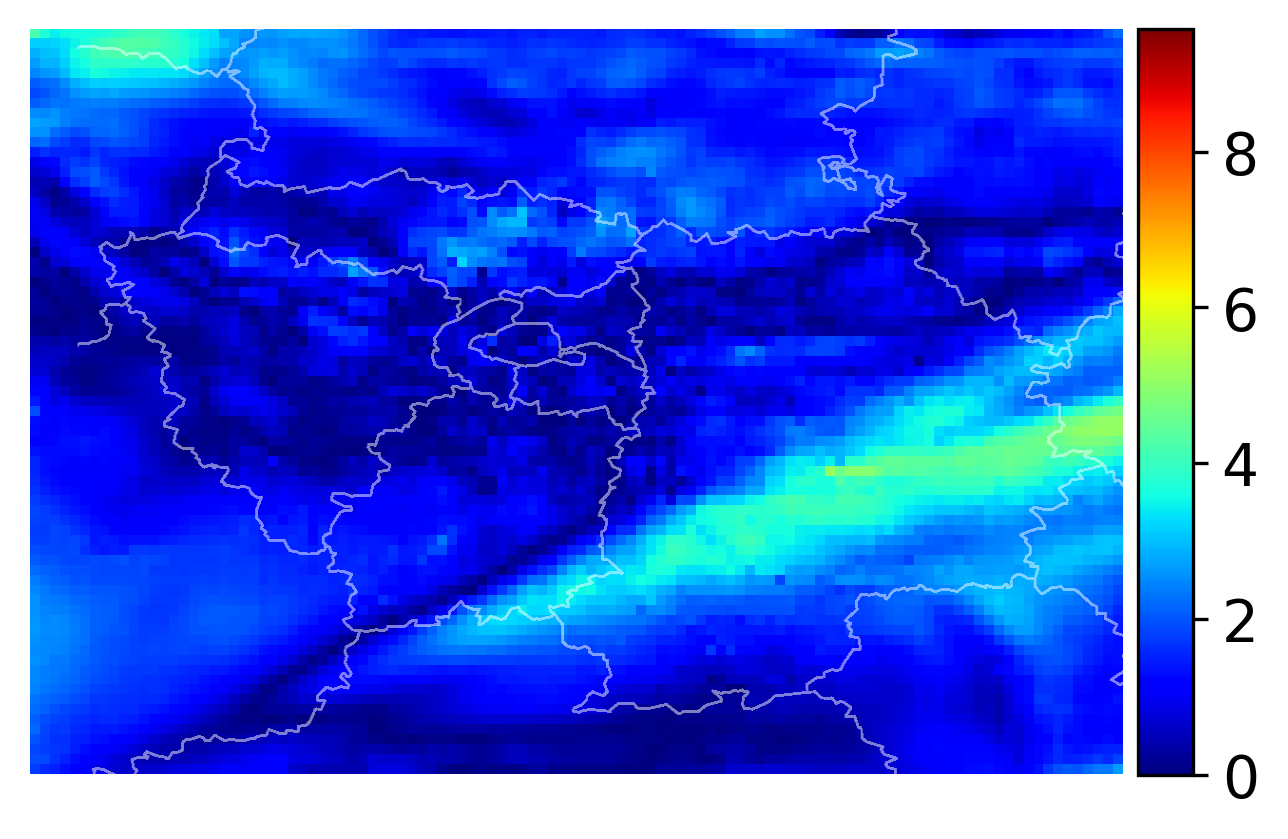}
            \caption*{Diffusion error}
        \end{subfigure}

    \end{minipage}%
    }

    \caption{\textbf{Prediction visualisation of each model on PM$_{10}$ simulation data.} Side-by-side comparison of the PM$_{10}$ predictions at 8 AM on the 1st of December for Kriging, VUNet, CLSTM and diffusion model}
    \label{fig:pm10_comparison_random}
\end{figure}

\begin{figure}[htbp]
    \centering
    
    \makebox[\linewidth][c]{%
\begin{minipage}{1.4\textwidth}
        \centering
        \begin{subfigure}{0.19\linewidth}
            \includegraphics[width=\linewidth]{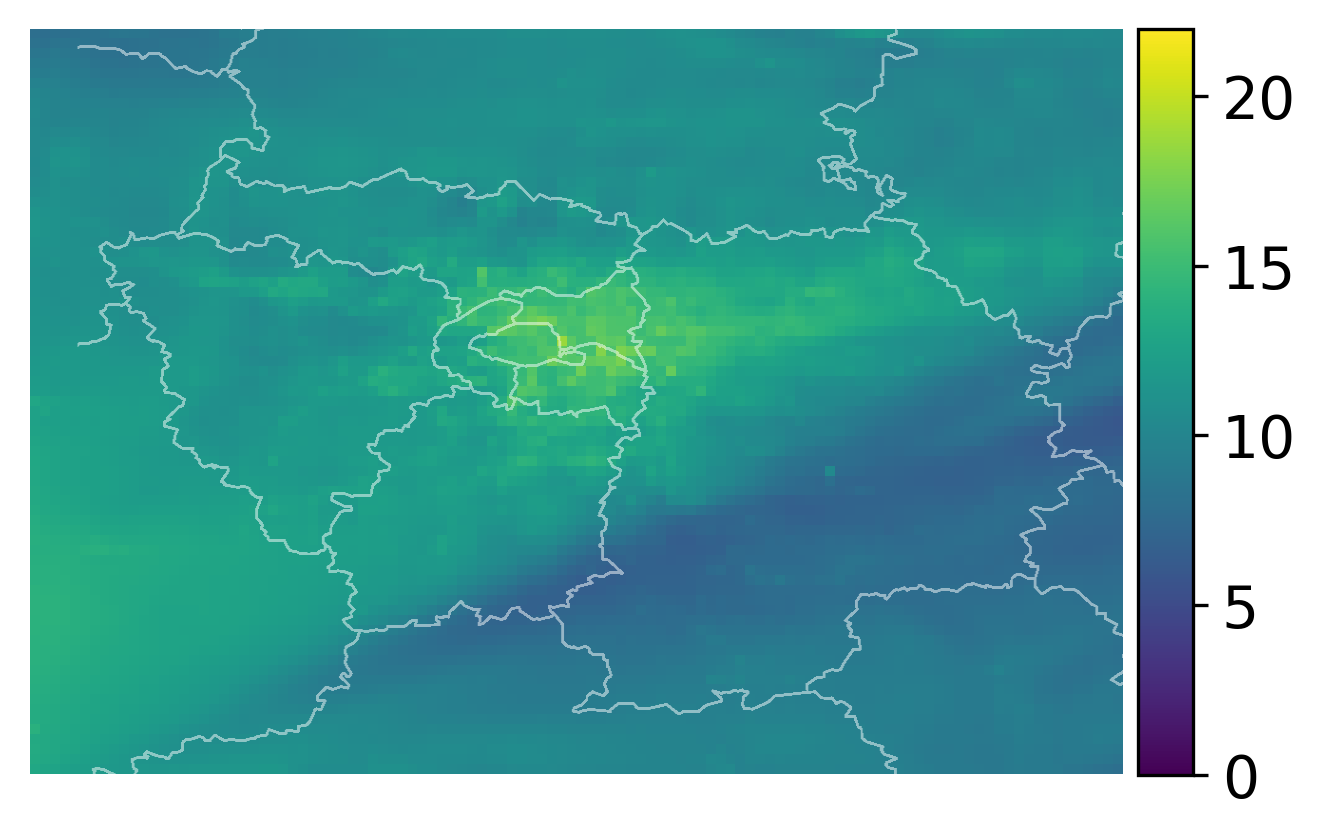}
            \caption*{PM$_{2.5}$ reference}
        \end{subfigure}
        \begin{subfigure}{0.19\linewidth}
            \includegraphics[width=\linewidth]{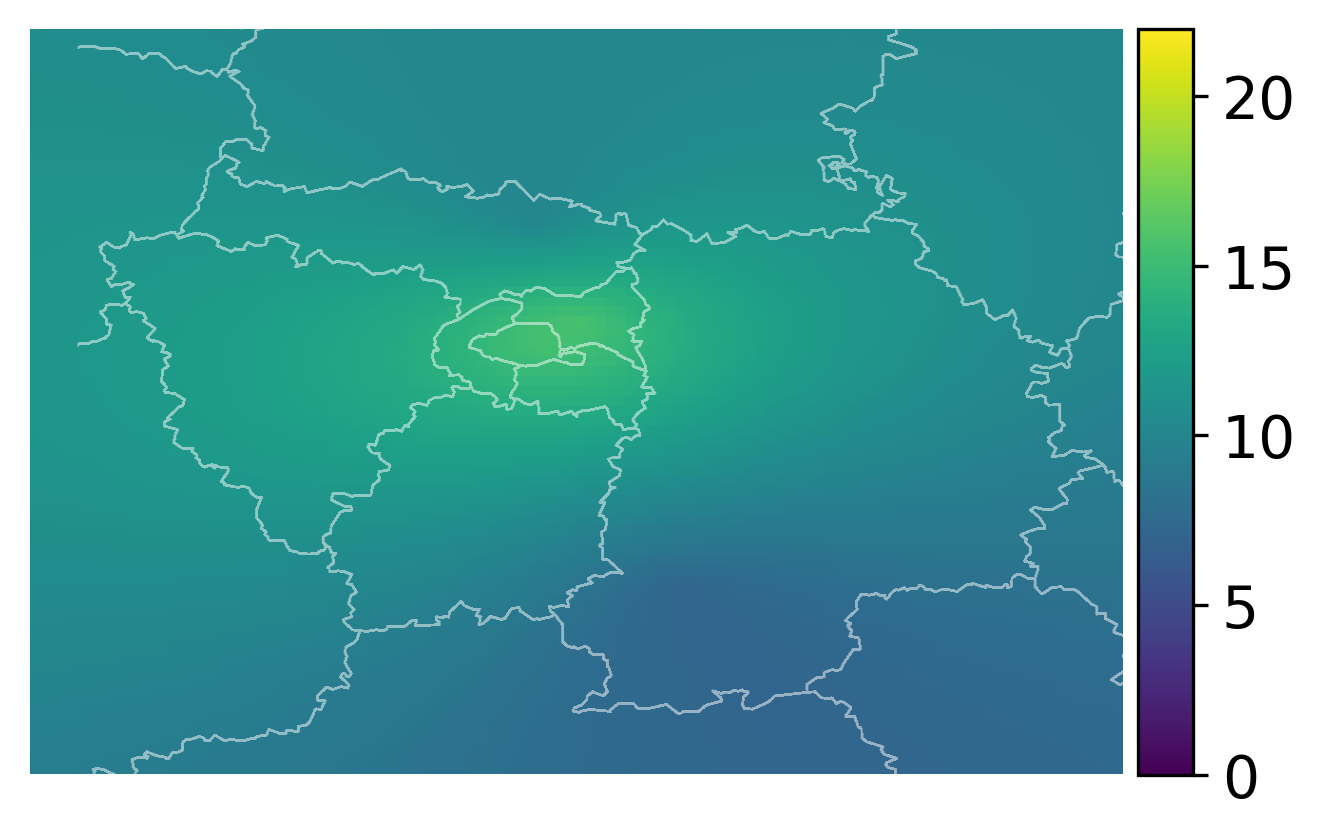}
            \caption*{Kriging}
        \end{subfigure}
        \begin{subfigure}{0.19\linewidth}
            \includegraphics[width=\linewidth]{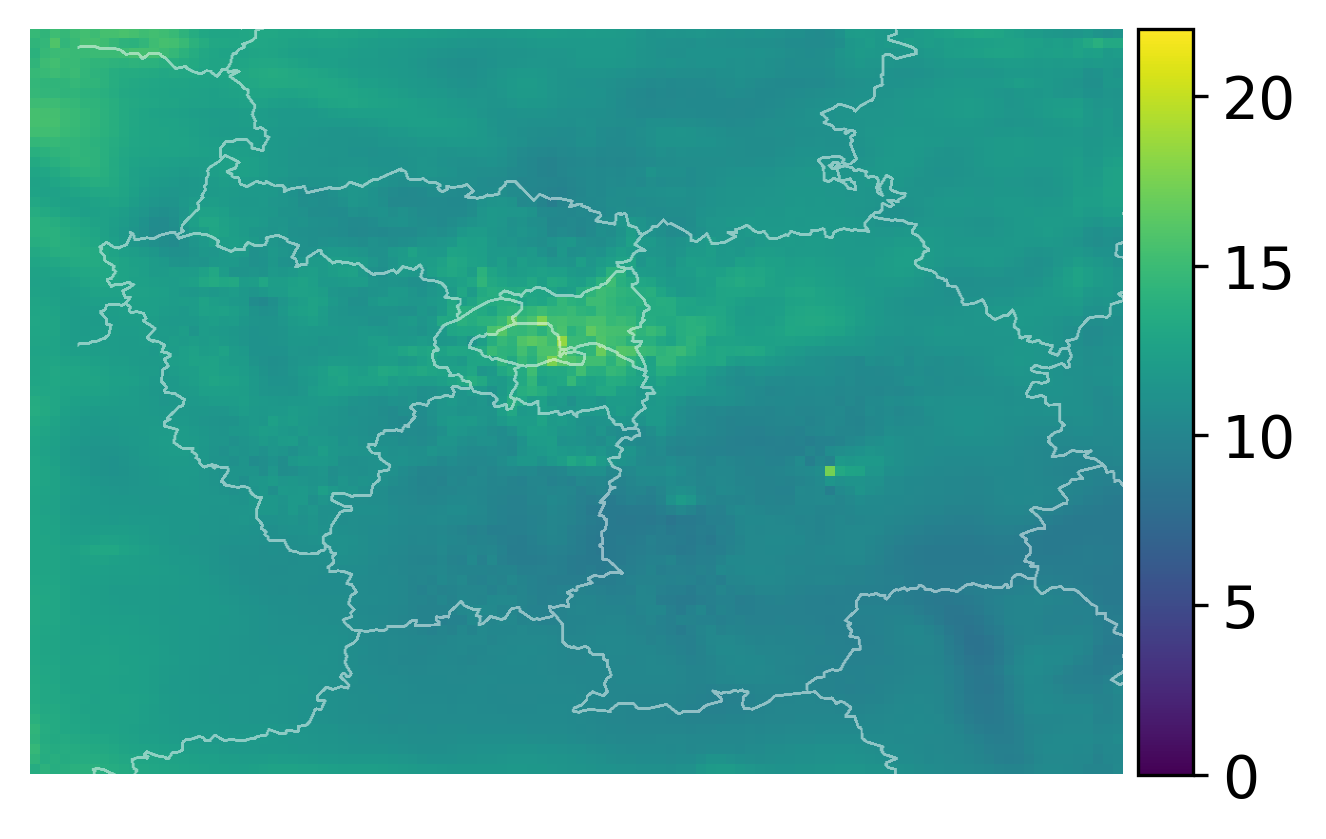}
            \caption*{VUNet}
        \end{subfigure}
        \begin{subfigure}{0.19\linewidth}
            \includegraphics[width=\linewidth]{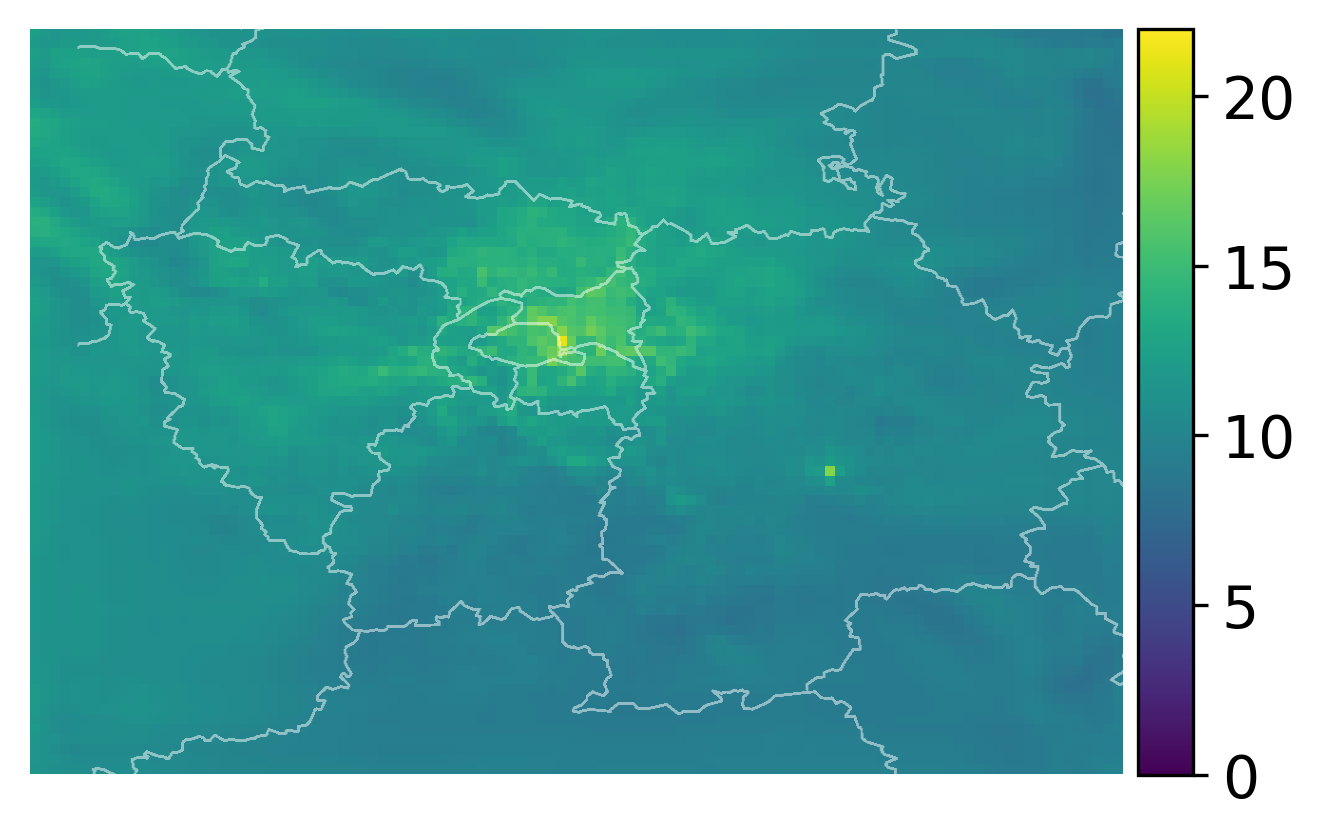}
            \caption*{CLSTM}
        \end{subfigure}
        \begin{subfigure}{0.19\linewidth}
            \includegraphics[width=\linewidth]{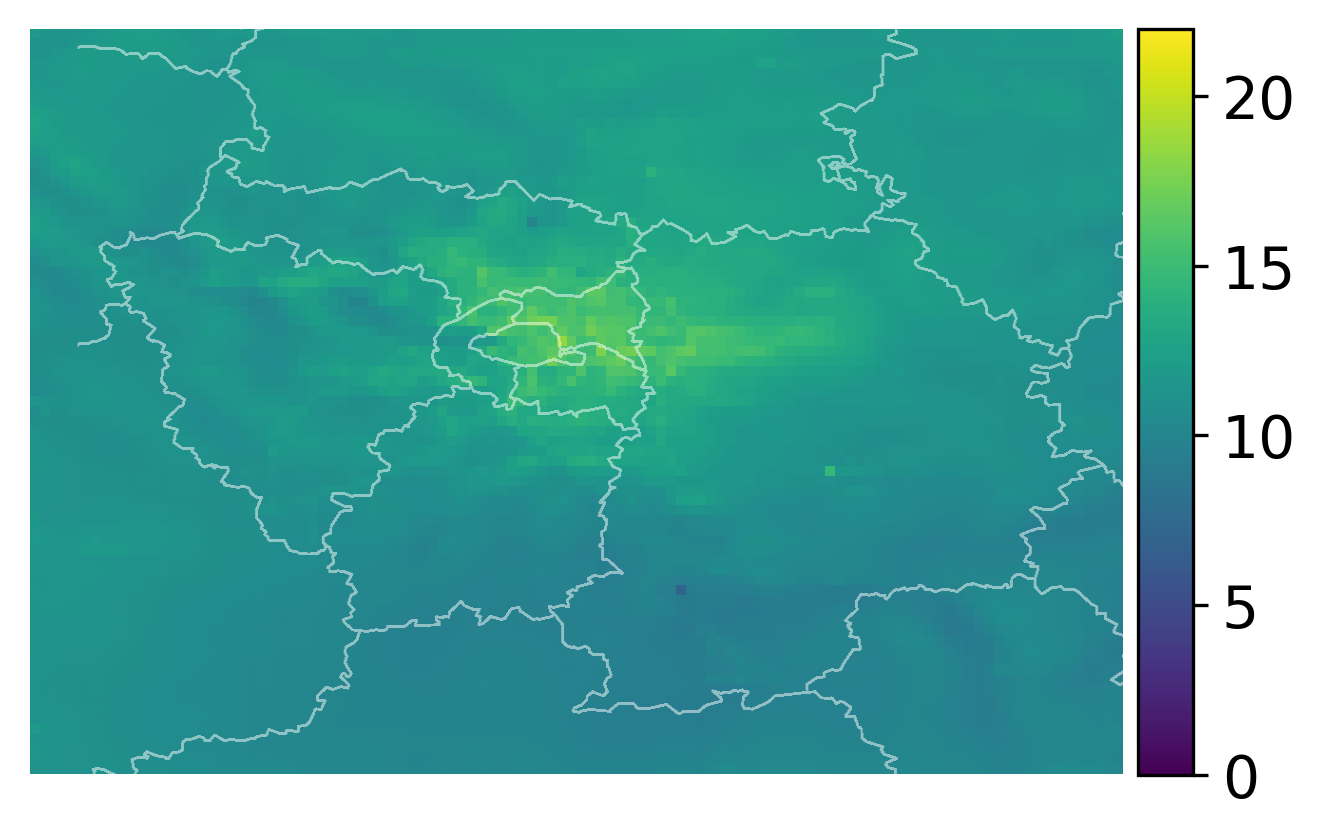}
            \caption*{Diffusion}
        \end{subfigure}
    \end{minipage}%
    }

    \makebox[\linewidth][c]{%
\begin{minipage}{1.4\textwidth}
    \centering
        \hspace{0.19\linewidth}
        \begin{subfigure}{0.19\linewidth}
            \includegraphics[width=\linewidth]{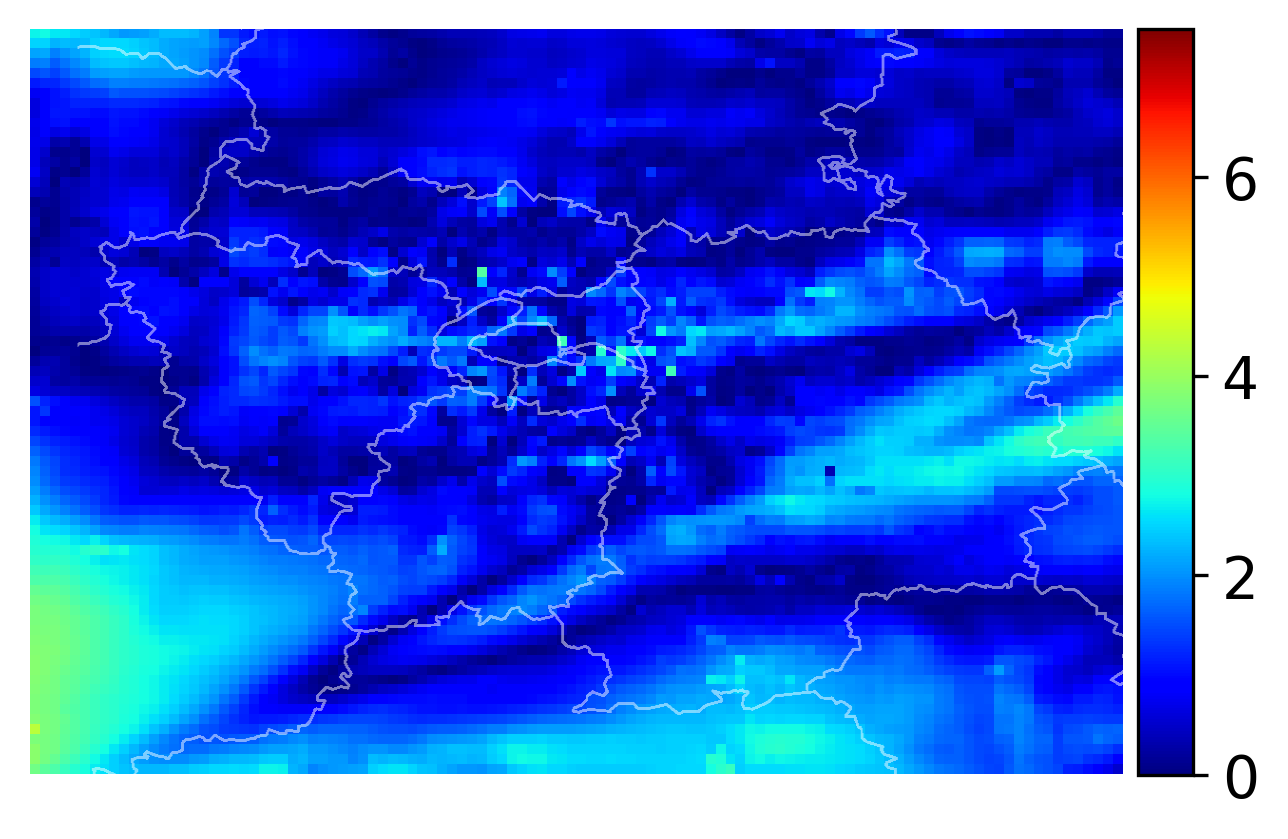}
            \caption*{Kriging Error}
        \end{subfigure}
        \begin{subfigure}{0.19\linewidth}
            \includegraphics[width=\linewidth]{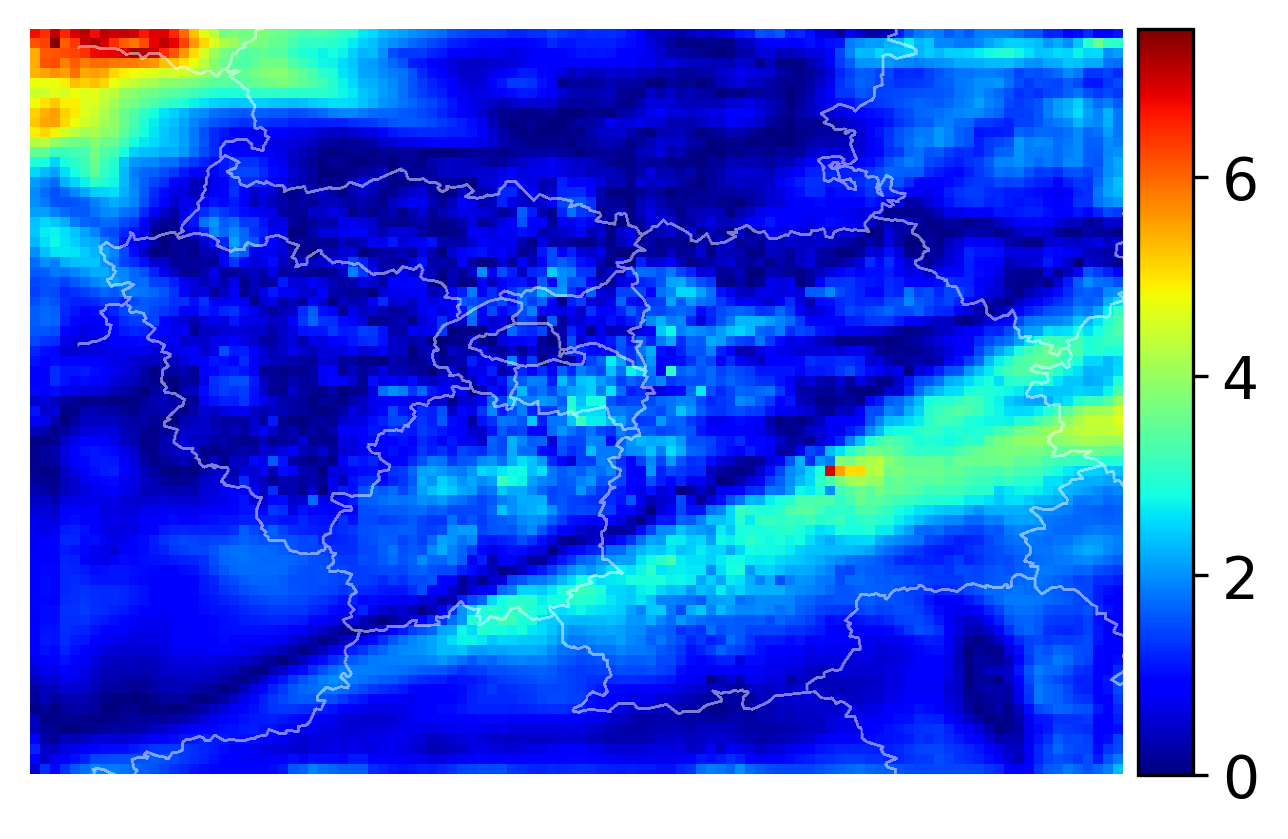}
            \caption*{VUNet Error}
        \end{subfigure}
        \begin{subfigure}{0.19\linewidth}
            \includegraphics[width=\linewidth]{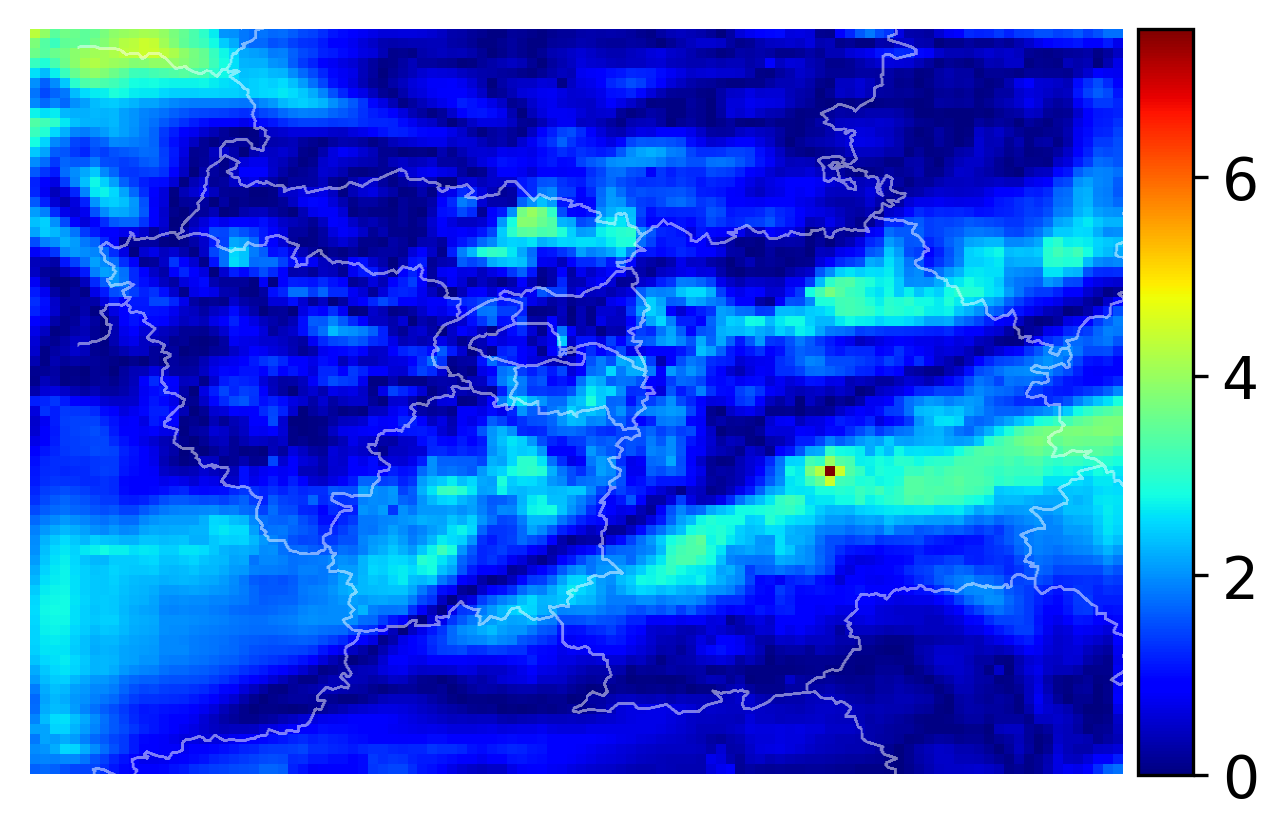}
            \caption*{CLSTM error}
        \end{subfigure}
        \begin{subfigure}{0.19\linewidth}
            \includegraphics[width=\linewidth]{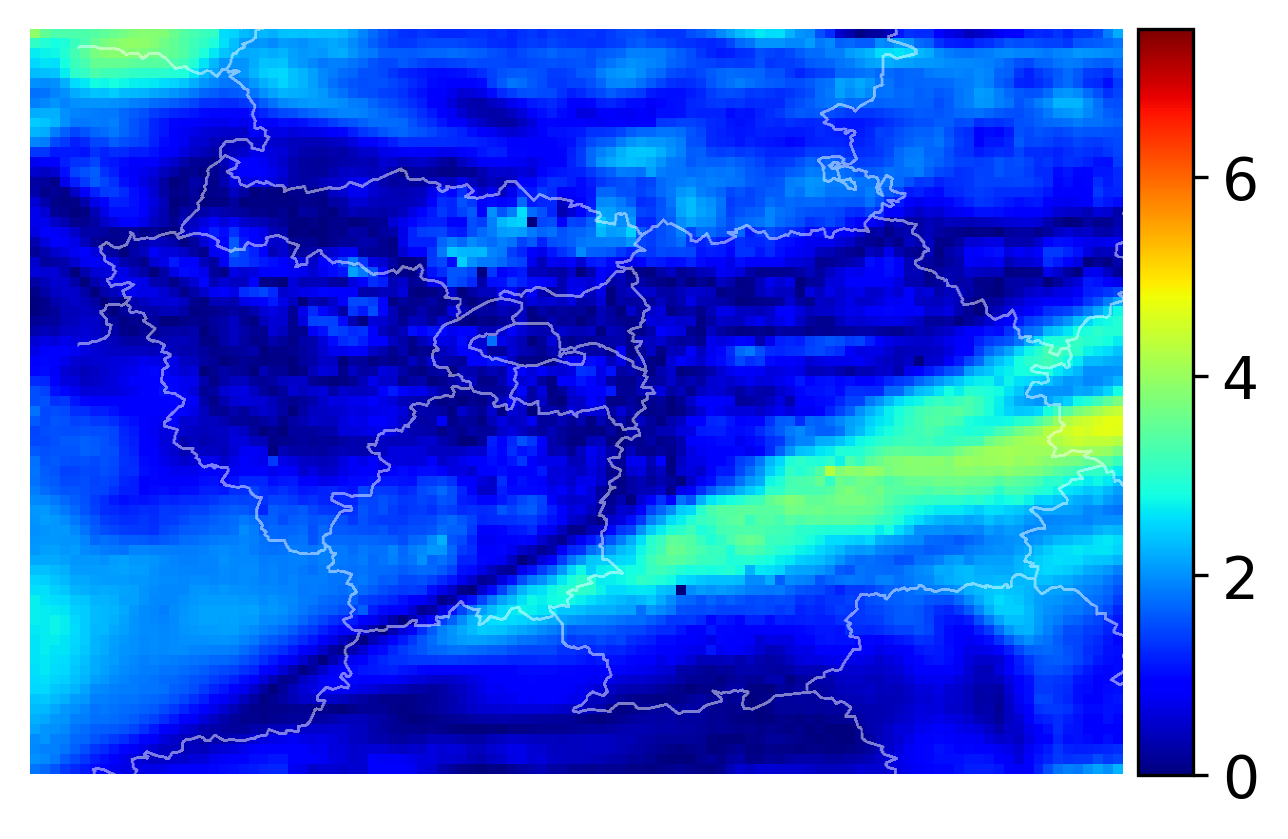}
            \caption*{Diffusion error}
        \end{subfigure}

    \end{minipage}%
    }

    \caption{\textbf{Prediction visualisation of each model on PM$_{2.5}$ simulation data.} Side-by-side comparison of the PM$_{2.5}$ predictions at 8 AM on the 1st of December for Kriging, VUNet, CLSTM and diffusion model}
    \label{fig:pm25_comparison_random}
\end{figure}

Figures \ref{fig:no2_comparison_random}, \ref{fig:o3_comparison_random}, \ref{fig:pm10_comparison_random}, \ref{fig:pm25_comparison_random} compare the predicted output of Kriging, VUNet, CLSTM and the diffusion model against the reference simulation data point from 1st December at 8 AM.

\vfill

\vfill

\vfill

\clearpage
\section{\an{Uncertainty and Statistical analysis}}

\subsection{Confidence interval}\label{ci}
\an{We quantify the uncertainty of the L2 relative error of each model using a temporal block bootstrap. This analysis (Table \ref{tab:bootstrap_ci_results}) shows the robustness of the model performance by accounting for the temporal dependence of the observed data.}

\an{In the test period of December, for each model, the hourly predictions were resampled using overlapping blocks of 24 consecutive hours, corresponding to a one-day temporal block. A total of 10,000 bootstrap samples were generated by randomly selecting blocks with replacement and concatenating them until the original sample length was recovered. The mean relative L2 error was then calculated for each bootstrap sample, producing an empirical distribution of the mean error. The $95\%$ confidence interval (CI) was obtained from the $2.5$th and $97.5$th percentiles of this bootstrap distribution. Thus, the resulting confidence intervals provide an estimate of the uncertainty in the mean error while accounting for temporal dependence rather than treating all hourly observations as independent. Table \ref{tab:bootstrap_ci_results} shows that the relatively narrow intervals indicate low uncertainty in the estimated error.}

\ans{To directly compare the model performance on observation data, we perform a paired temporal block bootstrap analysis of the mean relative L2 error by time matching the temporal blocks. The primary analysis used 24-hour block and an additional sensitivity analysis uses 48, 72, and 168 hours blocks to evaluate the effect of temporal block length. Table \ref{tab:paired_block_sensitivity} reports the difference in mean relative L2 error between each pair of models, with the model having higher error listed first. Values in brackets denote the 95\% confidence interval. Across all block lengths, the mean pairwise difference remains unchanged, and the confidence interval remains consistently positive. This indicates that the analysis remains robust to the choice of block length. In particular, CLSTM consistently achieves lower error than other models, while diffusion remains better than VUNet and ViTAE. }

\begin{table}[htbp]
\centering
\begin{adjustbox}{max width=\linewidth,center}
\begin{tabular}{llrrr}
\toprule
Dataset & Model & relative L2 error & 95\% CI Lower & 95\% CI Upper \\
\midrule
\multirow{4}{*}{Simulation data}
& CLSTM(t=8) & 0.1125 & 0.0946 & 0.1322 \\
& VUNet(t=4) & 0.1090 & 0.0911 & 0.1270 \\
& ViTAE(t=6) & 0.1299 & 0.1090 & 0.1497 \\
& Diffusion(E=20) & 0.1093 & 0.0900 & 0.1281 \\
\midrule
\multirow{4}{*}{Observation data}
& CLSTM(t=8) & 0.2307 & 0.2152 & 0.2480 \\
& VUNet(t=4) & 0.2671 & 0.2453 & 0.2844 \\
& ViTAE(t=6) & 0.3175 & 0.2924 & 0.3401 \\
& Diffusion(E=20) & 0.2544 & 0.2370 & 0.2712 \\
\bottomrule
\end{tabular}
\end{adjustbox}
\caption{\an{Mean L2 relative error and 95\% confidence intervals estimated using temporal block bootstrap with 24-hour blocks and 10,000 bootstraps on the simulation and observation data.}}
\label{tab:bootstrap_ci_results}
\end{table}

\begin{table}[htbp]

\begin{adjustbox}{max width=\linewidth,center}
\begin{tabular}{lcccc}
\toprule
Comparison & 24 h & 48 h & 72 h & 168 h \\
\midrule
VUNet -- CLSTM
& $0.0357\;[0.0196,0.0483]$
& $0.0357\;[0.0154,0.0492]$
& $0.0357\;[0.0135,0.0504]$
& $0.0357\;[0.0123,0.0439]$ \\

ViTAE -- CLSTM
& $0.0862\;[0.0666,0.1043]$
& $0.0862\;[0.0613,0.1068]$
& $0.0862\;[0.0592,0.1062]$
& $0.0862\;[0.0577,0.1001]$ \\

Diffusion -- CLSTM
& $0.0237\;[0.0139,0.0321]$
& $0.0237\;[0.0112,0.0337]$
& $0.0237\;[0.0097,0.0339]$
& $0.0237\;[0.0090,0.0304]$ \\

ViTAE -- VUNet
& $0.0508\;[0.0428,0.0602]$
& $0.0508\;[0.0422,0.0608]$
& $0.0508\;[0.0420,0.0601]$
& $0.0508\;[0.0427,0.0581]$ \\

VUNet -- Diffusion
& $0.0125\;[0.0027,0.0199]$
& $0.0125\;[0.0018,0.0183]$
& $0.0125\;[0.0011,0.0185]$
& $0.0125\;[0.0011,0.0145]$ \\

ViTAE -- Diffusion
& $0.0629\;[0.0502,0.0754]$
& $0.0629\;[0.0479,0.0746]$
& $0.0629\;[0.0472,0.0747]$
& $0.0629\;[0.0474,0.0703]$ \\
\bottomrule
\end{tabular}
\end{adjustbox}
\caption{\ans{Sensitivity of paired temporal block-bootstrap comparisons to the temporal block length on the observation data. Values represent the difference in mean relative L2 error between the two models, with the model having higher error listed first and the 95\% confidence interval in brackets. All comparisons are based on time-matched observations.}}
\label{tab:paired_block_sensitivity}
\end{table}

\subsection{Training seed variability}\label{training_seed}
Table \ref{tab:seed_results} shows the variation in mean relative L2 error on simulation and observation dataset when training seed is changed. The results are consistent, with less variability observed for CLSTM and diffusion model

\begin{table}[htbp]
\centering
\begin{adjustbox}{max width=\linewidth,center}
\begin{tabular}{llrrrr}
\toprule
Dataset & Model & Seed 0 & Seed 21 & Seed 42 & Mean $\pm$ Std \\
\midrule
\multirow{4}{*}{Simulation data}
& CLSTM(t=8) & 0.117 & 0.121 & 0.112 & $0.117 \pm 0.0035$ \\
& VUNet(t=4) & 0.121 & 0.115 & 0.109 & $0.115 \pm 0.005$ \\
& ViTAE(t=6) & 0.129 & 0.131 & 0.169 & $0.143 \pm 0.018$ \\
& Diffusion(E=20) & 0.1085 & 0.1087 & 0.109 & $ 0.1087 \pm 0.0002$ \\
\midrule
\multirow{4}{*}{Observation data}
& CLSTM(t=8) & 0.234 & 0.235 & 0.230 & $0.233 \pm 0.0019$ \\
& VUNet(t=4) & 0.302 & 0.282 & 0.267 & $0.284 \pm 0.015$ \\
& ViTAE(t=6) & 0.317 & 0.307 & 0.352 & $0.326 \pm 0.019$ \\
& Diffusion(E=20) & 0.256 & 0.265 & 0.256 & $ 0.259 \pm 0.004$ \\
\bottomrule
\end{tabular}
\end{adjustbox}
\caption{\an{Relative L2 error on test dataset of different models across three training seeds on simulation and observation data}}
\label{tab:seed_results}
\end{table}

\subsection{Uncertainty analysis of Kriging and VUNet}

\begin{figure}[htbp]
    \centering
    \includegraphics[width=1.0\linewidth]{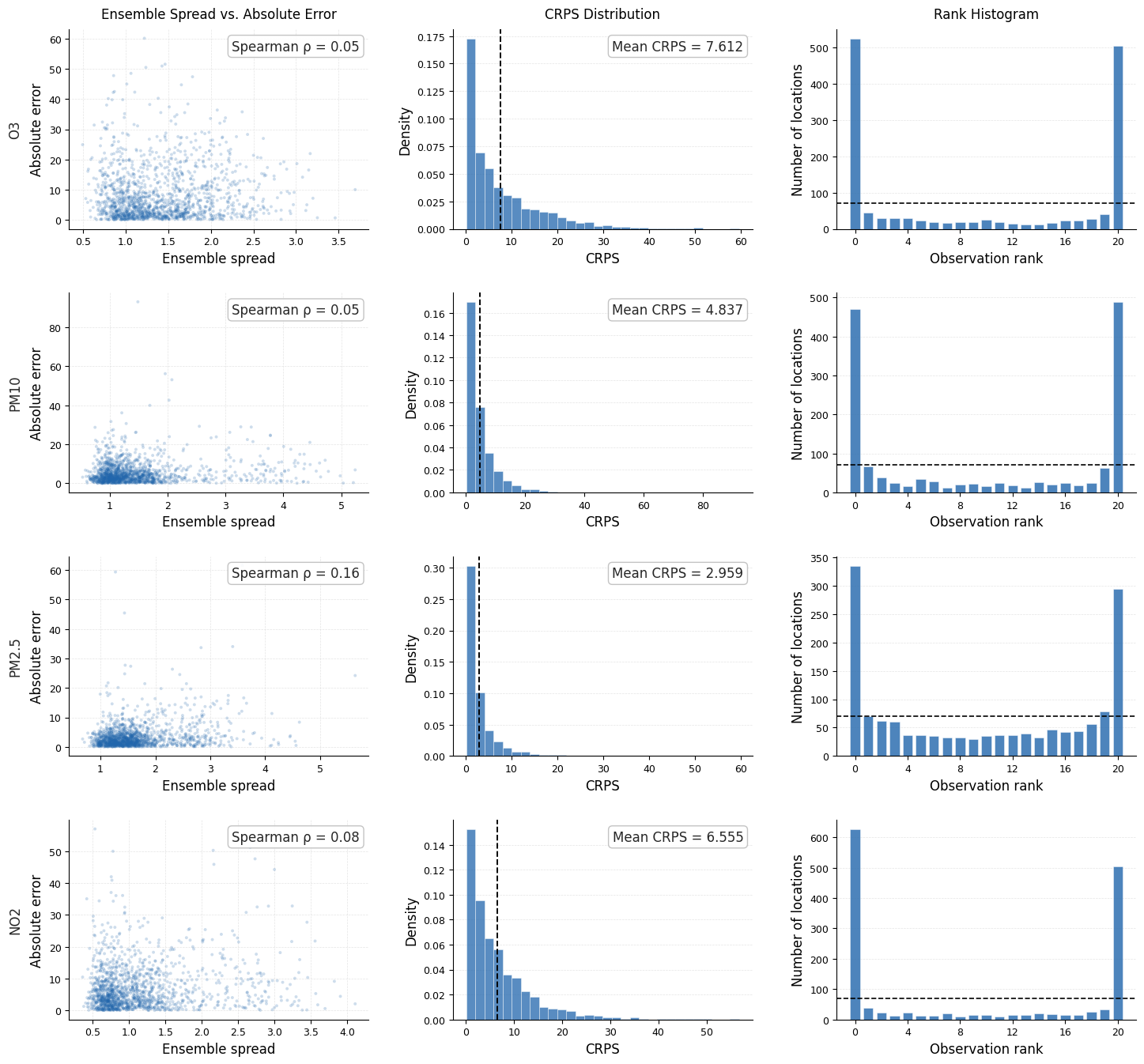}
    \caption{\an{\textbf{Uncertainty analysis of the Kriging model.} Uncertainty analysis of ensemble prediction of Kriging model on held-out stations of real observation data}}
    \label{fig:uq_kriging}
\end{figure}

\an{Ordinary Kriging provides a spatial prediction and its corresponding variance which represents the uncertainty due to spatial interpolation given the observed points and the fitted variogram. We generate 20 predictions by sampling from the Gaussian predictive distribution at each grid location. This allows us to perform uncertainty analysis and plot the ensemble spread, CRPS distribution and rank histogram (Figure \ref{fig:uq_kriging}).}

\begin{figure}[htbp]
    \centering
    \includegraphics[width=1.0\linewidth]{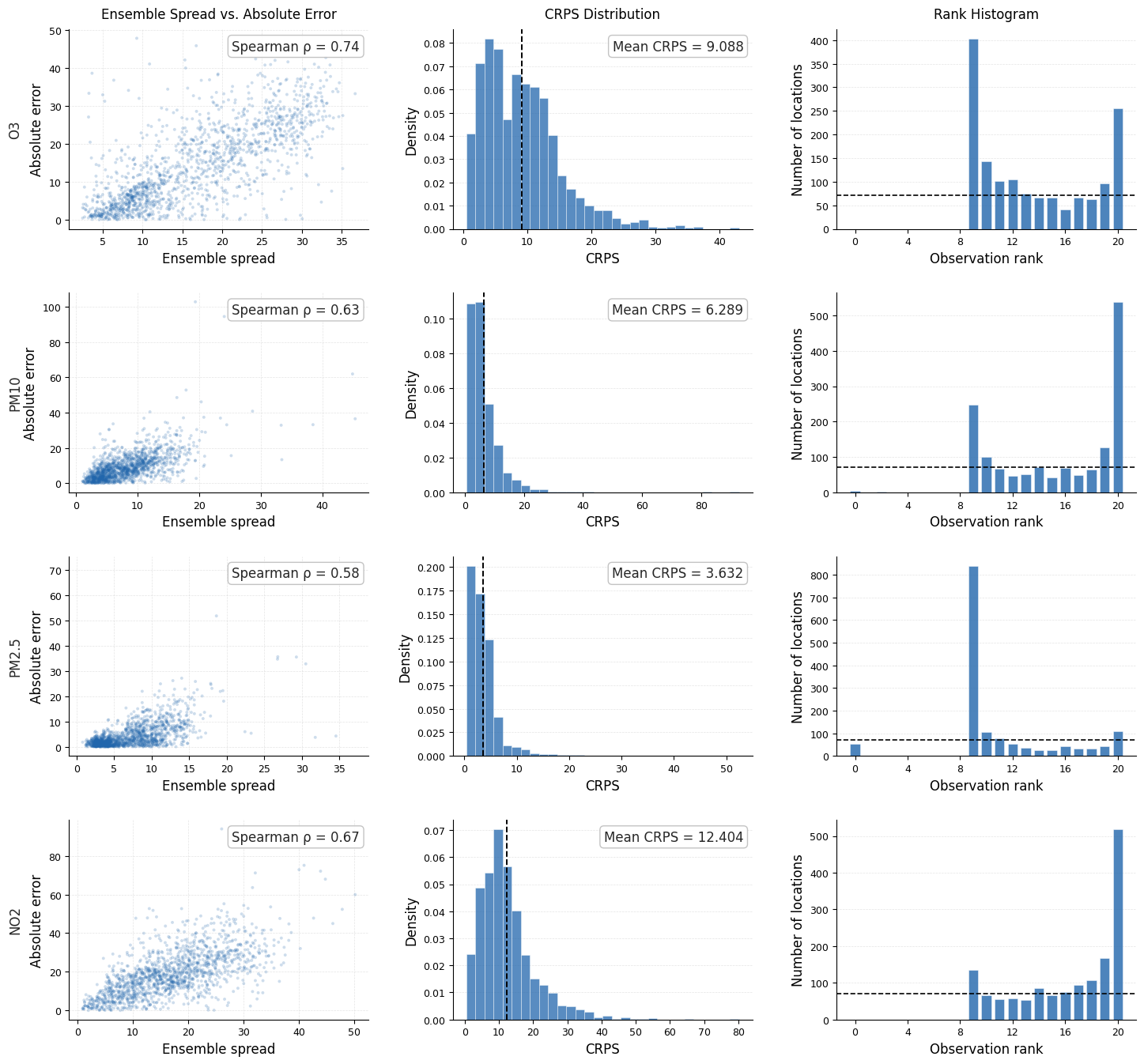}
    \caption{\an{\textbf{Uncertainty analysis of the VUNet model.}  Uncertainty analysis on ensemble prediction of VUNet on held-out stations of real observation data}}
    \label{fig:uq_vunet}
\end{figure}

\an{We train the deterministic VUNet model with 20 different weight initialisation seeds to produce an ensemble prediction. Supplementary Figure \ref{fig:uq_vunet} plots the uncertainty analysis for the deterministic VUNet model. The stochasticity is introduced by using 20 different model weight initialisations and the variability represents sensitivity to model parameters. In this experiment we test if the ensemble diversity arising from parameter initialisation can be a surrogate for the predictive uncertainty. The spread-error correlation is high because the ensemble error only arises from one source: model initialisation. The high concentration of the rank histogram in one bin indicates that the variability across independently trained networks is limited and does not represent the variability of observations. For PM$_{10}$ and NO$_2$, the ensemble is underestimating the observations and the spread is not sufficient to compensate for the bias. By contrast, for O$_3$ and PM$_{2.5}$ the observed value frequently falls within the relatively narrow range near the middle of the ensemble. This is because each VUNet model is trained to minimise the MSE, so every trained model produces similar estimates of the conditional mean. Thus, introducing variability from training seeds provides some ensemble variability but does not necessarily provide a calibrated representation of predictive uncertainty. }

\subsection{\ans{Empirical coverage of diffusion model}}

\ans{We evaluate the calibration of the 20-member diffusion ensemble by constructing intervals at nominal coverage levels of 50\%, 80\%, 90\%, and 95\%. For each interval, we measure empirical coverage, defined as the fraction of observations whose ground-truth values fall within the predicted interval, and the corresponding mean interval width. Table \ref{tab:coverage} reports the empirical coverage and mean interval width, showing underdispersion across all nominal levels. Inflating the ensemble spread may improve calibration, although its effectiveness would need to be evaluated in future work.}

\begin{table}[htbp]
\centering
\begin{adjustbox}{max width=\linewidth,center}
\begin{tabular}{lcccccccc}
\toprule
& \multicolumn{2}{c}{50\%}
& \multicolumn{2}{c}{80\%}
& \multicolumn{2}{c}{90\%}
& \multicolumn{2}{c}{95\%} \\
\cmidrule(lr){2-3}
\cmidrule(lr){4-5}
\cmidrule(lr){6-7}
\cmidrule(lr){8-9}
& Coverage (\%) & Width
& Coverage (\%) & Width
& Coverage (\%) & Width
& Coverage (\%) & Width \\
\midrule
O$_3$      & 37.01 & 13.21 & 61.57 & 23.30 & 71.00 & 27.11 & 75.03 & 29.43 \\
PM$_{10}$  & 39.64 &  7.88 & 63.12 & 14.22 & 70.12 & 17.13 & 75.03 & 19.04 \\
PM$_{2.5}$ & 36.25 &  6.43 & 59.37 & 11.58 & 67.52 & 14.08 & 73.11 & 15.83 \\
NO$_2$     & 38.96 & 14.92 & 62.58 & 26.41 & 68.37 & 30.77 & 72.48 & 33.59 \\
\bottomrule
\end{tabular}
\end{adjustbox}
\caption{\ans{Empirical coverage and interval width of the diffusion ensemble for nominal coverage levels of 50\%, 80\%, 90\%, and 95\%. For each nominal level, the two subcolumns report empirical coverage (\%) and interval width, respectively.}}
\label{tab:coverage}
\end{table}

\section{Diffusion-based generative model}\label{appendix_diffusion}

\subsection{Mathematical Formulation}\label{appendix_diff_math}

We work with a class of diffusion models called \textit{score-based stochastic differential equations}~\cite{song2020score}, where the data is continuously perturbed over time. In this section, variable $r$ represents the time parameter governing the noise addition process. The forward SDE adds noise to the original image as follows:
\begin{equation}
    \mathbf{dx} = \mathbf{f}(\mathbf{x},r) \mathrm{d}r + g(r) \mathrm{d}\mathbf{w},
\end{equation}
where $\mathbf{f}(\mathbf{x},r)$ is the drift coefficient, $g(r)$ is the diffusion coefficient, and $\mathbf{w}$ is the standard Wiener process. When going back in time (denoising), the reverse-time process can also be shown to be a diffusion process given by the equation:
\begin{equation}
    \mathrm{d}\mathbf{x} = [\mathbf{f}(\mathbf{x},r) - g^2(r) \nabla_x \log p_r (\mathbf{x})]\mathrm{d}r + g(r)\mathrm{d}\mathbf{w}.
    \label{eq:reverse}
\end{equation}
The term $\nabla_x \log p_r (\mathbf{x})$ is the \textit{score function}~\cite{song2020score}, which is estimated through a trained neural network. The choice of coefficients $\mathbf{f}(\mathbf{x},r)$ and $g(r)$ determines the noise addition process. The Elucidating Diffusion Model (EDM)~\cite{karras2022elucidating} provides a framework with $\mathbf{f}(r) = 0$ and $g(r) = \sqrt{2 \dot{\sigma}(r) \sigma(r) }$, where $\sigma(r)$ is the noise standard deviation added at time $r$.

Let $D_\theta (\mathbf{x}; \sigma)$ be the neural network parametrised by $\theta$ that returns the original image for a noisy image input $\mathbf{x}$. The training objective of the neural net is to minimise the mean squared error between the model output and the original image:
\begin{equation}
    {\mathbb E}_{\mathbf{x} \sim p_{\textrm{data}}} {\mathbb E}_{n\sim \mathcal{N}(\mathbf{0},\sigma^2\mathbf{I})} \left\|D_\theta (\mathbf{x}+\mathbf{n};\sigma) - \mathbf{x}\right\|_2^2.
    \label{loss-eq}
\end{equation}
From the above equation, Karras et al.~\cite{karras2022elucidating} proved that
\begin{equation}
    \nabla_x \log p(\mathbf{x}; \sigma) = \frac{D_{\theta}(\mathbf{x}; \sigma) - \mathbf{x}}{\sigma^2}.
\label{eq:denoiser}
\end{equation}
Thus, minimising the loss in equation \ref{loss-eq} provides a relation between the \textit{score function} and the neural net denoiser $D_{\theta}$. The reverse SDE in equation \ref{eq:reverse} is solved using DPM Solver++~\cite{lu2025dpm}. This algorithm is more stable and provides faster inference with high-quality generation. Compared to traditional DDIM samplers~\cite{song2020denoising}, which require hundreds of steps, DPM Solver++ can achieve high-quality samples within 15 to 20 steps~\cite{lu2025dpm}.

\subsection{Imputation}\label{appendix_diff_imputation}

Diffusion models generate random samples from the learned distribution. Instead, imputation~\cite{song2020score} is a technique to guide the back sampling process given observation points. We condition the denoising process with Voronoi tessellations of the observed data from monitoring stations. This is achieved by injecting the Voronoi embedding into the layers of the denoiser $D_{\theta}$ through a cross attention mechanism. \\

 Furthermore, we apply \textit{masked} back sampling using sparse observation points to guide generation. Since the values from monitoring stations are treated as ground truth, we preserve them during sample generation. \an{The observations are first re-noised to match the current noise level,
\begin{equation}
    \mathbf{\tilde{x}}_{r-1}^t = \mathbf{x}_0^t + \sigma_{r-1}\mathbf{\epsilon}, \quad \mathbf{\epsilon} \sim \mathcal{N}(0, \mathbf{I})
\end{equation}
and are then reinserted into the generated sample at the observed locations
\begin{equation}
    \mathbf{x}^t_{r-1} = \hat{\mathbf{x}}^t_{r-1} \odot (1 - \mathbf{\Omega}_t) + \mathbf{\tilde{x}}_{r-1}^t \odot \mathbf{\Omega}_t
\end{equation}
}
The DPM Solver++ predicts a one-step denoised image $\mathbf{x}_{r-1}^{t}$ given $\mathbf{y}^t$. The above process is repeated $R$ times to obtain the final generated sample.

\section{Evaluation Metrics}\label{appendix_metrics}

The model's performance is evaluated on the simulation data using relative L2 error, Mean Fractional Error (MFE), Mean Fractional Bias (MFB) and Structural Similarity Index Measure (SSIM) \cite{wang2004image}.

\begin{equation}
    \textrm{Relative L2} = \frac{1}{T} \sum_{t=1}^T \frac{\left\|\mathbf{y}_t - \mathbf{x}_t\right\|_2}{\left\|\mathbf{x}_t\right\|_2 + \an{\epsilon}}
\end{equation}

\begin{equation}
    \mathrm{MFE} =
\frac{1}{T \times n}
\sum_{t=1}^{T}
\sum_{i=1}^{n}
\frac{2\left| y_{t,i} - x_{t,i} \right|}
{x_{t,i} + y_{t,i} + \an{\epsilon}}
\end{equation}

\begin{equation}
    \mathrm{MFB} =
\frac{1}{T \times n}
\sum_{t=1}^{T}
\sum_{i=1}^{n}
\frac{2\left( y_{t,i} - x_{t,i} \right)}
{x_{t,i} + y_{t,i} + \an{\epsilon}}
\end{equation}

where $\mathbf{x}_t \in \mathbb{R}^{4\times n_x \times n_y }$ is the reference field, $\mathbf{y}_t \in \mathbb{R}^{4\times n_x \times n_y }$ is the reconstructed image and $T$ is the size of the test dataset. \an{A small term $\epsilon$ is added to the denominator to handle near-zero concentration values.} $y_{t,i}, x_{t,i}$ denote the pixel values; for example,
$\mathbf{x}_t = [x_{t,1}, \ldots, x_{t,n}]$ with $n = 4 \times n_x \times n_y$.

These metrics quantify the error at the pixel level but they do not necessarily reflect the preservation of structural information in reconstructed fields. To address this limitation, SSIM aims to compare patterns between images rather than individual pixel intensities, thereby providing a measure that is more closely aligned with human visual perception.

We lack ground truth data to evaluate real-world observation data. Thus, we apply the binary evaluation mask to the data before calculating the metrics. For example, the relative L2 error metric is computed as follows:

\begin{equation}
    \mathrm{Relative L2} = \frac{1}{T} \sum_{t=1}^T \frac{\left\| \mathbf{y}_t \odot \mathbf{\Omega}_t^{\textrm{eval}} - \mathbf{x}_t \odot \mathbf{\Omega}_t^{\textrm{eval}}\right\|_2}{\left\| \mathbf{x}_t \odot \mathbf{\Omega}_t^{\textrm{eval}} \right\|_2 + \an{\epsilon}}.
\end{equation}

To compute the power spectrum plots, each predicted output is converted to its 2D Fourier representation. It is then shifted such that the zero-frequency component is positioned at the centre. The power spectrum is defined as the squared magnitude of the Fourier coefficients. We then compute the radial average of the spectral image (the mean of values that are located equidistant from the centre). The radial mean spectrum, which is a function of frequency, is averaged over all samples and pollutants, resulting in the power spectrum plot.

\section{Data augmentation methods}\label{appendix_augmen}

This study employs a range of data augmentation techniques to reduce the distributional discrepancy between observed and simulated data. These methods were designed to produce realistic augmentations that respect the local correlations present in pollution fields. Each noise generation method has multiple hyperparameters, as well as a mean and standard deviation. Method-specific hyperparameters and their function are described in Table \ref{tab:noise_hyperparams}.

\begin{table}[htbp]
\caption{Table enumerating and describing noise generation method-specific hyperparameters.}
\label{tab:noise_hyperparams}
\centering
\begin{tabularx}{\linewidth}{|l|l|X|}
\hline
\textbf{Method} & \textbf{Hyperparameter} & \textbf{Function} \\[0.8ex]  %
\hline
Gaussian
    & sigma & The size of the Gaussian kernel. \\
\hline
\multirow{2}{*}{Time-aware Gaussian}
    & sigma & The size of the spatial Gaussian kernel. \\
    & time sigma & The size of the temporal Gaussian kernel. \\
\hline
\multirow{4}{*}{Perlin}
    & base resolution & Controls the coarseness of the base noise grid. \\
    & octaves & The number of noise layers used for fractal noise. \\
    & persistence & Controls the amplitude of each successive octave. \\
    & lacunarity & Controls the frequency of each successive octave. \\
\hline
Cross-correlation
    & - & - \\
\hline
\end{tabularx}
\end{table}

\subsection{Gaussian perturbation}

This perturbation method is the simplest and fastest method used in this study. The main idea is to generate white noise with the desired shape, usually in $\mathbb{R}^{t\times c\times n_x\times n_y}$, and to apply a 2D convolution over the last two dimensions. A kernel with Gaussian weights is used to avoid the generation of sharp transitions~\cite{gonzalez2009digital} while also providing a very rudimentary cloud-like appearance to the noise.

\subsection{Time-aware Gaussian perturbation}

This method extends the Gaussian perturbation approach by introducing temporal structure, generating perturbations that are correlated across consecutive time steps in a manner analogous to Gaussian process covariance models \cite{williams2006gaussian}. It is the only augmentation strategy considered in this study that explicitly incorporates temporal correlations. This is achieved by generating independent Gaussian noise examples before convolving along the time dimension with a Gaussian kernel.

\subsection{Perlin perturbation}

Perlin perturbation was initially developed to generate natural-looking textures in computer graphics~\cite{perlin_noise}, but over time has evolved into a more general tool for procedural generation. Notably, its use for realistic cloud generation~\cite{perlin_noise_clouds} is of particular interest for this study. With its ability to create smooth yet irregular patterns, this method could improve the models' robustness by introducing realistic cloud-like noise.

 Perlin perturbation generates a grid of randomly oriented vectors that define local directional changes. To compute the noise value at a given point, the algorithm calculates contributions from the surrounding grid corners and interpolates them. Fractal Perlin noise extends this concept by summing multiple layers ("octaves") of Perlin noise at increasing spatial frequencies ("lacunarity") and decreasing amplitudes ("persistence").

\subsection{Cross-correlation perturbation}
This method generates perturbations based on the cross-correlation structure among pixels in the training dataset. By estimating inter-pixel correlations and applying the corresponding covariance transformation, the method produces spatially correlated noise fields that better reflect the statistical properties of air-pollution patterns. As cross-correlation is an image-specific measure, the mean across the time dimension is computed. Using the Cholesky decomposition of the cross-correlation matrix, white noise is converted into cross-correlated noise.

 Among the augmentation strategies considered, this method generates the most statistically realistic perturbations, as it explicitly preserves the spatial correlation structure of the training data. However, this reliance on training-set statistics may limit the diversity of the generated samples, potentially reducing its effectiveness in improving model robustness to out-of-distribution observations.

\section{Model architectures}\label{appendix_model_figs}

\begin{table}[htbp]
\centering
\begin{adjustbox}{max width=\linewidth,center}
\begin{tabular}{lcc}
\hline
\textbf{Model} & \textbf{Trainable Parameters} & \textbf{Runtime (s)} \\
\hline
VUNet (t=4)     & 15,482,884 & 2.464 \\
ViTAE (t=6)     & 18,798,024 & 2.228 \\
CLSTM (t=8)    & 2,522,244  & 66.905 \\
Diffusion (E=20) & 27,529,988 & 197.911 \\
Kriging & - & 15.187 \\
\hline
\end{tabular}
\end{adjustbox}
\caption{Number of trainable parameters and the inference time of models on the test period of observation dataset. The diffusion model is an ensemble of 20 predictions.}
\label{tab:model_parameter}
\end{table}

\begin{figure}[H]
    \centering
    \includegraphics[width=\linewidth]{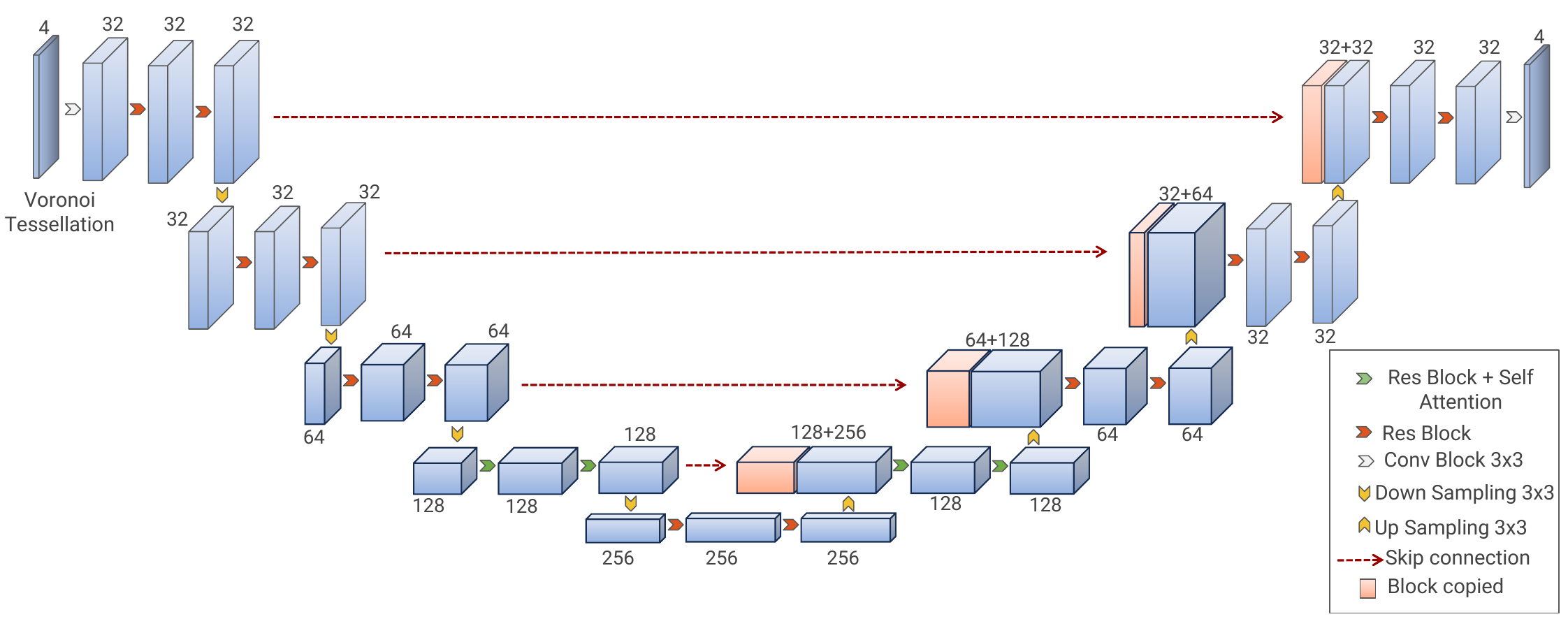}
    \caption{\textbf{Model architecture of the VUNet model.} The visual representation of layers of VUNet model}
    \label{fig:VUNet}
\end{figure}

\begin{figure}[H]
    \centering
    \includegraphics[scale=0.32]{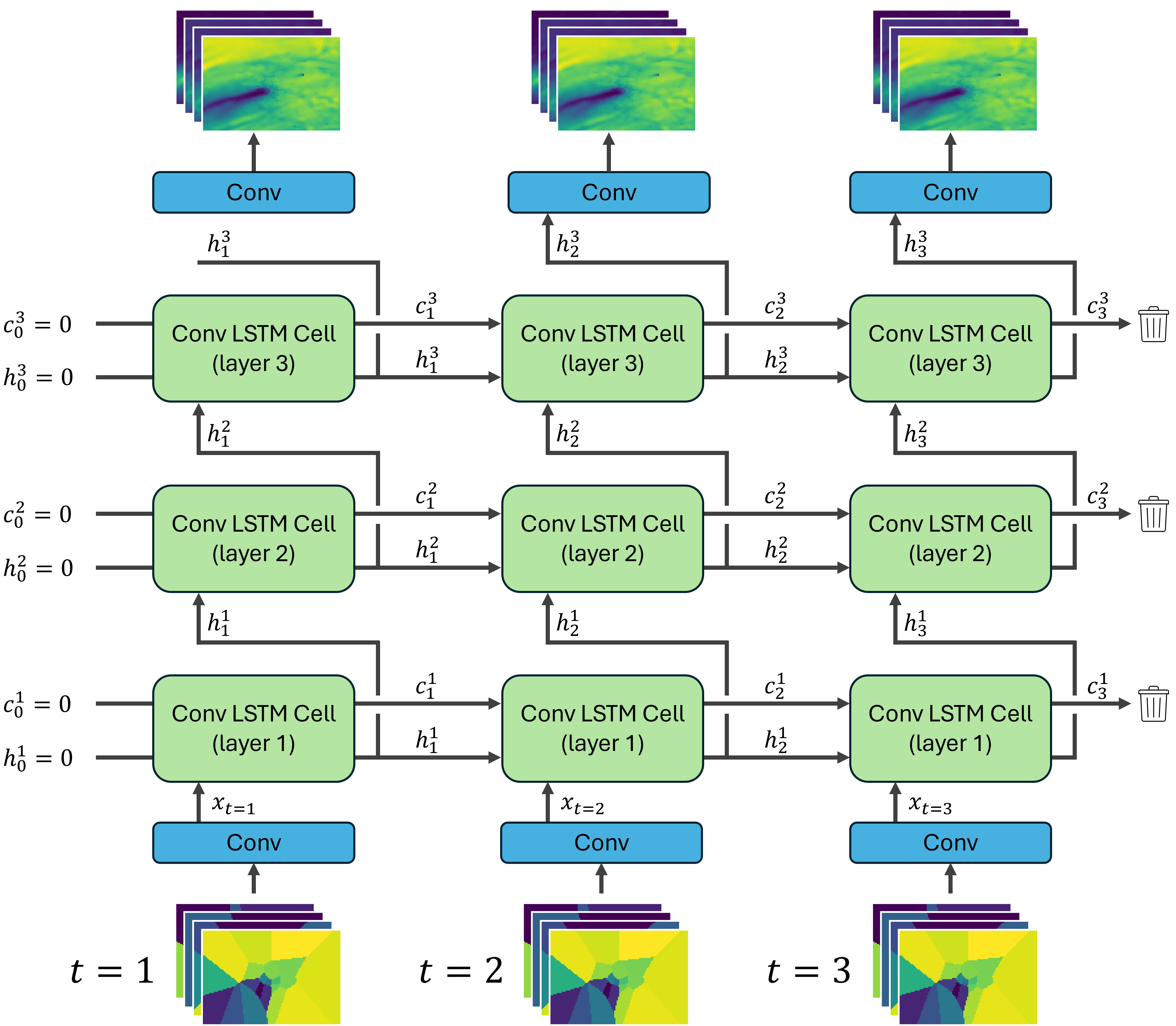}
    \caption{\textbf{Model architecture of the CLSTM model.} The visual representation of CLSTM model with time steps $k=3$.}
    \label{fig:clstm}
\end{figure}

VUNet (Figure \ref{fig:VUNet}) is a state-of-the-art CNN model that excels at capturing local spatial interactions, whereas ViTAE is a transformer-based model designed to model long-range and global dependencies. It consists of a transformer encoder followed by a UNet decoder. By contrast, CLSTM (Figure \ref{fig:clstm}) specialises in learning temporal sequences of spatial data. Kriging is a traditional interpolation method~\cite{RefWorks:RefID:36-kleijnen2009kriging} provided as a baseline for this study. From table \ref{tab:model_parameter}, ViTAE has the highest number of trainable parameters among the deterministic models, followed by VUNet and CLSTM.

\begin{figure}[H]
    \centering
    \includegraphics[width=\linewidth]{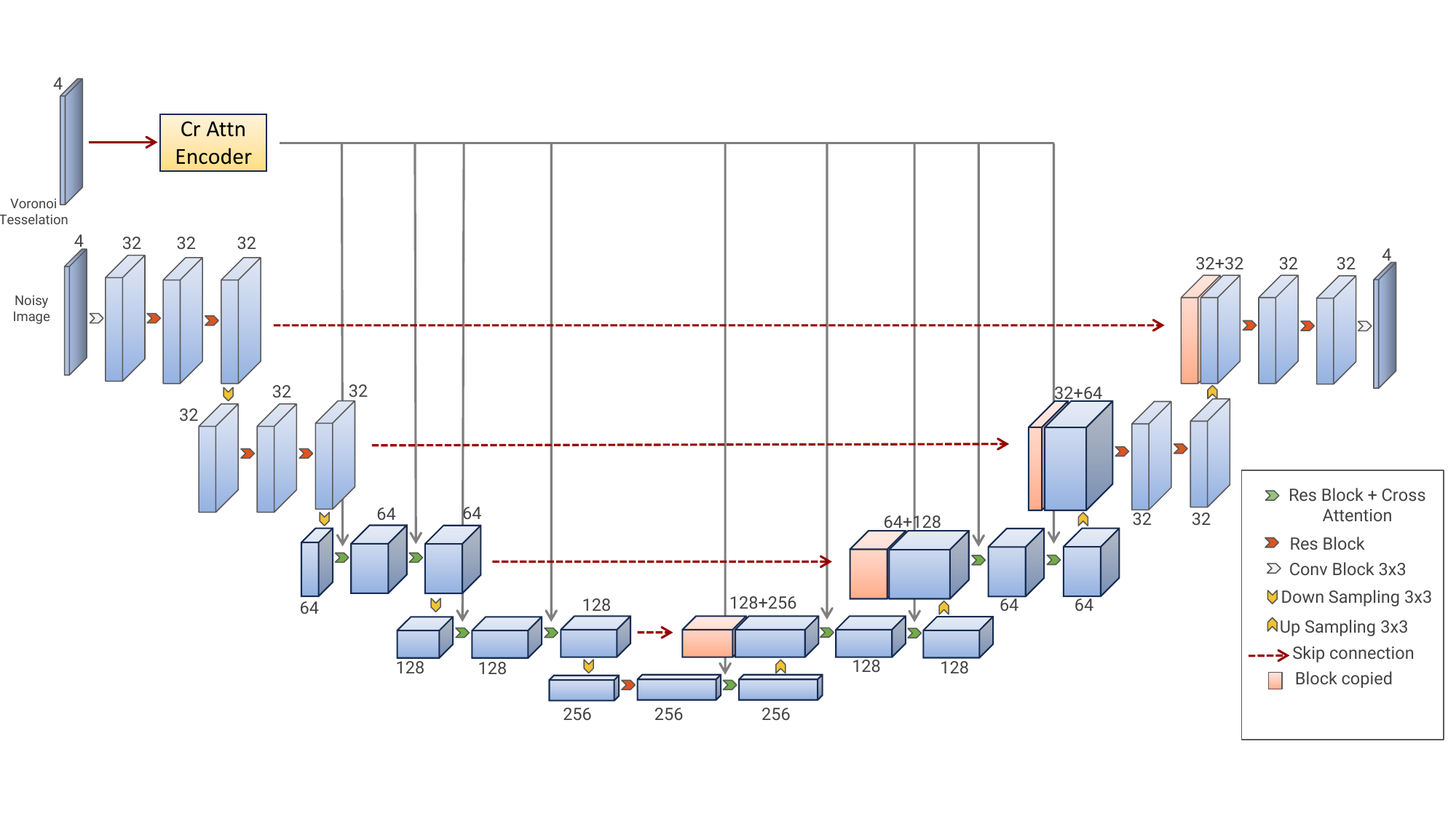}
    \caption{\textbf{Model architecture of the Denoiser.} The visual representation of the denoiser used in the diffusion model. }
    \label{fig:gen_model}
\end{figure}

The neural network used in the diffusion model is a simple UNet along with a Voronoi encoder (Figure \ref{fig:gen_model}) which performs cross-attention between the intermediate layers of the UNet and the latent-variable output of the Voronoi encoder.

\ans{All models are trained and evaluated using NVIDIA RTX 5000 Ada Generation GPU with 32 GB of GPU memory. The Kriging model is run on CPU with an Intel Xeon W-2133 processor (6 cores). Table \ref{tab:model_parameter} shows the evaluation runtime (in seconds) of the models on the test period (December) of observation data. }

\an{During training of diffusion process, instead of uniform noise sampling, the EDM framework \cite{karras2022elucidating} recommends noise to be sampled as $\ln \sigma \sim \mathcal{N}(P_{\textrm{mean}}, P_{\textrm{std}})$. We assign a slightly higher value of $P_{\textrm{mean}} = 1.2$, $P_{\textrm{std}} = 1.7$. Figure~\ref{fig:noise_clean_distribution_comparison} in the main text depicts that the data distribution resembles a non-Gaussian heavy-tailed distribution. This implies that there are fewer points with high pollutant concentration values. Since the model has to learn to reconstruct fields across a wide range of values, a higher noise distribution is chosen for the scheduler.}

\an{Table \ref{fig:hyperparameters} provides model hyperparameters used to train each model. These are selected by performing Bayesian optimisation using the TPE (Tree-structured Parzen Estimator) algorithm \cite{bergstra2011algorithms} to minimise the relative L2 error on the validation dataset.}

\begin{table}[htbp]
\centering
\begin{adjustbox}{max width=\linewidth,center}
\begin{tabular}{l l l}
\hline
\textbf{Model} & \textbf{Hyperparameter} & \textbf{Value} \\
\hline
\multirow{4}{*}{VUNet}
    & lr & 0.0001 \\
    & weight\_decay & 0.001 \\
    & kernel\_size & 5 \\
    & use\_norm & False \\
\hline
\multirow{6}{*}{CLSTM}
    & lr & 0.0001 \\
    & weight\_decay & 1e-05 \\
    & kernel\_size & 7 \\
    & use\_norm & False \\
    & n\_layers & 2 \\
    & hidden\_channels & 64 \\
\hline
\multirow{4}{*}{VITAE}
    & lr & 0.001 \\
    & weight\_decay & 1e-05 \\
    & $\lambda$ & 0.1 \\
    & kernel\_size & 5 \\
\hline
\multirow{5}{*}{Diffusion}
    & lr & 0.0001 \\
    & weight\_decay & 1e-07 \\
    & cross attention dim & 256 \\
    & noise schedule P\_mean & 1.2 \\
    & noise schedule P\_std & 1.7 \\
\hline
\multirow{7}{*}{Kriging}
    & variogram model & linear \\
    & nugget &  0.0288 \\
    & nlags &  9 \\
    & weight & True \\
    & anisotropy\_scaling & 1.641 \\
    & anisotropy\_angle & 179.1 \\
    & slope & 0.213 \\
\hline
\end{tabular}
\end{adjustbox}
\caption{Model Hyperparameters}
\label{fig:hyperparameters}
\end{table}

\end{document}